\documentclass[journal]{IEEEtran}

\usepackage{amsmath,amsfonts,amssymb}
\usepackage{algorithm}
\usepackage{algorithmic}
\usepackage{array}
\usepackage{booktabs}
\usepackage{multirow}
\usepackage{makecell}
\usepackage{graphicx}
\usepackage[caption=false,font=normalsize,labelfont=sf,textfont=sf]{subfig}
\usepackage{textcomp}
\usepackage{stfloats}
\usepackage{url}
\usepackage{xspace}
\usepackage[table]{xcolor}
\usepackage{pifont}
\usepackage{cite}
\usepackage{balance} 

\newcommand{\tbranch}{\hspace{0.6em}\makebox[1.15em][l]{
  \rule[-0.55ex]{0.45pt}{1.65ex}\kern-0.45pt\rule[0.25ex]{0.85em}{0.45pt}}}
\newcommand{\tlast}{\hspace{0.6em}\makebox[1.15em][l]{
  \rule[0.25ex]{0.45pt}{0.85ex}\kern-0.45pt\rule[0.25ex]{0.85em}{0.45pt}}}

\newcommand{\cmark}{\ding{51}}
\newcommand{\xmark}{\ding{55}}

\renewcommand{\dbltopfraction}{0.95}


\newcommand{\fitwidth}{\ifdim\width>\linewidth\linewidth\else\width\fi}

\makeatletter
\long\def\@makecaption#1#2{%
\ifx\@captype\@IEEEtablestring%
\footnotesize\bgroup\par\centering\@IEEEtabletopskipstrut{\normalfont\footnotesize #1}\\{\normalfont\footnotesize #2}\par\addvspace{0.5\baselineskip}\egroup%
\@IEEEtablecaptionsepspace
\else
\@IEEEfigurecaptionsepspace
\setbox\@tempboxa\hbox{\normalfont\footnotesize {#1.}\nobreakspace\nobreakspace #2}%
\ifdim \wd\@tempboxa >\hsize%
\setbox\@tempboxa\hbox{\normalfont\footnotesize {#1.}\nobreakspace\nobreakspace}%
\parbox[t]{\hsize}{\normalfont\footnotesize\noindent\unhbox\@tempboxa#2}%
\else%
\hbox to\hsize{\normalfont\footnotesize\box\@tempboxa\hfil}%
\fi\fi}
\makeatother

\newcommand{\method}{BridgeVLA\xspace}
\newcommand{\memmethod}{BridgeVLA++\xspace}
\newcommand{\eg}{\textit{e.g.}\xspace}

\usepackage[hidelinks,breaklinks=true,bookmarksnumbered=true]{hyperref}

\begin{document}

\bstctlcite{IEEEexample:BSTcontrol}

\title{\memmethod: A Data-Efficient, Generalizable, and Memory-Augmented Vision-Language-Action Framework for 3D Manipulation}
\author{
Peiyan Li\textsuperscript{*}\thanks{\textsuperscript{*} Equal Contribution.}, Yuze Zhu\textsuperscript{*}, Yixiang Chen, Qisen Ma, Yuan Xu, Jiabing Yang, He Guan\\
Yan Huang\textsuperscript{\textdagger}\thanks{\textsuperscript{\textdagger} Corresponding Author.}, Hongtao Wu, Xiao Ma,
Tao Kong, Liang Wang, \textit{Fellow, IEEE}, Tieniu Tan, \textit{Fellow, IEEE}%
\thanks{Peiyan Li, Yuze Zhu, Yixiang Chen, Qisen Ma, Yuan Xu, Jiabing Yang, Yan Huang, Liang Wang and Tieniu Tan are with the New Laboratory of Pattern Recognition (NLPR), Institute of Automation, Chinese Academy of Sciences, Beijing, China, and with the School of Artificial Intelligence, University of Chinese Academy of Sciences, Beijing, China.}%
\thanks{He Guan is with FiveAges, Beijing, China. Yan Huang is also with FiveAges.}%
\thanks{Hongtao Wu, Xiao Ma, and Tao Kong contribute to this work when they were with ByteDance Seed.}}

\markboth{IEEE Transactions on Pattern Analysis and Machine Intelligence}%
{\memmethod: Memory-Augmented Vision-Language-Action Learning for 3D Manipulation}

\maketitle

\begin{abstract}
Leveraging pre-trained vision-language models (VLMs) to construct vision-language-action (VLA) models has emerged as a promising paradigm
for 3D robot manipulation. However, existing 3D VLA methods remain data-hungry, exhibit limited generalization under distribution shifts, and lack explicit
memory of past observations. These limitations hinder their application to data-scarce, open-world, and memory-dependent manipulation scenarios.
Our previous work, BridgeVLA, improves data efficiency and generalization by preserving the input--output alignment of a pre-trained VLM during
3D action learning: raw point clouds are projected into multi-view images, and intermediate heatmaps are predicted before generating robot actions.
In this work, we develop BridgeVLA++ by equipping BridgeVLA with a unified spatio-temporal memory architecture that models persistent spatial context and temporal interaction history. The resulting memory-augmented framework can reason over observation histories while preserving BridgeVLA's data efficiency and generalization capabilities. Extensive experiments show that our framework achieves strong performance on spatial manipulation tasks while exhibiting robust generalization. BridgeVLA++ further achieves state-of-the-art performance on two challenging memory-dependent manipulation benchmarks without sacrificing the data efficiency and generalization of the original BridgeVLA. In addition, BridgeVLA++ performs effectively in bimanual manipulation settings and is validated on an additional real-world robotic platform, demonstrating its scalability across tasks, environments, and robotic platforms. These results establish BridgeVLA++ as a unified 3D vision-language-action framework that simultaneously supports data-efficient learning, robust generalization, and effective memory-aware robot manipulation.\footnote{Project website: \textcolor[HTML]{0969DA}{\url{https://bridgevla-plus.github.io/}}.}
\end{abstract}


\section{Introduction}
\label{sec:introduction}

\begin{figure*}[t!]
  \centering
  \includegraphics[width=\linewidth]{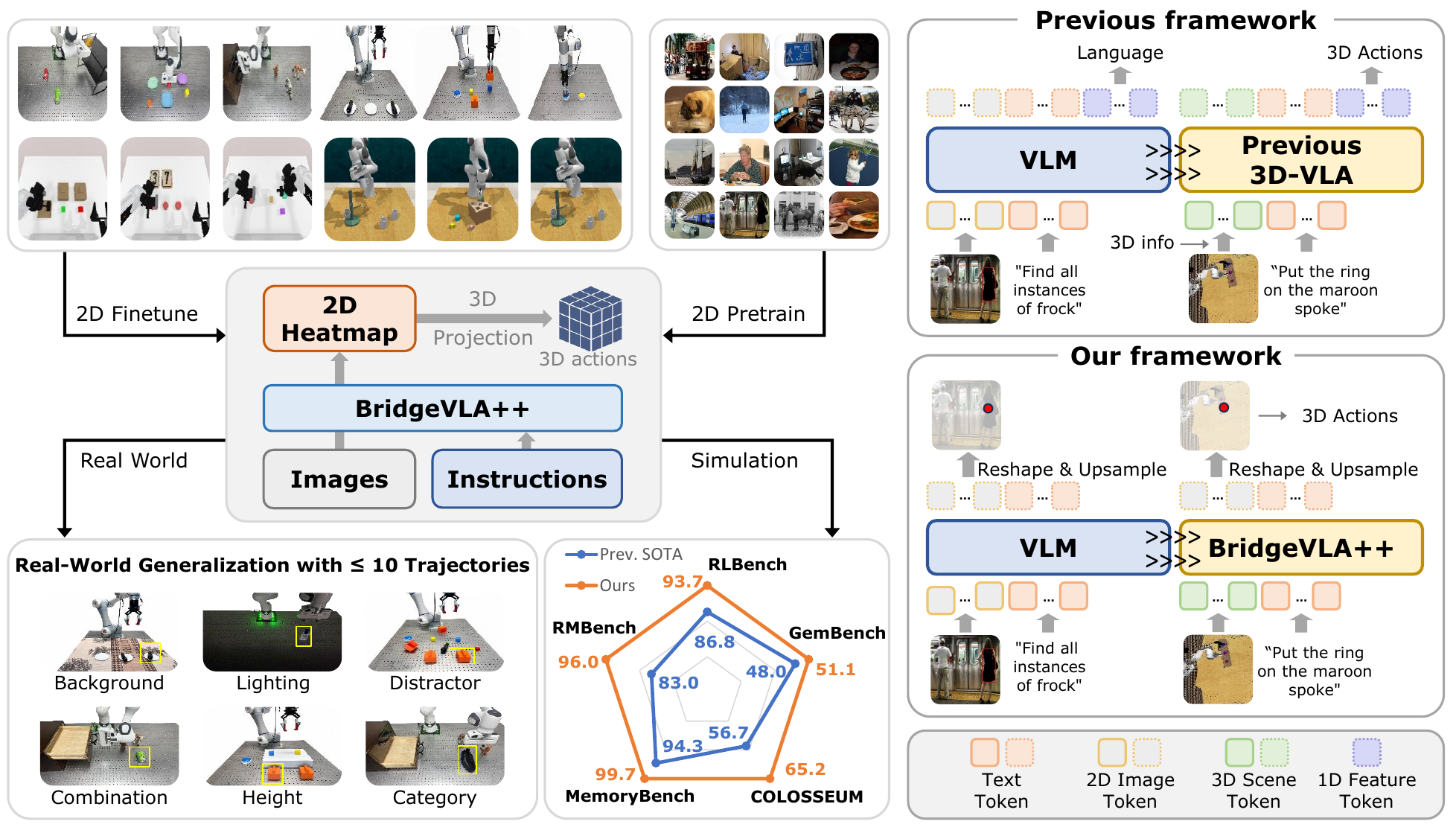}
  \caption{\textbf{Overview.}
  \method\ is a 3D VLA framework that aligns its inputs and outputs in a
  unified 2D image space. It is pre-trained on object grounding using 2D heatmaps
  and fine-tuned on action prediction for 3D manipulation.
  \memmethod extends \method with a unified spatio-temporal memory architecture
  in which temporal memory preserves interaction history to determine
  \emph{what to do next}, whereas spatial memory restores previously observed
  geometry to determine \emph{where exactly to act}.
  Experiments in both simulated and real-world settings demonstrate that
  \memmethod effectively handles memory-dependent and memory-free tasks while
  preserving \method's data efficiency and generalization capabilities.}
  \label{fig:teaser}
\end{figure*}

\IEEEPARstart{L}{everaging} pre-trained vision-language models (VLMs) to
construct vision-language-action (VLA) models has become a promising
approach to learning generalizable and robust robot manipulation policies
\cite{kimopenvla,intelligence2025pi_,yu2026wall,li2023vision,brohan2023rt}.
However, most VLA models operate on 2D images and require large amounts of
robot data.
In contrast, 3D manipulation policies exploit geometric
structure and achieve higher sample efficiency
\cite{shridhar2023perceiver,3d-da,gervet2023act3d,goyal2023rvt,goyal2024rvt}.
This raises a question: can a unified 3D VLA model combine the
semantic generalization of pre-trained VLMs with the geometric efficiency
of 3D manipulation policies?

Existing attempts to build 3D VLAs do not fully resolve this challenge
\cite{zhen20243d,qu2025spatialvla}.
Many methods encode actions as token sequences and predict them
autoregressively, thereby discarding the spatial correspondence between 3D
observations and actions that underlies the efficiency of prior 3D policies.
Moreover, introducing 3D inputs into a VLM creates a
modality gap from its 2D image pre-training.
The resulting misalignment limits both the transfer of VLM
priors and the exploitation of explicit 3D structure.

Beyond data efficiency and generalization, memory presents an additional
challenge.
Most VLA and 3D manipulation policies predict each action primarily from the
current observation.
They therefore struggle when the correct action depends on previous
interactions or when task-relevant geometry observed earlier becomes
occluded during execution.
A capable 3D VLA should retain both temporal task context and persistent
spatial information while preserving its original data efficiency and
generalization ability.

To address the first two challenges, our previous work introduced
\method, a 3D VLA framework based on input--output alignment.
\method projects point-cloud observations into multi-view orthographic
images \cite{goyal2023rvt,goyal2024rvt} and processes them with a
pre-trained VLM.
Instead of predicting actions as tokens, it predicts a 2D translational
heatmap for each view and back-projects the heatmap maxima into a 3D
end-effector position.
A scalable object-grounding pre-training stage further teaches the VLM to
predict language-conditioned heatmaps before robot-policy fine-tuning.
As a result, both pre-training and downstream manipulation are performed in
the same 2D visual-localization space, enabling data-efficient and
generalizable 3D action learning.

In this article, we extend \method into \memmethod by introducing a unified
spatio-temporal memory architecture.
The temporal memory maintains selected historical observations, allowing the policy to distinguish visually similar
situations occurring at different task stages and to determine
\emph{what to do next}.
The spatial memory preserves geometric information from an earlier,
less-occluded observation and re-renders the stored scene, recovering target regions that may be hidden by the robot or manipulated objects and helping the policy
determine \emph{where exactly to act}.
Such scene-level memory representation can also be shared across two arms,
enabling a natural extension to bimanual manipulation with a common backbone
and arm-specific action heads.

We evaluate \method and \memmethod on five simulation benchmarks.
The original \method achieves state-of-the-art performance on
RLBench \cite{james2020rlbench},
COLOSSEUM \cite{pumacay2024colosseum}, and
GemBench \cite{garcia2024towards}, demonstrating strong sample efficiency
and out-of-distribution generalization.
With the proposed memory architecture, \memmethod establishes
state-of-the-art results on two memory-dependent benchmarks,
RMBench \cite{chen2026rmbench} and
MemoryBench \cite{fang2025sam2act}.
\memmethod matches or improves upon \method on the original
benchmarks, showing that memory-dependent reasoning is gained without
sacrificing its original performance.

We further validate the framework on two real-world robot embodiments,
Franka Research 3 and Dobot CR5A, covering both memory-independent and memory-dependent tasks.
On memory-independent tasks, \method outperforms a strong baseline by
32\% on average and remains robust under visual perturbations, unseen object
categories, and unseen instructions.
On memory-dependent tasks, \memmethod improves the average success rate from
20.0\% to 93.3\% over \method.
These results demonstrate that the proposed extension has cross-embodiment scalability while preserving the
data efficiency and generalization ability of the original framework.

The main contributions of this article are summarized as follows:
\begin{itemize}
  \item We present \method, a data-efficient and generalizable 3D VLA
  framework that aligns VLM pre-training and 3D manipulation learning in a
  shared  2D heatmap space.

  \item We introduce a scalable language-conditioned heatmap pre-training
  strategy that transfers object-grounding knowledge to downstream robot
  action prediction.

  \item We propose \memmethod, a unified spatio-temporal memory architecture
  that combines temporal interaction history and persistent spatial
  information to determine both \emph{what to do next} and
  \emph{where exactly to act}.

  \item We conduct extensive experiments on five simulation benchmarks and
  two real-world robot embodiments, demonstrating state-of-the-art
  performance, strong data efficiency and generalization, bimanual manipulation, effective
  memory-dependent reasoning, and
  cross-embodiment scalability.
\end{itemize}

This article is an extension to our NeurIPS 2025 conference paper
\cite{li2026bridgevla}.
The major extensions are:
\begin{itemize}
  \item a unified spatio-temporal memory architecture that equips
  \method with explicit memory-dependent reasoning;

  \item evaluation on two additional memory-dependent benchmarks,
  RMBench and MemoryBench;

  \item an extension from single-arm to bimanual manipulation;

  \item new real-world experiments on different embodiments and tasks, together with
  additional analyses showing that the memory extension preserves the data
  efficiency and generalization ability of the original framework.
\end{itemize}

The remainder of this article is organized as follows.
Sec.~\ref{sec:related_work} reviews related work.
Sec.~\ref{sec:bridgevla} introduces the original \method framework,
and Sec.~\ref{sec:memmethod} presents the proposed \memmethod
architecture.
Sec.~\ref{sec:experiments} reports simulation and real-world
experiments together with ablation studies.
Finally, Sec.~\ref{sec:conclusion} concludes the article.
Implementation and evaluation details are provided in
Appendices~\ref{app:arch_details}--\ref{app:compute_resources}, and the
full experimental results in
Appendices~\ref{app:colosseum_results}--\ref{app:real_settings_vis}.
\section{Related Work}
\label{sec:related_work}

\subsection{Language-Conditioned Visuomotor Policies}

Most language-conditioned visuomotor policies employ transformers to process
2D visual inputs and directly generate 3D actions for manipulation
\cite{brohan2022rt,brohan2023rt,kimopenvla,black2024pi_0,
li2023vision}.
Among these approaches, developing large vision-language-action (VLA)
models, most often by leveraging pre-trained vision-language models (VLMs),
has become increasingly popular because of their effectiveness in learning
complex manipulation skills
\cite{brohan2023rt,kimopenvla,li2023vision,black2024pi_0,
pertsch2025fast,intelligence2025pi_06star,intelligence2026pi_07,
generalist2025gen0,generalist2026gen1,genesis2026gene265,
sunday2026act2preview}.
However, such 2D image-based policies typically require substantial
data-collection effort, often relying on large trajectory datasets to
generalize effectively across tasks~\cite{o2024open}.
In contrast, 3D manipulation policies have demonstrated strong potential for
data-efficient learning by exploiting the spatial structure inherent in 3D
observations.
One line of work directly processes point clouds
\cite{chen2023polarnet,yuan2023m2t2,yang2025fp3,gervet2023act3d,
3d-da,garcia2024towards}.
For example, Act3D~\cite{gervet2023act3d} constructs a 3D feature cloud by
lifting image features onto the observed point cloud and predicts
translational actions by classifying candidate 3D points in the workspace.
Another line of work represents the observation space using voxels and
predicts translational actions within the same voxel space, thereby aligning
input observations and output actions in a shared spatial representation
\cite{shridhar2023perceiver,james2022coarse}.
More recently, RVT~\cite{goyal2023rvt} and
RVT-2~\cite{goyal2024rvt} leverage orthographic projections of 3D point
clouds to convert 3D signals into 2D images, avoiding the high computational
cost of directly processing native 3D representations.
Unlike the above methods, our base framework, \method, seeks to unify the
semantic effectiveness of VLA models with the data efficiency of 3D
manipulation policies within a single cohesive framework.

\subsection{3D Vision-Language-Action (VLA) Models}

While 2D VLA models have been extensively studied, 3D VLA models
\cite{zhen20243d,yang2025fp3,li2025pointvla,
qu2025spatialvla} remain relatively underexplored.
Zhen \textit{et al.}~\cite{zhen20243d} build 3D-VLA on top of a 3D-based large
language model (LLM) and train it to perform 3D reasoning, multimodal goal
generation, and robot planning.
Lift3D~\cite{jia2024lift3d} enhances 2D foundation models
(\eg, DINOv2~\cite{oquab2023dinov2}) with implicit and explicit 3D robotic
representations for learning 3D manipulation policies.
FP3~\cite{yang2025fp3} employs a transformer to fuse information from point
clouds, proprioceptive states, and language instructions.
PointVLA~\cite{li2025pointvla} uses a VLM and a point-cloud encoder to process
2D images and 3D point clouds, respectively, and injects the resulting 3D
features additively into a few selected blocks of an otherwise frozen action
expert, thereby avoiding retraining of the pre-trained VLA.
SpatialVLA~\cite{qu2025spatialvla} introduces Ego3D positional encoding to
inject 3D information into 2D visual observations and adopts adaptive action
grids to represent robot motions in a more transferable manner.
In contrast to these architectural modifications, \method incorporates 3D
spatial priors without introducing a dedicated 3D encoder or modifying the
core architecture of the VLM: it projects 3D point clouds into multi-view
orthographic images~\cite{goyal2023rvt,goyal2024rvt} and formulates action
prediction as 2D spatial heatmap estimation, thereby keeping both inputs
and outputs within the native 2D domain of the pre-trained VLM.
A concurrent work, OG-VLA~\cite{singh2025ogvla}, explores a similar
orthographic-projection design, albeit generating heatmaps with an
auxiliary image-diffusion decoder.
However, these 3D VLA models operate without any form of memory, which
motivates the line of work on memory-dependent manipulation reviewed next.

\subsection{Memory-Dependent Manipulation}
Most of the above-mentioned VLA models and 3D manipulation policies,
including our \method, adopt a strictly Markovian formulation, predicting
actions solely from the current observation.
Such a formulation becomes inadequate when a task requires temporal
context, or when critical spatial geometry is occluded during execution.
On the one hand, to incorporate \emph{temporal context}, early approaches
attended to the entire observation history~\cite{guhur2023instruction},
resulting in computational costs that scaled poorly with episode length.
Recent methods instead maintain bounded or structured memory through
mechanisms such as explicit memory banks
(\eg, SAM2Act+~\cite{fang2025sam2act}), generative
world models~\cite{li2026lingbotva,yang2026wla,yang2026memorywam},
hierarchical planners~\cite{chen2026rmbench}, visual
traces~\cite{zheng2024tracevla}, or specialized cognitive or gated memory
modules~\cite{lei2025robomemory,shi2026memoryvla,gao2026gatedmemory}.
On the other hand, to address \emph{spatial occlusion}, existing solutions
typically track object poses explicitly over time~\cite{zhou2024yoso} or
maintain persistent geometric representations of the
workspace~\cite{zheng2026memworld}.
In contrast to the above methods, \memmethod offers a simple yet effective
solution to both challenges.
By jointly designing temporal memory for retaining historical context and
spatial memory for recovering occluded geometry, \memmethod achieves strong
performance on memory-dependent manipulation tasks using only a lightweight
attention module.
\section{\method}
\label{sec:bridgevla}

\begin{figure*}[t!]
  \centering
  \includegraphics[width=\textwidth]{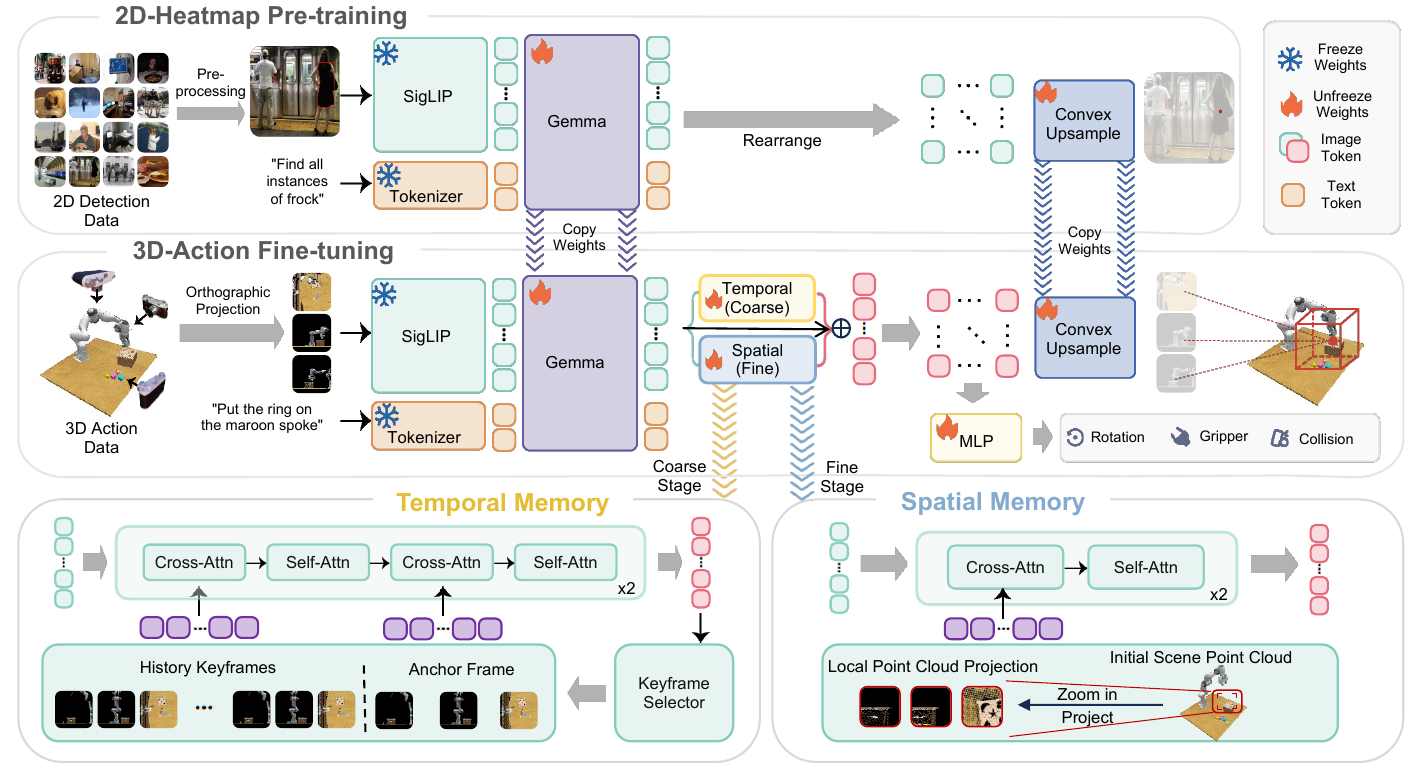}
  \caption{\textbf{Model Architecture.}
  \textbf{Top:} \method first learns language-conditioned 2D heatmap prediction
  from detection data and transfers the resulting weights to 3D action
  fine-tuning. During a policy forward pass, the observed point cloud is
  rendered into orthographic views and processed with the language instruction
  by the VLM to produce multi-view visual tokens and heatmaps. The coarse
  heatmaps localize a 3D waypoint, around which the point cloud is cropped,
  magnified, and re-rendered for a shared-weight fine pass. The fine heatmaps
  determine the final translation, while global tokens and local tokens at the
  coarse waypoint are fed to an MLP to predict rotation, gripper state, and
  collision avoidance. \textbf{Bottom:} \memmethod augments this forward flow
  with temporal and spatial memories. At the coarse stage, the
  current tokens cross-attend to an anchor frame and selected historical
  keyframes to determine \emph{what to do next}. At the fine stage, they
  cross-attend to view-aligned spatial tokens obtained by re-rendering the
  initial point cloud under the current zoom, providing less-occluded geometry
  to determine \emph{where exactly to act}.}
  \label{fig:bridgevla_arch}
\end{figure*}

The key idea of \method is to align both the input and output of 3D
manipulation learning within a shared 2D space.
Specifically, \method formulates 3D manipulation as multi-view 2D heatmap
prediction.
A scalable pre-training stage first learns language-conditioned heatmap
grounding from large-scale 2D data
(Sec.~\ref{sec:bridgevla:pretrain}).
During downstream policy fine-tuning, the observed 3D scene is rendered into
multiple orthographic views, and the predicted heatmaps are back-projected
to recover the 3D end-effector translation of the next keyframe
(Sec.~\ref{sec:bridgevla:finetune}).
Fig.~\ref{fig:bridgevla_arch} provides an overview of the complete
framework, including the spatio-temporal memory extension introduced in
Sec.~\ref{sec:memmethod}.

\subsection{Problem Formulation}
\label{sec:bridgevla:problem}

We consider language-conditioned multi-task 3D manipulation learned from a
set of expert demonstrations
$\mathcal{D}=\{\tau^i\}_{i=1}^{N}$.
Each demonstration is represented as
\begin{equation}
    \tau^i =
    \left(
        l^i,
        \left\{
            \left(\mathbf{o}_t^i,\mathbf{a}_t^i\right)
        \right\}_{t=1}^{H_i}
    \right),
\end{equation}
where $l^i$ is a language instruction,
$\mathbf{o}_t^i$ is the observation at step $t$, and
$\mathbf{a}_t^i$ is the corresponding expert action.
The observation $\mathbf{o}_t$ consists of one or more RGB-D images captured
by calibrated cameras.

Following prior keyframe-based manipulation methods
\cite{johns2021coarse,shridhar2023perceiver,goyal2023rvt},
the policy is queried at a sparse set of decision points and predicts the
end-effector configuration of the next keyframe.
For single-arm manipulation, the action is represented as
\begin{equation}
    \mathbf{a}_t =
    \left(
        \mathbf{x}_t,
        \mathbf{R}_t,
        g_t,
        c_t
    \right),
    \label{eq:action}
\end{equation}
where $\mathbf{x}_t\in\mathbb{R}^{3}$ is the target end-effector
translation, $\mathbf{R}_t\in SO(3)$ is the target rotation,
$g_t\in\{0,1\}$ denotes the gripper state, and
$c_t\in\{0,1\}$ is a collision-avoidance flag used by the motion planner.
The collision flag is omitted for benchmarks that do not provide this
action component.

The original \method learns a language-conditioned policy that predicts the
next best pose from the current observation:
\begin{equation}
    \mathbf{a}_t =
    \pi_{\mathrm{B}}
    \left(
        \mathbf{o}_t,l
    \right).
    \label{eq:bridgevla_policy}
\end{equation}
After each prediction, a motion planner or benchmark-specific low-level
controller executes the target action.
The observation is then refreshed, and the policy predicts the next
keyframe.
This process continues until task completion or a predefined step limit is
reached.

Equation~\eqref{eq:bridgevla_policy} defines the memory-free formulation of
\method, in which each action is predicted from the current observation
alone.
Sec.~\ref{sec:memmethod} extends this formulation by conditioning the
policy on information retained from earlier interactions.

\subsection{2D-Heatmap Pre-Training}
\label{sec:bridgevla:pretrain}

The original VLM backbone is pre-trained to generate token sequences, whose
outputs do not directly preserve the spatial structure required for precise
robot action prediction.
To align VLM pre-training with downstream policy learning, we introduce an
additional pre-training stage that teaches the model to ground
language-specified objects through 2D heatmap prediction.

We use the 120K object-detection split of
RoboPoint~\cite{yuan2024robopoint} as the pre-training dataset.
Each training sample consists of an image, a text prompt describing one or
more objects of interest, and the bounding boxes of the corresponding
objects.
For each target object $i$, we construct a spatially truncated Gaussian
probability map:
\begin{equation}
H_i^{\mathrm{gt}}(\mathbf{x})
=
\begin{cases}
p_i(\mathbf{x}),
&
p_i(\mathbf{x}) \geq p_{\min},
\\
0,
&
\text{otherwise},
\end{cases}
\label{eq:single_object_probability_map}
\end{equation}
where $\mathbf{x}=(u,v)$ denotes a pixel location and
\begin{equation}
p_i(\mathbf{x})
=
\exp
\left(
-\frac{
\left\|
\mathbf{x}-\widehat{\mathbf{x}}_i
\right\|_2^2
}{
2\sigma^2
}
\right).
\end{equation}
Here, $\widehat{\mathbf{x}}_i$ is the center of the bounding box of object
$i$, $\sigma$ controls the spatial extent of the Gaussian, and $p_{\min}$
is the truncation threshold.

When multiple target objects are specified in the same prompt, their
probability maps are averaged and normalized to construct the final
ground-truth heatmap:
\begin{equation}
H_{\mathrm{avg}}(\mathbf{x})
=
\frac{1}{N_{\mathrm{obj}}}
\sum_{i=1}^{N_{\mathrm{obj}}}
H_i^{\mathrm{gt}}(\mathbf{x}),
\end{equation}
\begin{equation}
H^{\mathrm{gt}}(\mathbf{x})
=
\frac{
H_{\mathrm{avg}}(\mathbf{x})
}{
\sum_{\mathbf{x}'\in\Omega}
H_{\mathrm{avg}}(\mathbf{x}')
},
\label{eq:multi_object_probability_map}
\end{equation}
where $N_{\mathrm{obj}}$ is the number of target objects and $\Omega$
denotes the image domain.
Examples of the resulting heatmap annotations are shown in
Fig.~\ref{fig:pretrain_dataset}.

As illustrated in Fig.~\ref{fig:bridgevla_arch}, the input image and the
text prompt describing the objects of interest are jointly processed by the
VLM backbone.
We employ PaliGemma~\cite{beyer2024paligemma}, which consists of a SigLIP
vision encoder~\cite{zhai2023sigmoid} and a Gemma transformer
backbone~\cite{team2024gemma}.
PaliGemma is originally pre-trained to take one or more images together with
a prefix text and autoregressively generate a suffix text.
Although causal attention is used for suffix-token generation, image tokens
and prefix-text tokens interact through bidirectional attention.
Consequently, each output image token is conditioned on both the visual
observation and the language query.

To recover spatial structure from these output tokens, we rearrange the
image tokens according to their original patch positions, forming a
two-dimensional feature grid.
A convex-upsampling module~\cite{teed2020raft} then decodes this grid into a
heatmap with the same spatial resolution as the input image.
Unlike fixed interpolation operations such as bilinear or nearest-neighbor
upsampling, convex upsampling predicts spatially varying interpolation
weights, allowing the decoder to recover finer localization details.

The model is optimized using the cross-entropy loss
\begin{equation}
L_{\mathrm{pre}}
=
-\sum_{\mathbf{x}\in\Omega}
H^{\mathrm{gt}}(\mathbf{x})
\log
\widehat{H}(\mathbf{x}),
\label{eq:pretrain_heatmap_loss}
\end{equation}
where $\widehat{H}$ is the predicted heatmap after spatial softmax
normalization.

This pre-training stage changes the output interface of the VLM from
unstructured token generation to language-conditioned spatial localization.
Unlike 3D VLA methods that represent robot actions as token
sequences~\cite{zhen20243d,qu2025spatialvla}, our model produces a
spatially structured 2D heatmap.
The formulation is also scalable because any vision-language dataset whose
annotations can be converted into spatial targets, such as object centers,
keypoints, or segmentation regions, can in principle be used for
pre-training.
The resulting VLM backbone and heatmap decoder are subsequently transferred
to 3D action fine-tuning.

\subsection{3D Action Fine-Tuning}
\label{sec:bridgevla:finetune}

During downstream policy learning, \method preserves the 2D input and
heatmap-output interface established during pre-training while using
explicit 3D geometry for robot action prediction.

Given RGB-D images captured by one or more calibrated cameras, we first
reconstruct a colored point cloud of the observed scene.
Following RVT~\cite{goyal2023rvt} and
RVT-2~\cite{goyal2024rvt}, the point cloud is rendered into three
orthographic projection images corresponding to the top, front, and right
views.
The three rendered images and the language instruction are then processed
by the pre-trained VLM backbone to predict one translational heatmap for
each view.

Notably, the VLM operates purely on images and language: no proprioceptive
signals, such as robot joint states or end-effector poses, are fed into its
forward pass.
This design preserves the image--language input format used during
pre-training and reduces the distribution shift between 2D heatmap
pre-training and 3D policy fine-tuning.

\paragraph{Translation prediction}
To recover the translational action, we uniformly sample candidate 3D
locations within the robot workspace.
Each candidate location is projected onto the three orthographic views, and
its score is obtained by aggregating the corresponding heatmap values:
\begin{equation}
s_t(\mathbf{x})
=
\sum_{v=1}^{V}
\widehat{H}_{t,v}
\left(
\Pi_v(\mathbf{x})
\right),
\qquad V=3,
\label{eq:translation_score}
\end{equation}
where $\Pi_v$ denotes the projection onto view $v$, and
$\widehat{H}_{t,v}$ is the predicted heatmap for that view.
The candidate with the highest score is selected as the end-effector
translation of the next keyframe:
\begin{equation}
\widehat{\mathbf{x}}_t
=
\arg\max_{\mathbf{x}}
s_t(\mathbf{x}).
\label{eq:translation_prediction}
\end{equation}
This procedure preserves the geometric correspondence between the
multi-view observations, heatmap outputs, and 3D translational actions.

\paragraph{Rotation, gripper, and collision prediction}
The non-translational action components are predicted jointly from
multi-view features that combine global scene context with local evidence
around the predicted translation.
For each orthographic view, we obtain a global feature by max-pooling all
output image tokens.
We then project the predicted 3D translation onto the view and take the
output image token at this 2D location as the local feature.
The global and local features of the three views are concatenated and passed
to a single three-layer MLP, whose output vector is split into the rotation,
gripper, and collision-avoidance predictions.
We take these features only from the coarse stage, described next.

The end-effector rotation is represented in the continuous 6D form
of~\cite{zhou2019continuity}, from which the rotation matrix is recovered by
Gram--Schmidt orthonormalization.
The gripper state and collision-avoidance flag are each predicted by a
two-class softmax over a pair of logits.

\paragraph{Coarse-to-fine refinement}
A prediction over the complete workspace provides global localization but
may lack the precision required for fine-grained manipulation.
Following prior work~\cite{james2022coarse,goyal2024rvt}, \method adopts a
coarse-to-fine refinement strategy.

The first forward pass predicts a coarse translation from orthographic views
covering the complete workspace.
The point cloud is then cropped and magnified using a cuboid centered at the
predicted coarse translation.
A second set of orthographic images is rendered from this zoomed local point
cloud and processed by the same VLM backbone.
The translation predicted by the fine pass is used as the final
end-effector position for execution.
The coarse and fine passes share model parameters and differ only in the
spatial range represented by their input projections.

\paragraph{Fine-tuning objective}
The fine-tuning objective contains four components:
\begin{equation}
L_{\mathrm{base}}
=
L_{\mathrm{trans}}
+
L_{\mathrm{rot}}
+
L_{\mathrm{gripper}}
+
L_{\mathrm{collision}}.
\label{eq:base_loss}
\end{equation}

The translation loss $L_{\mathrm{trans}}$ supervises heatmap prediction
using cross-entropy.
For each orthographic view, the ground-truth translational heatmap is
constructed using the normalized single-target probability map defined in
Eq.~\eqref{eq:single_object_probability_map}, where
$\widehat{\mathbf{x}}_i$ becomes the projected pixel location of the
ground-truth end-effector translation at the next keyframe.
The loss is applied to heatmaps predicted at both coarse and fine
stages.

The rotation loss $L_{\mathrm{rot}}$ is the squared Frobenius norm of the
difference between the rotation matrix recovered from the predicted 6D
representation and the ground-truth rotation matrix.
The gripper and collision terms are two-class cross-entropy losses.
For benchmarks that do not provide a collision-avoidance label,
$L_{\mathrm{collision}}$ is omitted.

To improve geometric robustness, random rigid-body transformations are
applied jointly to the input point cloud and ground-truth action during
training.
Additional implementation and optimization details are provided in
Appendix~\ref{app:finetune_hparams}.

The coarse-to-fine design not only improves spatial precision, but also
exposes two complementary stages at which information from earlier
observations can be incorporated.
The coarse stage reasons over the complete workspace and determines the next
target region; it could therefore benefit from temporal context
about previous interactions and completed sub-goals.
The fine stage performs precise localization within a zoomed local crop,
where task-relevant geometry may be occluded by the robot or manipulated
objects; it could therefore benefit from a persistent spatial reference.

Motivated by this stage-specific decomposition, Sec.~\ref{sec:memmethod}
introduces \memmethod, which augments the coarse-stage representation with
temporal memory and the fine-stage representation with spatial memory.
The resulting framework extends \method from a policy conditioned only on
the current observation to a memory-conditioned policy, while preserving
its original heatmap-based action interface, action parameterization, and
input--output alignment.

\section{\memmethod}
\label{sec:memmethod}

\subsection{Overview}
\label{sec:memmethod:formulation}

The preceding section introduces \method as a memory-free policy that
predicts each action from the current observation and language instruction.
Although this formulation is effective for many manipulation tasks, the
current observation alone may be insufficient when the policy must track
previously completed sub-goals or utilize task-relevant geometry that becomes occluded during execution.

To address these limitations, we extend \method to a memory-conditioned
policy:
\begin{equation}
    \mathbf{a}_t
    =
    \pi_{\mathrm{M}}
    \left(
        \mathbf{o}_t,
        l,
        \mathcal{M}_t
    \right),
    \label{eq:mem_policy}
\end{equation}
where $\mathcal{M}_t$ denotes the episode memory available at decision step
$t$.
We decompose the memory into two complementary components:
\begin{equation}
    \mathcal{M}_t
    =
    \left(
        \mathcal{T}_t,
        \mathcal{S}_t
    \right),
    \label{eq:memory_decomposition}
\end{equation}
where $\mathcal{T}_t$ is a temporal memory that summarizes the interaction
history, and $\mathcal{S}_t$ is a spatial memory that preserves previously
observed scene geometry.

The two memories complement the coarse-to-fine action prediction of
\method.
Temporal memory is incorporated at the coarse stage to help the policy
determine \emph{what to do next} from the execution history.
Spatial memory is incorporated at the fine stage to help the policy
determine \emph{where exactly to act} when the target geometry is partially
occluded.
Both memories are represented and processed in the visual token space of the
VLM, allowing them to be integrated without modifying the original
heatmap-based action interface.

\subsection{Temporal Memory for Coarse-Stage Reasoning}
\label{sec:memmethod:temporal}

The coarse stage determines the approximate target region and provides the
features used to predict rotation, gripper state, and collision avoidance.
Because these decisions may depend on both recent interactions and overall
task progress, we augment the coarse-stage representation with temporal
memory.
We formulate the temporal memory as
\begin{equation}
    \mathcal{T}_t
    =
    \left(
        \mathbf{A}_0,
        \mathcal{H}^{\mathrm{nbr}}_t,
        \mathcal{H}^{\mathrm{sub}}_t
    \right),
    \label{eq:temporal_memory}
\end{equation}
where $\mathbf{A}_0$ denotes the initial anchor views,
$\mathcal{H}^{\mathrm{nbr}}_t$ contains recent neighboring keyframes, and
$\mathcal{H}^{\mathrm{sub}}_t$ contains adaptively selected sub-goal
keyframes.
All three components are stored as coarse-stage visual tokens.
Together, they provide a fixed reference to the initial scene, short-term
context about recent transitions, and longer-term evidence of completed
sub-goals.

\subsubsection{Initial Anchor Views}
\label{sec:memmethod:anchor}

At the beginning of each episode, the initial point cloud is rendered into
the same three orthographic views used by the coarse stage.
The resulting visual tokens are stored as $\mathbf{A}_0$.
Comparing the current representation with $\mathbf{A}_0$ helps the policy
identify scene changes that have occurred since the beginning of the
episode.
Because the coarse-stage virtual cameras remain fixed throughout execution,
the anchor views provide a consistent global reference for reasoning about
task progress.

\subsubsection{History Keyframes}
\label{sec:memmethod:bank}

The initial anchor captures changes relative to the beginning of the episode,
but does not describe the sequence of interactions that produced the current
state.
We therefore maintain a dynamic history buffer containing neighboring and
sub-goal keyframes.
The neighboring-keyframe memory
$\mathcal{H}^{\mathrm{nbr}}_t$ stores the most recent $n$ executed
keyframes, with $n=2$ in all our experiments.
It provides short-term context about recent state transitions.
The sub-goal-keyframe memory
$\mathcal{H}^{\mathrm{sub}}_t$ stores representative historical
observations selected by the adaptive module described in
Sec.~\ref{sec:memmethod:gate}.
These observations record informative milestones and provide longer-term
evidence of completed sub-goals.
Both types of keyframes are cached as coarse-stage visual tokens.
Together, they represent immediate interaction context and longer-term task
progress.
Memory capacity, frame ordering, and buffer-management details are provided
in Appendix~\ref{app:finetune_hparams}.

\subsubsection{Adaptive Sub-Goal Keyframe Selection}
\label{sec:memmethod:gate}

Storing every historical keyframe would introduce redundancy.
We therefore introduce a lightweight adaptive selection module
$D_{\phi}$ to determine whether the current keyframe contains
informative evidence of task progress to be retained in
$\mathcal{H}^{\mathrm{sub}}_t$.
A learnable query token attends to the image tokens after temporal-memory
integration and summarizes them into a single vector, from which a small MLP
predicts a retention probability.
A keyframe is retained when its predicted probability exceeds a predefined
threshold.
The selection module operates on the memory-conditioned tokens
rather than the tokens before memory integration.
This allows it to assess whether the current observation provides information
that is not already represented in $\mathcal{T}_t$, thereby preserving
informative milestones while avoiding redundant observations.

\subsection{Spatial Memory for Occlusion-Robust Fine Localization}
\label{sec:memmethod:spatial}

\begin{figure}[t]
  \centering
  \includegraphics[width=\linewidth]{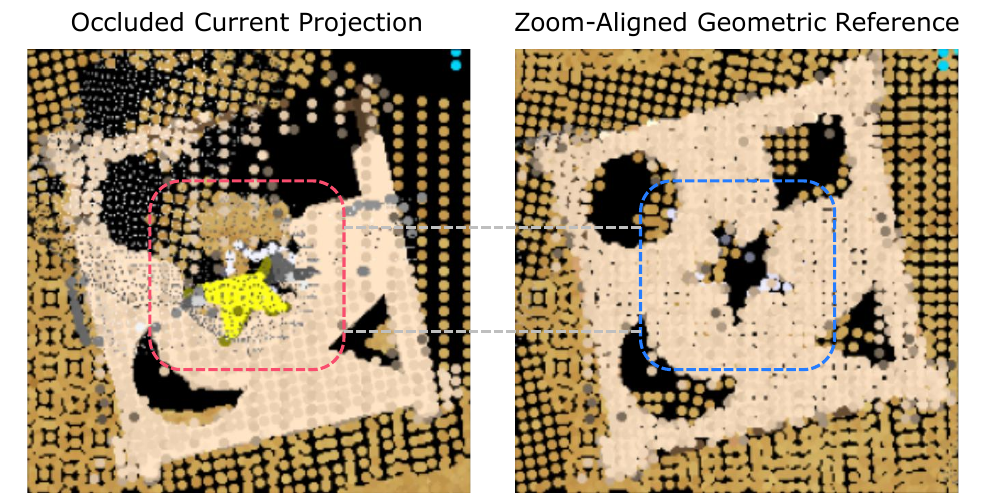}
  \caption{\textbf{Occlusion-robust fine localization using spatial
  memory.}
  \emph{Left:} the current zoomed observation, partially occluded by the
  gripper and manipulated object.
  \emph{Right:} the stored point cloud $\mathbf{P}_0$ re-rendered under
  the same predicted waypoint and zoom transformation, providing a
  spatially aligned, less-occluded reference of the same local region.}
  \label{fig:anchor_memory}
\end{figure}

Temporal memory helps the coarse stage identify the next target region, but
does not directly help recover the fine-grained geometry required for precise
localization.
At the fine stage, the robot arm, gripper, or manipulated object may partially
occlude the target region in the current local crop.
For example, Fig.~\ref{fig:anchor_memory} shows a case in which the gripper
and grasped object obscure the target receptacle.

To provide a persistent geometric reference, we store the colored point cloud
of the initial observation as $\mathbf{P}_0$.
Because this observation is captured before substantial robot interaction,
it typically provides a less-occluded view of the workspace.
Unlike a fixed 2D image, the stored point cloud can be re-rendered using the
same viewpoint and zoom configuration as any subsequent fine-stage crop.
Once the coarse stage predicts a waypoint
$\widehat{\mathbf{x}}^{\mathrm{c}}_t$, we apply the same zoom operation to
the current point cloud and the stored reference $\mathbf{P}_0$.
The zoomed reference is then rendered through the fine-stage virtual cameras
and encoded into visual tokens:
\begin{equation}
    \mathcal{S}_t
    =
    \Phi
    \left(
        \mathrm{Render}
        \left(
            \mathrm{Zoom}
            \left(
                \mathbf{P}_0;
                \widehat{\mathbf{x}}^{\mathrm{c}}_t
            \right)
        \right)
    \right),
    \label{eq:spatial_memory}
\end{equation}
where $\Phi$ denotes the VLM backbone applied to the re-rendered views
together with the language instruction, and $\mathcal{S}_t$ is the spatial
memory associated with the current fine-stage crop.

Because the current local observation and the re-rendered spatial reference
share the same virtual camera configuration, they are geometrically aligned
at the view level.
Accordingly, we let tokens from each current view attend only to memory tokens from
the corresponding reference view.
The current observation represents the latest state of the scene, whereas
the spatial memory provides previously visible geometry that may now be
occluded.
Thus, the spatial memory complements rather than replaces the current
observation.
This adaptive zoom alignment enables a single canonical point cloud
$\mathbf{P}_0$ to provide spatial references for different local regions
throughout the episode.
Because the required crop depends on the current coarse waypoint,
$\mathcal{S}_t$ is rendered and encoded on demand at every decision step.

\subsection{Memory Integration}
\label{sec:memmethod:integration}

After constructing the temporal and spatial memories, we inject them into
the corresponding stages of \method using compact attention modules.
Temporal memory conditions the coarse-stage representation, whereas spatial
memory conditions the fine-stage representation.

To avoid repeatedly processing historical projection images, the temporal
buffer stores their encoded visual tokens rather than the raw images.
Each cached observation is represented by a token grid in
$\mathbb{R}^{V\times N\times d}$, where $V$ is the number of orthographic
views, $N$ is the number of visual tokens per view, and $d$ is the token
dimension.
These are the VLM backbone's language-conditioned output image tokens, 
and can be reused throughout the episode without re-encoding.

Let $\mathbf{Z}^{\mathrm{c}}_t$ and
$\mathbf{Z}^{\mathrm{f}}_t$ denote the current coarse- and fine-stage visual
tokens, respectively.
We obtain the memory-conditioned representations as
\begin{equation}
    \widetilde{\mathbf{Z}}^{\mathrm{c}}_t
    =
    F_{\mathrm{temp}}
    \left(
        \mathbf{Z}^{\mathrm{c}}_t,
        \mathcal{T}_t
    \right),
    \qquad
    \widetilde{\mathbf{Z}}^{\mathrm{f}}_t
    =
    F_{\mathrm{spa}}
    \left(
        \mathbf{Z}^{\mathrm{f}}_t,
        \mathcal{S}_t
    \right),
    \label{eq:memory_integration}
\end{equation}
where $F_{\mathrm{temp}}$ and $F_{\mathrm{spa}}$ denote the temporal and
spatial memory-injection modules.
Each block consists of two attention layers.
Within each layer, the current visual tokens serve as queries, while the
corresponding memory tokens serve as keys and values in cross-attention.
The retrieved information is subsequently fused with the current
representation through self-attention and feed-forward updates.
The output preserves the shape of the input token grid:
\begin{equation}
    \widetilde{\mathbf{Z}}^{s}_t
    \in
    \mathbb{R}^{V\times N\times d},
    \qquad
    s\in\{\mathrm{c},\mathrm{f}\}.
    \label{eq:memory_output_shape}
\end{equation}
Consequently, the convex-upsampling modules and action-prediction heads of
\method can be applied without modification.
The temporal and spatial memory-injection modules introduce approximately
168M and 84M parameters, respectively, while the adaptive sub-goal-selection
module introduces approximately 18M parameters.
Despite this modest architectural overhead, the memory modules substantially
improve performance on memory-dependent tasks, as evaluated in
Sec.~\ref{sec:experiments}.

\subsection{Bimanual Extension}
\label{sec:memmethod:bimanual}

The temporal and spatial memories encode the shared episode state and
workspace geometry rather than arm-specific information.
This scene-level formulation allows \memmethod to extend naturally to
bimanual manipulation with only lightweight modifications.
Specifically, we introduce arm-specific action heads by duplicating the
convex-upsampling module and the MLP-based action heads, while sharing the VLM backbone, temporal memory, spatial memory, and adaptive selection module between the two arms.
At the coarse stage, the arm-specific heads operate on the shared
memory-conditioned representation and predict a separate coarse waypoint
for each arm.
At the fine stage, each arm independently constructs a zoomed local crop
around its predicted coarse waypoint and refines its final translation.
The resulting bimanual action is represented as
\begin{equation}
    \mathbf{a}^{\mathrm{bi}}_t
    =
    \left(
        \mathbf{a}^{\mathrm{left}}_t,
        \mathbf{a}^{\mathrm{right}}_t
    \right),
    \label{eq:bimanual_action}
\end{equation}
where each arm-specific action follows the representation defined in
Eq.~\eqref{eq:action}.
This shared-trunk, arm-specific-head design successfully supports bimanual action prediction without duplicating the computationally expensive VLM backbone or the episodic memory modules.

\subsection{Training and Inference Details}
\label{sec:memmethod:training}

\subsubsection{Training}

During training, the temporal and spatial memories associated with each
sample are constructed from preceding observations in the corresponding
expert demonstration.
The initial observation provides the temporal anchor views and the spatial memory reference, while neighboring and annotated sub-goal keyframes are
selected from earlier execution steps.

To preserve geometric consistency, the random rigid-body augmentation
introduced in Sec.~\ref{sec:bridgevla:finetune} is applied consistently to
the current observation, the associated memory observations, and the
ground-truth actions.
The complete training objective is
\begin{equation}
    L
    =
    L_{\mathrm{base}}
    +
    \lambda_{\mathrm{check}}
    L_{\mathrm{check}},
    \label{eq:memory_training_loss}
\end{equation}
where $L_{\mathrm{base}}$ is the action-prediction loss defined in
Eq.~\eqref{eq:base_loss}, and $L_{\mathrm{check}}$ is a binary
cross-entropy loss that supervises whether the current keyframe should be retained as a sub-goal keyframe.
For bimanual tasks, $L_{\mathrm{base}}$ includes the action losses of both
arms, whereas the shared adaptive selection module is supervised once using
$L_{\mathrm{check}}$.

\subsubsection{Inference}

At the beginning of an episode, the initial observation is used to construct
the temporal anchor views and the spatial point-cloud reference
$\mathbf{P}_0$.
The remaining temporal-memory slots are initialized with zero padding.
After each executed action, the encoded image tokens of the current
observation are inserted into the temporal buffer as a neighboring
keyframe.
The adaptive selection module also determines whether the observation should
be retained as a sub-goal keyframe.
When the buffer exceeds its predefined capacity, the oldest entries are
removed according to the memory-management strategy described in
Appendix~\ref{app:finetune_hparams}.
Because temporal memory stores encoded image tokens rather than raw
projection images, historical observations do not require repeated visual encoding.
For spatial memory, we retain the colored point cloud of the initial observation and re-render and re-encode it at each decision step, since the required local crop depends on the dynamically predicted coarse waypoint. As only a single reference observation is processed, the resulting computational overhead remains low.
Overall, the memory extension introduces 269.77M additional parameters,
corresponding to a 9.2\% increase over the 2.92B-parameter backbone.
For the inference latency, \method\ takes 0.35 seconds per prediction step and \memmethod\ 0.57 seconds on a single NVIDIA RTX 4090 GPU, a gap that is minor compared with the observation transmission and motion execution that dominate each keyframe-based control step.
\section{Experiments}
\label{sec:experiments}

In this section, we conduct extensive evaluations in both simulation and
real-world environments to assess \method and its memory-augmented extension,
\memmethod.
Specifically, our experiments are designed to answer the following research
questions:
\begin{itemize}
    \item[Q1:] How effectively do \method and \memmethod learn 3D manipulation
    compared with state-of-the-art methods when sufficient demonstrations are
    available?

    \item[Q2:] How robust are \method and \memmethod under
    out-of-distribution conditions, including distractors, lighting changes,
    background variations, novel object--skill combinations, and unseen
    object categories?

    \item[Q3:] How important are the proposed architectural components,
    including heatmap-based action decoding, 2D heatmap pre-training, and
    unified spatio-temporal memory, to the overall performance?

    \item[Q4:] Can \method and \memmethod be deployed effectively across
    different real-world robot platforms while retaining high sample
    efficiency, such as learning each task from only 10 demonstrations?

    \item[Q5:] How effectively can \memmethod address memory-dependent manipulation tasks?
\end{itemize}

\subsection{RLBench: General 3D Manipulation}
\label{sec:exp:bridgevla}
\label{sec:exp:rlbench}

To evaluate the base model's capacity for general 3D manipulation, we primarily evaluate \method\ and \memmethod\ on the RLBench benchmark.

\begin{table*}[!t]
\centering
\caption{\textbf{Results and ablation studies on RLBench.}
Success rates (SR, \%) across 18 tasks, together with the average SR and
average rank (lower is better).
The upper rows report prior methods, followed by \method
(the memory-free base policy, also referred to as \textbf{Base}) and
\memmethod; the indented rows below each report its architectural and
memory ablations.
Results are presented as mean$_{\pm\mathrm{std}}$ over five random seeds,
with 25 evaluation episodes per seed.
\textbf{w/o}~$\mathcal{S}$ and \textbf{w/o}~$\mathcal{T}$ denote
\memmethod without spatial and temporal memory, respectively.
Evaluation protocols and training details are provided in
Appendices~\ref{app:finetune_hparams} and~\ref{app:eval_protocol}.
The best result in each column is highlighted in bold.}
\label{tab:rlbench}
\scriptsize
\setlength{\tabcolsep}{3pt}
\setlength{\aboverulesep}{0pt}
\setlength{\belowrulesep}{0pt}
\renewcommand{\arraystretch}{1.15}
\resizebox{\fitwidth}{!}{%
\begin{tabular}{l *{10}{c}}
\toprule
& \textbf{Avg.} & \textbf{Avg.} & \textbf{Close} & \textbf{Drag} & \textbf{Insert}
& \textbf{Meat off} & \textbf{Open} & \textbf{Place} & \textbf{Place}
& \textbf{Push} \\
\multirow{-2}{*}{\textbf{Method}}
& \textbf{SR (\%) $\uparrow$} & \textbf{Rank $\downarrow$} & \textbf{Jar} & \textbf{Stick} & \textbf{Peg}
& \textbf{Grill} & \textbf{Drawer} & \textbf{Cups} & \textbf{Wine}
& \textbf{Buttons} \\
\midrule
PerAct~\cite{shridhar2023perceiver} & 49.4 & 11.33 & 55.2$_{\pm 4.7}$ & 89.6$_{\pm 4.1}$ & 5.6$_{\pm 4.1}$ & 70.4$_{\pm 2.0}$ & 88.0$_{\pm 5.7}$ & 2.4$_{\pm 3.2}$ & 44.8$_{\pm 7.8}$ & 92.8$_{\pm 3.0}$ \\
Act3D~\cite{gervet2023act3d} & 65.0 & 9.17 & 92.0 & 92.0 & 27.0 & 94.0 & 93.0 & 3.0 & 80.0 & 99.0 \\
RVT~\cite{goyal2023rvt} & 62.9 & 9.08 & 52.0$_{\pm 2.5}$ & 99.2$_{\pm 1.6}$ & 11.2$_{\pm 3.0}$ & 88.0$_{\pm 2.5}$ & 71.2$_{\pm 6.9}$ & 4.0$_{\pm 2.5}$ & 91.0$_{\pm 5.2}$ & \textbf{100.0}$_{\pm 0.0}$ \\
3D Diffuser Actor~\cite{3d-da} & 81.3 & 6.19 & 96.0$_{\pm 2.5}$ & \textbf{100.0}$_{\pm 0.0}$ & 65.6$_{\pm 4.1}$ & 96.8$_{\pm 1.6}$ & 89.6$_{\pm 4.1}$ & 24.0$_{\pm 7.6}$ & 93.6$_{\pm 4.8}$ & 98.4$_{\pm 2.0}$ \\
RVT-2~\cite{goyal2024rvt} & 81.4 & 5.97 & \textbf{100.0}$_{\pm 0.0}$ & 99.0$_{\pm 1.7}$ & 40.0$_{\pm 0.0}$ & 99.0$_{\pm 1.7}$ & 74.0$_{\pm 11.8}$ & 38.0$_{\pm 4.5}$ & 95.0$_{\pm 3.3}$ & \textbf{100.0}$_{\pm 0.0}$ \\
SAM2Act~\cite{fang2025sam2act} & 86.8$_{\pm 0.5}$ & 5.47 & 99.0$_{\pm 2.0}$ & 99.0$_{\pm 2.0}$ & 84.0$_{\pm 5.7}$ & 98.0$_{\pm 2.3}$ & 83.0$_{\pm 6.0}$ & 47.0$_{\pm 6.0}$ & 93.0$_{\pm 3.8}$ & \textbf{100.0}$_{\pm 0.0}$ \\
\midrule
\textbf{\method\ (ours)} & 90.5$_{\pm 1.1}$ & 4.75 & \textbf{100.0}$_{\pm 0.0}$ & 97.6$_{\pm 3.6}$ & 91.2$_{\pm 1.8}$ & \textbf{100.0}$_{\pm 0.0}$ & 99.2$_{\pm 1.8}$ & 58.4$_{\pm 4.6}$ & 89.6$_{\pm 8.3}$ & \textbf{100.0}$_{\pm 0.0}$ \\
\tbranch w/ discretized rotation & 88.2 & 4.86 & \textbf{100.0}$_{\pm 0.0}$ & \textbf{100.0}$_{\pm 0.0}$ & 88.0$_{\pm 2.8}$ & \textbf{100.0}$_{\pm 0.0}$ & \textbf{100.0}$_{\pm 0.0}$ & 58.4$_{\pm 10.0}$ & 88.0$_{\pm 2.8}$ & 98.4$_{\pm 2.2}$ \\
\tbranch w/o heatmap decoding & 31.4 & 12.78 & 49.3$_{\pm 2.3}$ & 65.3$_{\pm 2.3}$ & 0.0$_{\pm 0.0}$ & 81.3$_{\pm 4.6}$ & 74.7$_{\pm 10.1}$ & 1.3$_{\pm 2.3}$ & 32.0$_{\pm 14.4}$ & 54.7$_{\pm 6.1}$ \\
\tlast w/ 3D position input & 56.2 & 10.14 & 96.0$_{\pm 0.0}$ & 58.7$_{\pm 6.1}$ & 26.7$_{\pm 2.3}$ & 96.0$_{\pm 0.0}$ & 97.3$_{\pm 2.3}$ & 14.7$_{\pm 4.6}$ & 81.3$_{\pm 8.3}$ & 86.7$_{\pm 2.3}$ \\
\specialrule{1pt}{0pt}{0pt}
\textbf{\memmethod\ (ours)} & \textbf{93.7$_{\pm 0.6}$} & \textbf{3.64} & \textbf{100.0}$_{\pm 0.0}$ & 98.4$_{\pm 2.2}$ & \textbf{99.2}$_{\pm 1.8}$ & \textbf{100.0}$_{\pm 0.0}$ & 99.2$_{\pm 1.8}$ & \textbf{76.8}$_{\pm 11.5}$ & \textbf{95.2}$_{\pm 4.4}$ & \textbf{100.0}$_{\pm 0.0}$ \\
\tbranch w/o spatial memory & 92.0$_{\pm 0.5}$ & 3.81 & \textbf{100.0}$_{\pm 0.0}$ & \textbf{100.0}$_{\pm 0.0}$ & 95.2$_{\pm 3.3}$ & \textbf{100.0}$_{\pm 0.0}$ & 99.2$_{\pm 1.8}$ & 74.4$_{\pm 3.6}$ & 78.4$_{\pm 3.6}$ & \textbf{100.0}$_{\pm 0.0}$ \\
\tlast w/o temporal memory & 91.9$_{\pm 1.0}$ & 3.81 & \textbf{100.0}$_{\pm 0.0}$ & 99.2$_{\pm 1.8}$ & 82.4$_{\pm 2.2}$ & \textbf{100.0}$_{\pm 0.0}$ & 97.6$_{\pm 2.2}$ & 57.6$_{\pm 7.3}$ & 90.4$_{\pm 4.6}$ & \textbf{100.0}$_{\pm 0.0}$ \\
\bottomrule
\addlinespace[5pt]
\toprule
& \textbf{Put in} & \textbf{Put in} & \textbf{Put in} & \textbf{Screw}
& \textbf{Slide} & \textbf{Sort} & \textbf{Stack} & \textbf{Stack}
& \textbf{Sweep to} & \textbf{Turn} \\
\multirow{-2}{*}{\textbf{Method}}
& \textbf{Cupboard} & \textbf{Drawer} & \textbf{Safe} & \textbf{Bulb}
& \textbf{Block} & \textbf{Shape} & \textbf{Blocks} & \textbf{Cups}
& \textbf{Dustpan} & \textbf{Tap} \\
\midrule
PerAct~\cite{shridhar2023perceiver} & 28.0$_{\pm 4.4}$ & 51.2$_{\pm 4.7}$ & 84.0$_{\pm 3.6}$ & 17.6$_{\pm 2.0}$ & 74.0$_{\pm 13.0}$ & 16.8$_{\pm 4.7}$ & 26.4$_{\pm 3.2}$ & 2.4$_{\pm 2.0}$ & 52.0$_{\pm 0.0}$ & 88.0$_{\pm 4.4}$ \\
Act3D~\cite{gervet2023act3d} & 51.0 & 90.0 & 95.0 & 47.0 & 93.0 & 8.0 & 12.0 & 9.0 & 92.0 & 94.0 \\
RVT~\cite{goyal2023rvt} & 49.6$_{\pm 3.2}$ & 88.0$_{\pm 5.7}$ & 91.2$_{\pm 3.0}$ & 48.0$_{\pm 5.7}$ & 81.6$_{\pm 5.4}$ & 36.0$_{\pm 2.5}$ & 28.8$_{\pm 3.9}$ & 26.4$_{\pm 8.2}$ & 72.0$_{\pm 0.0}$ & 93.6$_{\pm 4.1}$ \\
3D Diffuser Actor~\cite{3d-da} & 85.6$_{\pm 4.1}$ & 96.0$_{\pm 3.6}$ & 97.6$_{\pm 2.0}$ & 82.4$_{\pm 2.0}$ & 97.6$_{\pm 3.2}$ & 44.0$_{\pm 4.4}$ & 68.3$_{\pm 3.3}$ & 47.2$_{\pm 8.5}$ & 84.0$_{\pm 4.4}$ & \textbf{99.2}$_{\pm 1.6}$ \\
RVT-2~\cite{goyal2024rvt} & 66.0$_{\pm 4.5}$ & 96.0$_{\pm 0.0}$ & 96.0$_{\pm 2.8}$ & 88.0$_{\pm 4.9}$ & 92.0$_{\pm 2.8}$ & 35.0$_{\pm 7.1}$ & 80.0$_{\pm 2.8}$ & 69.0$_{\pm 5.9}$ & \textbf{100.0}$_{\pm 0.0}$ & 99.0$_{\pm 1.7}$ \\
SAM2Act~\cite{fang2025sam2act} & 75.0$_{\pm 3.8}$ & 99.0$_{\pm 2.0}$ & 98.0$_{\pm 2.3}$ & 89.0$_{\pm 2.0}$ & 86.0$_{\pm 4.0}$ & 64.0$_{\pm 4.6}$ & 76.0$_{\pm 8.6}$ & 78.0$_{\pm 4.0}$ & 99.0$_{\pm 2.0}$ & 96.0$_{\pm 5.7}$ \\
\midrule
\textbf{\method\ (ours)} & 91.2$_{\pm 1.8}$ & 96.0$_{\pm 0.0}$ & 95.2$_{\pm 3.3}$ & 93.6$_{\pm 6.1}$ & 95.2$_{\pm 3.3}$ & 55.2$_{\pm 5.2}$ & 84.8$_{\pm 9.1}$ & 88.8$_{\pm 3.3}$ & \textbf{100.0}$_{\pm 0.0}$ & 92.8$_{\pm 3.3}$ \\
\tbranch w/ discretized rotation & 73.6$_{\pm 4.6}$ & \textbf{99.2}$_{\pm 1.8}$ & \textbf{99.2}$_{\pm 1.8}$ & 87.2$_{\pm 6.6}$ & 96.0$_{\pm 2.8}$ & 60.8$_{\pm 7.7}$ & 76.8$_{\pm 8.7}$ & 81.6$_{\pm 3.6}$ & 87.2$_{\pm 1.8}$ & 92.8$_{\pm 3.3}$ \\
\tbranch w/o heatmap decoding & 5.3$_{\pm 2.3}$ & 0.0$_{\pm 0.0}$ & 58.7$_{\pm 22.7}$ & 2.7$_{\pm 2.3}$ & 64.0$_{\pm 0.0}$ & 4.0$_{\pm 4.0}$ & 0.0$_{\pm 0.0}$ & 0.0$_{\pm 0.0}$ & 32.0$_{\pm 4.0}$ & 40.0$_{\pm 10.6}$ \\
\tlast w/ 3D position input & 10.7$_{\pm 2.3}$ & 78.7$_{\pm 2.3}$ & 97.3$_{\pm 4.6}$ & 16.0$_{\pm 4.0}$ & 72.0$_{\pm 0.0}$ & 21.3$_{\pm 8.3}$ & 17.3$_{\pm 2.3}$ & 4.0$_{\pm 4.0}$ & 53.3$_{\pm 2.3}$ & 84.0$_{\pm 0.0}$ \\
\specialrule{1pt}{0pt}{0pt}
\textbf{\memmethod\ (ours)} & \textbf{92.0}$_{\pm 0.0}$ & \textbf{99.2}$_{\pm 1.8}$ & 92.8$_{\pm 5.9}$ & 95.2$_{\pm 5.2}$ & 96.0$_{\pm 4.0}$ & 72.0$_{\pm 6.3}$ & 85.6$_{\pm 4.6}$ & \textbf{98.4}$_{\pm 2.2}$ & 97.6$_{\pm 2.2}$ & 89.6$_{\pm 4.6}$ \\
\tbranch w/o spatial memory & 90.4$_{\pm 2.2}$ & 91.2$_{\pm 1.8}$ & 96.0$_{\pm 4.9}$ & 95.2$_{\pm 1.8}$ & 97.6$_{\pm 3.6}$ & 60.8$_{\pm 3.3}$ & \textbf{91.2}$_{\pm 3.3}$ & 92.8$_{\pm 4.4}$ & 99.2$_{\pm 1.8}$ & 95.2$_{\pm 4.4}$ \\
\tlast w/o temporal memory & 88.8$_{\pm 4.4}$ & 98.4$_{\pm 3.6}$ & 92.8$_{\pm 3.3}$ & \textbf{96.0}$_{\pm 4.0}$ & \textbf{100.0}$_{\pm 0.0}$ & \textbf{73.6}$_{\pm 5.4}$ & 88.8$_{\pm 3.3}$ & 93.6$_{\pm 2.2}$ & \textbf{100.0}$_{\pm 0.0}$ & 94.4$_{\pm 4.6}$ \\
\bottomrule
\end{tabular}%
}
\end{table*}

\paragraph{Setup} 
RLBench~\cite{james2020rlbench} serves as a standard multi-task suite for evaluating manipulation policies.
It implements tasks in CoppeliaSim~\cite{rohmer2013v} using a Franka Panda robot equipped with a parallel-jaw gripper, with observations provided by four RGB-D cameras (front, left shoulder, right shoulder, and wrist). 
Following previous works~\cite{shridhar2023perceiver,goyal2023rvt,goyal2024rvt}, we evaluate on 18 tasks spanning non-prehensile manipulation such as \textit{Slide Block to Target}, pick-and-place tasks like \textit{Stack Cups}, and high-precision insertion tasks including \textit{Sort Shape}. 
We train on 100 demonstrations per task and report the mean success rate over five evaluation runs of 25 episodes per task.

\paragraph{Baselines}
We compare \method\ with state-of-the-art baselines encompassing both 2D and 3D methods.
(1) \textbf{PerAct}~\cite{shridhar2023perceiver} operates in the voxel space and predicts the action with a Perceiver transformer~\cite{jaegleperceiver}.
(2) \textbf{Act3D}~\cite{gervet2023act3d} predicts the next keyframe action by selecting the point with the highest score from a set of randomly sampled points in the workspace.
(3) \textbf{RVT}~\cite{goyal2023rvt} uses a multi-view transformer to aggregate information from multiple orthographic views of the point cloud observation.
(4) \textbf{3D Diffuser Actor}~\cite{3d-da} generates 3D trajectories via a diffusion process conditioned on the 3D observation and language instruction.
(5) \textbf{RVT-2}~\cite{goyal2024rvt} further improves the precision of its prior via a coarse-to-fine strategy.
(6) \textbf{SAM2Act}~\cite{fang2025sam2act}, the previous state-of-the-art method on this benchmark, builds upon the multi-view transformer and integrates the SAM2 visual foundation model to strengthen scene representation.

\paragraph{Results} 
Table~\ref{tab:rlbench} summarizes the performance comparison. \method\ achieves a 90.5\% average success rate across the 18 tasks, outperforming SAM2Act by 3.7 absolute percentage points, establishing a new state-of-the-art and addressing Q1.
This improvement is particularly pronounced in precision-critical tasks such as \emph{Stack Cups}, underscoring the efficacy of our dense per-view heatmap representation for fine-grained localization. The high success rates showcase its strong capability in precise manipulation. 
The primary failure modes emerge in occlusion-heavy tasks, where the robot's arm obscures the target during the fine-localization stage.
These occlusions are later resolved by \memmethod's spatio-temporal memory, which elevates the overall success rate to 93.7\% and brings significant gains to \emph{Sort Shape} (+16.8\%) and \emph{Place Cups} (+18.4\%).
The indented rows of Table~\ref{tab:rlbench} report ablated variants of both models, which we analyze in Sec.~\ref{sec:exp:ablation}.
\begin{table*}[!t]
\centering
\caption{\textbf{Results on RMBench.}
Success rates (\%) over 100 episodes per task on the nine dual-arm
RMBench tasks~\cite{chen2026rmbench}, grouped by task memory
complexity.
Baseline numbers are quoted
from~\cite{chen2026rmbench,yang2026memorywam,yang2026wla}, with group
averages recomputed from per-task results where a source reports none.
Best result per task in bold; ``--'' denotes tasks not evaluated by the
source.}
\label{tab:rmbench_main}
\footnotesize
\setlength{\tabcolsep}{4pt}
\setlength{\aboverulesep}{0pt}
\setlength{\belowrulesep}{0pt}
\renewcommand{\arraystretch}{1.25}
\begin{tabular}{l c cccccc ccccc}
\toprule
& & \multicolumn{6}{c}{\textbf{$M(1)$ tasks}}
& \multicolumn{5}{c}{\textbf{$M(n)$ tasks}} \\
\cmidrule(lr){3-8} \cmidrule(lr){9-13}
& \textbf{Overall}
& \textbf{Observe \&} & \textbf{Rearrange} & \textbf{Put Back} & \textbf{Swap} & \textbf{Swap} &
& \textbf{Battery} & \textbf{Blocks} & \textbf{Cover} & \textbf{Press} & \\
\multirow{-2}{*}{\textbf{Method}}
& \textbf{Avg.}
& \textbf{Pick Up} & \textbf{Blocks} & \textbf{Block} & \textbf{Blocks} & \textbf{T} & \multirow{-2}{*}{\textbf{\textit{Avg.}}}
& \textbf{Try} & \textbf{Ranking Try} & \textbf{Blocks} & \textbf{Button} & \multirow{-2}{*}{\textbf{\textit{Avg.}}} \\
\midrule
DP~\cite{chi2024diffusionpolicy}        & 5.8  & 1  & 0   & 0   & 11  & 20 & 6.4  & 10 & 10  & 0  & 0  & 5.0  \\
ACT~\cite{zhao2023learning}             & 5.9  & 1  & 29  & 0   & 2   & 2  & 6.8  & 19 & 0   & 0  & 0  & 4.8  \\
$\pi_{0.5}$~\cite{intelligence2025pi_}  & 10.4 & 9  & 13  & 11  & 24  & 15 & 14.4 & 16 & 6   & 0  & 0  & 5.5  \\
X-VLA~\cite{zheng2025xvla}              & 9.8  & 9  & 13  & 18  & 16  & 3  & 11.8 & 26 & 1   & 2  & 0  & 7.3  \\
Mem-0~\cite{chen2026rmbench}            & 42.0 & 4  & 89  & 90  & 67  & 14 & 52.8 & 28 & 18  & 68 & 0  & 28.5 \\
Fast-WAM~\cite{yuan2026fastwam}         & 5.9  & 0  & 0   & 0   & 0   & 7  & 1.4  & 20 & 26  & 0  & 0  & 11.5 \\
LingBot-VA~\cite{li2026lingbotva}       & 78.2 & 13 & \textbf{100} & \textbf{100} & 99  & 88 & 80.0 & 41 & \textbf{100} & 79 & 84 & 76.0 \\
MemoryWAM~\cite{yang2026memorywam}      & 83.0 & 27 & \textbf{100} & \textbf{100} & \textbf{100} & 94 & 84.2 & 41 & \textbf{100} & 98 & 87 & 81.5 \\
\midrule
\textbf{\method\ (ours)}                & 18.9 & 75 & 0   & 1   & 11 & 8  & 19.0 & 72 & 0   & 3  & 0  & 18.8 \\
\textbf{\memmethod\ (ours)}             & \textbf{96.0} & \textbf{81} & \textbf{100} & \textbf{100} & 99 & \textbf{96} & \textbf{95.2} & \textbf{96} & \textbf{100} & \textbf{99} & \textbf{93} & \textbf{97.0} \\
\bottomrule
\end{tabular}
\end{table*}

\begin{table*}[!t]
\centering
\caption{\textbf{Results on COLOSSEUM.}
Success rates (\%) across the 14 COLOSSEUM evaluation
settings~\cite{pumacay2024colosseum}: the 12 individual perturbation
axes, the original RLBench variations (``RLBench''), and all
perturbations applied jointly (``All Perturb.'').
MO and RO denote perturbations of the manipulated object and the
receptacle object, respectively.
``Avg.\ Rank'' is the average rank across the 14 settings over all
listed methods (lower is better).
R3M-MLP, MVP-MLP, PerAct, and RVT are quoted
from~\cite{pumacay2024colosseum}; RVT-2, \method, and \memmethod\ were
trained and evaluated by us (mean$\pm$variance over three test
repetitions; Appendix~\ref{app:eval_protocol}).
Best result per column in bold.}
\label{tab:colosseum}
\footnotesize
\setlength{\tabcolsep}{4pt}
\setlength{\aboverulesep}{0pt}
\setlength{\belowrulesep}{0pt}
\renewcommand{\arraystretch}{1.25}
\resizebox{\fitwidth}{!}{%
\begin{tabular}{l*{8}{c}}
\toprule
& \textbf{Avg.} & \textbf{Avg.} & \textbf{All} & \textbf{MO} & \textbf{RO} & \textbf{MO} & \textbf{RO} & \textbf{MO} \\
\multirow{-2}{*}{\textbf{Method}}
& \textbf{SR (\%) $\uparrow$} & \textbf{Rank $\downarrow$} & \textbf{Perturb.} & \textbf{Color} & \textbf{Color} & \textbf{Texture} & \textbf{Texture} & \textbf{Size} \\
\midrule
R3M-MLP~\cite{nair2022r3m}          & 0.8  & 6.71 & 0.6  & 0.4  & 0.0  & 0.0  & 0.0   & 1.8 \\
MVP-MLP~\cite{xiao2022masked}       & 1.6  & 6.00 & 0.8  & 1.2  & 0.0  & 0.4  & 0.0   & 4.44 \\
PerAct~\cite{shridhar2023perceiver} & 27.9 & 4.71 & 7.2  & 24.0 & 29.2 & 28.8 & 17.71 & 35.6 \\
RVT~\cite{goyal2023rvt}             & 35.4 & 4.29 & 6.4  & 26.0 & 31.3 & 44.8 & 41.1  & 35.3 \\
RVT-2~\cite{goyal2024rvt}           & 56.7 & 2.86 & 15.6$_{\pm 0.8}$ & 53.0$_{\pm 0.9}$ & 54.6$_{\pm 0.6}$ & 59.7$_{\pm 0.7}$ & 56.7$_{\pm 1.4}$ & 60.9$_{\pm 0.9}$ \\
\midrule
\textbf{\method (ours)} & 64.0 & \textbf{1.50} & 18.7$_{\pm 2.2}$ & 60.5$_{\pm 1.1}$ & \textbf{63.8}$_{\pm 0.1}$ & 63.5$_{\pm 1.5}$ & \textbf{68.4}$_{\pm 3.3}$ & 69.3$_{\pm 1.0}$ \\
\textbf{\memmethod (ours)} & \textbf{65.2} & 1.64 & \textbf{38.9}$_{\pm 0.8}$ & \textbf{68.7}$_{\pm 0.7}$ & 62.7$_{\pm 0.6}$ & \textbf{65.7}$_{\pm 0.4}$ & 65.5$_{\pm 1.2}$ & \textbf{71.5}$_{\pm 0.3}$ \\
\bottomrule
\addlinespace[5pt]
& \textbf{RO} & \textbf{Light} & \textbf{Table} & \textbf{Table} & & \textbf{Background} & & \textbf{Camera} \\
\multirow{-2}{*}{\textbf{Method}}
& \textbf{Size} & \textbf{Color} & \textbf{Color} & \textbf{Texture} & \multirow{-2}{*}{\textbf{Distractor}} & \textbf{Texture} & \multirow{-2}{*}{\textbf{RLBench}} & \textbf{Pose} \\
\midrule
R3M-MLP~\cite{nair2022r3m}          & 0.0  & 1.0  & 1.4  & 0.2  & 1.6  & 1.2  & 2.0  & 0.8 \\
MVP-MLP~\cite{xiao2022masked}       & 0.0  & 1.6  & 1.6  & 1.0  & 3.8  & 2.2  & 2.0  & 2.6 \\
PerAct~\cite{shridhar2023perceiver} & 29.3 & 29.1 & 30.4 & 23.2 & 27.1 & 33.5 & 39.4 & 36.3 \\
RVT~\cite{goyal2023rvt}             & 40.5 & 34.0 & 30.0 & 45.2 & 18.8 & 46.4 & 53.4 & 42.2 \\
RVT-2~\cite{goyal2024rvt}           & 53.4$_{\pm 1.5}$ & 58.0$_{\pm 1.1}$ & 62.6$_{\pm 0.9}$ & 56.6$_{\pm 0.9}$ & 60.8$_{\pm 0.5}$ & 68.7$_{\pm 1.1}$ & 68.8$_{\pm 1.3}$ & 64.4$_{\pm 0.5}$ \\
\midrule
\textbf{\method (ours)} & 61.7$_{\pm 0.8}$ & \textbf{69.7}$_{\pm 1.2}$ & \textbf{75.7}$_{\pm 0.9}$ & \textbf{71.3}$_{\pm 0.7}$ & 51.8$_{\pm 1.5}$ & \textbf{74.8}$_{\pm 1.0}$ & \textbf{73.1}$_{\pm 0.2}$ & \textbf{73.8}$_{\pm 0.3}$ \\
\textbf{\memmethod (ours)} & \textbf{62.0}$_{\pm 0.7}$ & 68.2$_{\pm 1.0}$ & 71.5$_{\pm 0.3}$ & 69.2$_{\pm 0.7}$ & \textbf{61.6}$_{\pm 0.5}$ & 69.5$_{\pm 1.2}$ & 68.5$_{\pm 0.6}$ & 68.7$_{\pm 0.7}$ \\
\bottomrule
\end{tabular}%
}
\end{table*}

\subsection{COLOSSEUM \& GemBench: Generalization}
\label{sec:exp:additional}

To further evaluate the robustness and generalization capabilities of
\method and \memmethod (Q2), we conduct experiments on
COLOSSEUM~\cite{pumacay2024colosseum} and
GemBench~\cite{garcia2024towards}.
Both benchmarks extend RLBench to evaluate out-of-distribution
generalization.
COLOSSEUM introduces 12 perturbation axes that are unseen during training,
including variations in object color, texture, and size, as well as changes
in background, lighting, distractors, and camera pose.
Together with the original RLBench setting and a combined
all-perturbations setting, it comprises 14 evaluation conditions in total.
GemBench evaluates hierarchical systematic generalization to novel rigid and
articulated objects, as well as unseen object--color compositions.
As reported in Table~\ref{tab:colosseum}, \method achieves a
state-of-the-art average success rate of 64.0\% on COLOSSEUM, outperforming
strong recent 2D and 3D manipulation methods, including RVT-2, 3D-LOTUS,
and 3D Diffuser Actor.
In particular, it exceeds RVT-2 by more than 7 percentage points.
Similarly, \method achieves a state-of-the-art average success rate of
50.0\% on GemBench, as shown in Table~\ref{tab:gembench}.
These results demonstrate the strong robustness of \method under diverse
out-of-distribution conditions.
Importantly, the memory-augmented \memmethod preserves this generalization
capability.
It matches or slightly improves upon \method on both benchmarks, achieving
65.2\% versus 64.0\% on COLOSSEUM and 51.1\% versus 50.0\% on GemBench.
Thus, introducing spatio-temporal memory does not compromise the
out-of-distribution robustness of the original framework.
Additional baseline details and analyses are provided in
Appendices~\ref{app:colosseum_results} and~\ref{app:gembench_results}.

\subsection{RMBench: Memory-Dependent Bimanual Manipulation}
\label{sec:exp:bridgevla_plus}
\label{sec:exp:rmbench}

To answer Q5, we explicitly evaluate \memmethod\ on memory-dependent manipulation tasks, utilizing the RMBench suite.

\paragraph{Setup}
RMBench~\cite{chen2026rmbench} is a dual-arm benchmark specifically designed to test episodic reasoning. Its nine tasks cannot be solved from the current frame alone, requiring the policy to retain past observations across short-term $M(1)$ and long-term $M(n)$ horizons. This concurrently validates our bimanual extension (Sec.~\ref{sec:memmethod:bimanual}). Following the benchmark protocol, we train on 50 demonstrations per task and report success rates over 100 evaluation episodes.

\paragraph{Baselines} 
We compare against strong memory-augmented manipulation methods including Mem-0, MemoryWAM, and several other baseline variants from the benchmark. We also compare against our memory-free base model, \method, to quantify the direct impact of the memory modules.

\paragraph{Results} 
Table~\ref{tab:rmbench_main} summarizes the results. 
The memory-free base model, \method, suffers a severe performance drop, yielding an 18.9\% overall success rate, confirming the necessity of episodic memory.
In stark contrast, \memmethod\ achieves a near-perfect 96.0\% overall success rate, outperforming the strongest memory-augmented baseline MemoryWAM by 13.0 points and the reference Mem-0 by 54.0 points.
It ranks best or tied for best on eight of the nine tasks.
Temporal memory $\mathcal{T}$ proves indispensable for these long-horizon tasks. For instance, in \emph{Battery Try}, a trial-and-error sorting task requiring the tracking of past attempts, \memmethod\ achieves 96\% success compared to the strongest baseline's 41\% (MemoryWAM).
We dissect the distinct roles of the spatial and temporal memories underlying these gains in Sec.~\ref{sec:exp:ablation}.

\subsection{MemoryBench: Single-Arm Memory Validation}

We additionally validate \memmethod\ on MemoryBench~\cite{fang2025sam2act}, a suite of single-arm memory-dependent scenarios. 
In these tasks, \memmethod\ achieves a $99.7{\pm}0.3\%$ success rate, firmly confirming its generalized efficacy in handling episodic memory requirements beyond bimanual coordination. A detailed per-task breakdown and comparison against baselines are provided in Appendix~\ref{app:memorybench_results} (Table~\ref{tab:memorybench}).

\subsection{Real-World Experiments}
\label{sec:exp:real_world}

To address Q4, we deploy both models on real robotic hardware. We first evaluate the base \method\ on general manipulation, probing its sample efficiency and generalization under diverse real-world disturbances; we then evaluate \memmethod, validating the spatio-temporal memory in the physical world while verifying that the memory extension leaves the base manipulation competence intact.

\begin{figure*}[t]
  \centering
  \includegraphics[width=\textwidth]{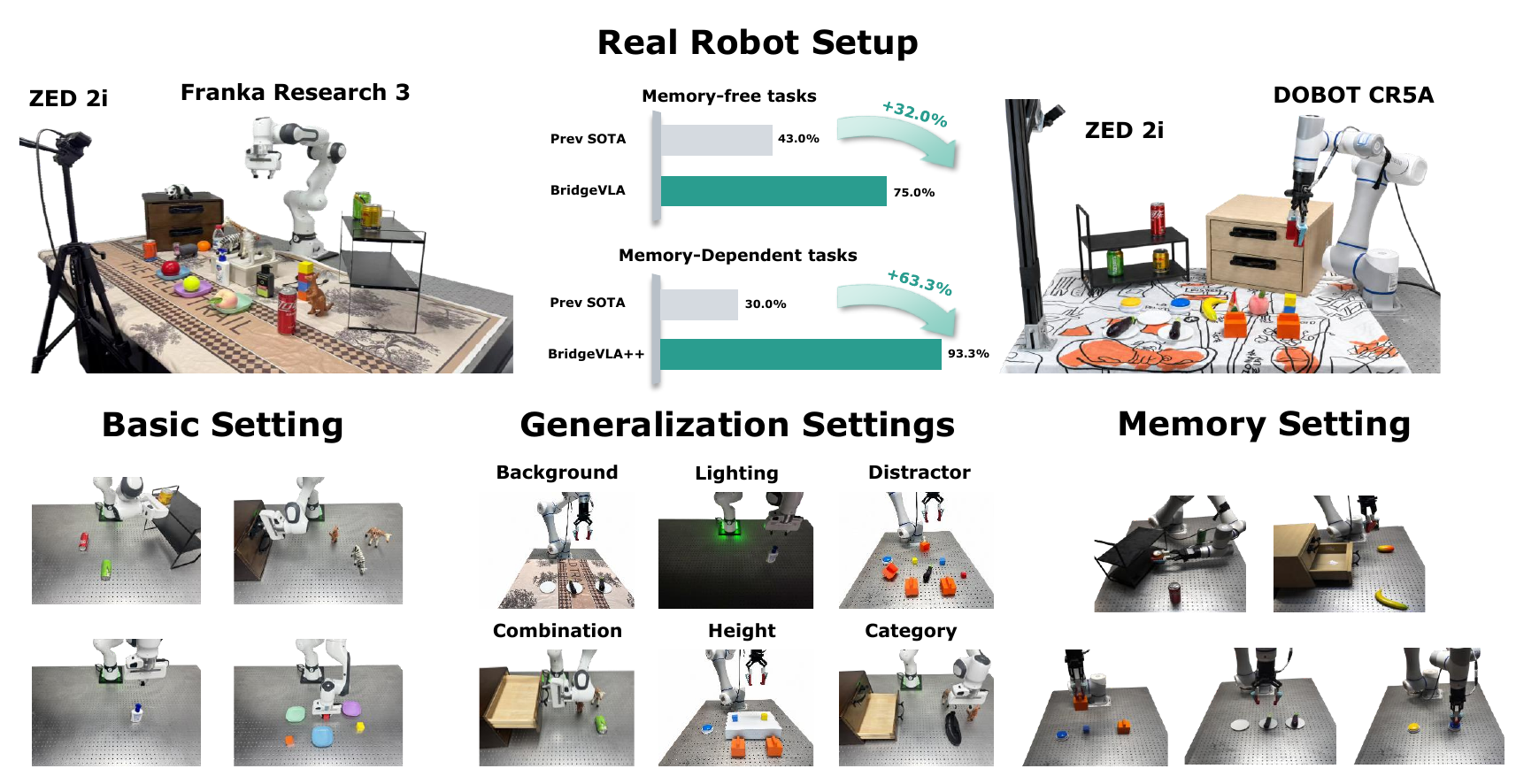}
  \caption{\textbf{Real-Robot Evaluation Setup.}
  \emph{Top left:} the general-manipulation platform---a 7-DoF Franka
  Research~3 arm observed by a static ZED~2i stereo camera.
  \emph{Top right:} the memory platform---a 6-DoF Dobot CR5A arm with
  the same camera configuration.
  \emph{Bottom:} the evaluation settings of the two suites.
  The bars at the top show the average gain of our models over the
  strongest prior method in each task group: RVT-2 across the seven
  Franka settings and SAM2Act+ in the Dobot basic setting.}
  \label{fig:real_setup}
\end{figure*}

\subsubsection{General Manipulation}

\paragraph{Setup}
We evaluate \method\ on a 7-DoF Franka Research 3 manipulator with a parallel-jaw gripper, observed via a static ZED~2i depth camera (Fig.~\ref{fig:real_setup}). The evaluation covers 13 tasks ranging from simple pick-and-place to complex long-horizon manipulation, with 10 expert demonstrations per task for training. Beyond the basic setting, we design six challenging generalization settings: Distractor, Lighting, Background, Height, Combination (unseen object--skill pairings), and Category (unseen object categories). Full setup details, per-task results, and analyses are provided in Appendix~\ref{app:real_franka}.
\begin{table*}[!t]
\centering
\caption{\textbf{Real-Robot Results on Franka in the Basic Setting.}
Success counts on the 13 real-robot tasks; all evaluations are our
own, with 10 trials per task per method
(Appendix~\ref{app:eval_protocol}).
All methods are trained with 10 demonstrations per task, except
SpatialVLA~(50 demos) and the \method~(3 demos) reference row; ACT is
trained single-task, as it is not language-conditioned.
Best result per column in bold (the 3-demonstration row is excluded
from the comparison).}
\label{tab:real_robot}
\footnotesize
\setlength{\tabcolsep}{4pt}
\setlength{\aboverulesep}{0pt}
\setlength{\belowrulesep}{0pt}
\renewcommand{\arraystretch}{1.25}
\resizebox{\fitwidth}{!}{%
\begin{tabular}{l*{7}{c}}
\toprule
& \textbf{Avg.} & \textbf{Soda Can} & \textbf{Giraffe} & \textbf{Red Block} & \textbf{Press} & \textbf{RedBull} & \textbf{RedBull} \\
\multirow{-2}{*}{\textbf{Method}}
& \textbf{SR (\%) $\uparrow$} & \textbf{Bottom Shelf} & \textbf{Lower Drawer} & \textbf{Blue Plate} & \textbf{Sanitizer} & \textbf{Top Shelf} & \textbf{Bottom Shelf} \\
\midrule
SpatialVLA~(50)~\cite{qu2025spatialvla} & 28.5 & 1/10 & 1/10 & 5/10 & 6/10 & 3/10 & 1/10 \\
SpatialVLA~(10)~\cite{qu2025spatialvla} & 3.1  & 0/10 & 0/10 & 0/10 & 2/10 & 0/10 & 0/10 \\
$\pi_{0.5}$~\cite{intelligence2025pi_}            & 20.0 & 2/10 & 1/10 & 4/10 & 4/10 & 1/10 & 1/10 \\
ACT~\cite{zhao2023learning}             & 21.5 & 2/10 & 2/10 & 3/10 & 2/10 & 3/10 & 1/10 \\
RVT-2~\cite{goyal2024rvt}               & 90.0   & \textbf{10/10} & 8/10 & 8/10 & \textbf{10/10} & 9/10 & \textbf{10/10} \\
\midrule
\method~(3 demos) & 95.4 & 9/10 & 10/10 & 10/10 & 10/10 & 9/10 & 10/10 \\
\textbf{\method} & \textbf{96.9} & 9/10 & \textbf{9/10} & \textbf{10/10} & \textbf{10/10} & \textbf{10/10} & \textbf{10/10} \\
\bottomrule
\addlinespace[5pt]
& \textbf{Coke} & \textbf{Orange Block} & \textbf{Red Block} & \textbf{Yellow Block} & \textbf{Zebra} & \textbf{Zebra} & \textbf{Wolf} \\
\multirow{-2}{*}{\textbf{Method}}
& \textbf{Top Shelf} & \textbf{Green Plate} & \textbf{Purple Plate} & \textbf{Green Plate} & \textbf{Upper Drawer} & \textbf{Lower Drawer} & \textbf{Upper Drawer} \\
\midrule
SpatialVLA~(50)~\cite{qu2025spatialvla} & 2/10 & 6/10 & 3/10 & 5/10 & 2/10 & 0/10 & 2/10 \\
SpatialVLA~(10)~\cite{qu2025spatialvla} & 0/10 & 1/10 & 1/10 & 0/10 & 0/10 & 0/10 & 0/10 \\
$\pi_{0.5}$~\cite{intelligence2025pi_}           & 2/10 & 4/10 & 3/10 & 3/10 & 0/10 & 0/10 & 1/10 \\
ACT~\cite{zhao2023learning}             & 2/10 & 2/10 & 3/10 & 4/10 & 1/10 & 2/10 & 1/10 \\
RVT-2~\cite{goyal2024rvt}               & \textbf{10/10} & \textbf{10/10} & 9/10 & 9/10 & 7/10 & 8/10 & \textbf{9/10} \\
\midrule
\method~(3 demos) & 10/10 & 10/10 & 10/10 & 10/10 & 9/10 & 10/10 & 7/10 \\
\textbf{\method} & \textbf{10/10} & \textbf{10/10} & \textbf{10/10} & \textbf{10/10} & \textbf{9/10} & \textbf{10/10} & \textbf{9/10} \\
\bottomrule
\end{tabular}%
}
\end{table*}

\paragraph{Results}
To demonstrate the advantages of \method over existing manipulation
policies, we compare it with four representative baselines spanning different
model categories: SpatialVLA~\cite{qu2025spatialvla}, a 3D VLA model;
$\pi_{0.5}$~\cite{intelligence2025pi_}, a 2D VLA model;
ACT~\cite{zhao2023learning}, a 2D non-VLA policy; and
RVT-2~\cite{goyal2024rvt}, a 3D non-VLA policy.

We first evaluate all methods under the basic setting.
For each task, every method is evaluated over 10 trials.
To ensure a fair comparison, we photograph each test scene and manually
reproduce the same scene configuration across methods.
The results are reported in Table~\ref{tab:real_robot}.
When trained with only 10 trajectories per task, most baselines fail almost
completely, whereas the two methods that explicitly exploit 3D spatial
structure, RVT-2 and \method, achieve substantially stronger performance.
Notably, although SpatialVLA also incorporates 3D information, it remains
considerably less data-efficient.
Even when its training set is increased to 50 trajectories per task, its
success rate remains substantially lower than that of \method.
This result suggests that incorporating 3D information alone is insufficient
for constructing a data-efficient 3D VLA model; the architectural design used
to align the observation and action spaces is also critical.

\begin{figure}[t]
  \centering
  \includegraphics[width=\columnwidth]{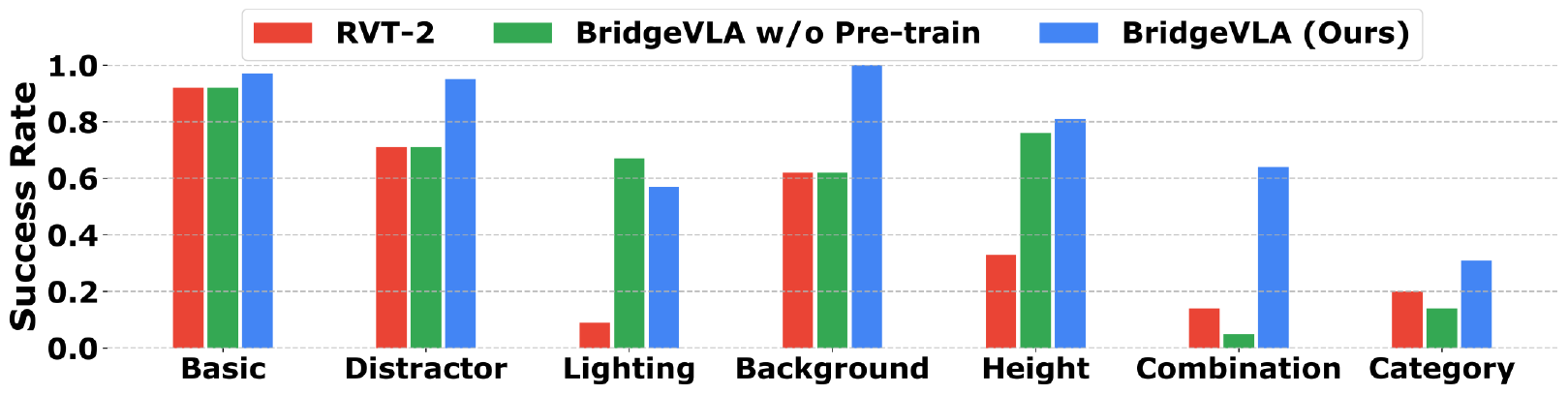}
  \caption{\textbf{Real-Robot Generalization Results.}
  Average success rate over the 13 Franka tasks in the basic setting and
  the six generalization settings;
  the \emph{w/o Pre-train} bars ablate the 2D-heatmap pre-training
  (Sec.~\ref{sec:exp:ablation}).}
  \label{fig:real_results}
\end{figure}

Remarkably, when the training data are further reduced to only three
demonstrations per task, \method still achieves a success rate of 95.4\%,
highlighting its exceptional sample efficiency and directly addressing Q4.
Because only RVT-2 and \method achieve reliable performance under the basic
setting, we further compare these two methods across the remaining
generalization settings.
As summarized in Fig.~\ref{fig:real_results}, \method consistently
outperforms RVT-2 across all seven settings, with an average improvement of
32\%.
The gains are particularly pronounced under novel lighting conditions and
unseen object--skill combinations, demonstrating that \method can
effectively transfer the semantic knowledge of the pre-trained VLM to
real-world manipulation.
Additional details on the experimental setup, baseline implementations,
per-task results, and analyses are provided in
Appendix~\ref{app:real_franka}.

\subsubsection{Memory-Augmented Manipulation}
\paragraph{Setup}
To validate \memmethod\ in the real world, we deploy the model on a Dobot CR5A equipped with an external RGB-D camera to capture the workspace (Fig.~\ref{fig:real_setup}). The evaluation suite pairs three memory-dependent tasks---\emph{Cover Blocks}, \emph{Press Button}, and \emph{Swap Eggplant}---with two memory-free standard manipulation tasks, \emph{Put in Drawer} and \emph{Put on Shelf}. The former probe whether the spatio-temporal memory transfers to physical hardware; the latter verify that the memory extension does not erode the policy's general manipulation capability. Each memory-dependent task is driven by a single language instruction, while \emph{Put in Drawer} and \emph{Put on Shelf} use two instructions that differ in target height. Similar to the setup on the Franka platform, we train on 10 demonstrations per language instruction and evaluate every instruction over 10 trials in each of five settings---Basic, Distractor, Background, Height, and Lighting---against memory-free \method\ and the memory-augmented SAM2Act+~\cite{fang2025sam2act}. Full details are provided in Appendix~\ref{app:real_dobot}.

\paragraph{Results}
Consistent with our simulation findings, the proposed spatio-temporal memory effectively transfers to physical hardware. In the basic setting (Table~\ref{tab:real_dobot_basic}), \memmethod\ achieves an average success rate of 93.3\% on the three memory-dependent tasks, yielding a threefold improvement over the memory-augmented SAM2Act+ (30.0\%). Conversely, the memory-free \method\ largely fails (20.0\%), confirming that these tasks intrinsically require episodic memory. This substantial performance margin over SAM2Act+ stems from architectural differences in memory management. Because SAM2Act+ indiscriminately stores every step and retrieves only a fixed temporal window, near-duplicate frames tend to dilute critical historical information. \memmethod\ overcomes this limitation via its targeted spatio-temporal memory design (Sec.~\ref{sec:memmethod}). Furthermore, owing to its pre-trained VLM backbone, \memmethod\ exhibits strong robustness against visual disturbances (Table~\ref{tab:real_dobot_generalization}). Importantly, this memory integration comes at no cost to general manipulation capabilities: on the two memory-free tasks, \memmethod\ matches or exceeds \method\ in both success rate and generalization across all settings. Collectively, these results demonstrate that \memmethod\ successfully acquires memory-dependent competencies in real-world scenarios while fully preserving the foundational manipulation skills and visual robustness of the base model. Per-instruction counts are detailed in Table~\ref{tab:app:real_dobot_all}.
\begin{table}[!t]
\centering
\caption{\textbf{Per-task success rates on the real Dobot platform in the
basic setting.}
Ten trials per language instruction; \emph{Mem.}\ marks memory-augmented
policies.
\emph{Put in Drawer} and \emph{Put on Shelf} are each evaluated with two
instructions (upper and lower target) and are reported as their average;
per-instruction counts are given in Table~\ref{tab:app:real_dobot_all}.}
\label{tab:real_dobot_basic}
\footnotesize
\setlength{\tabcolsep}{3pt}
\setlength{\aboverulesep}{0pt}
\setlength{\belowrulesep}{0pt}
\renewcommand{\arraystretch}{1.25}
\begin{tabular}{lcccccc}
\toprule
& & \multicolumn{3}{c}{\textbf{Memory-Dependent}}
& \multicolumn{2}{c}{\textbf{Memory-Free}} \\
\cmidrule(lr){3-5}\cmidrule(lr){6-7}
\textbf{Method}
& \textbf{Mem.}
& \textbf{\makecell{Cover\\Blocks}}
& \textbf{\makecell{Press\\Button}}
& \textbf{\makecell{Swap\\Eggplant}}
& \textbf{\makecell{Put in\\Drawer}}
& \textbf{\makecell{Put on\\Shelf}} \\
\midrule
SAM2Act+~\cite{fang2025sam2act} & \cmark & 20.0\% & 0.0\% & 70.0\% & 60.0\% & 20.0\% \\
\textbf{\method} & \xmark & 0.0\% & 0.0\% & 60.0\% & 100.0\% & 90.0\% \\
\textbf{\memmethod} & \cmark & 100.0\% & 100.0\% & 80.0\% & 100.0\% & 100.0\% \\
\bottomrule
\end{tabular}
\end{table}

\begin{table}[!t]
\centering
\caption{\textbf{Success rates on the real Dobot platform across all
evaluation settings.}
Each entry is the mean over the three memory-dependent or two
memory-free tasks of Table~\ref{tab:real_dobot_basic}; \emph{Avg.}\
averages the four disturbance settings.
\emph{Mem.}\ marks memory-augmented policies; per-instruction counts
are given in Table~\ref{tab:app:real_dobot_all}.}
\label{tab:real_dobot_generalization}
\footnotesize
\setlength{\tabcolsep}{3pt}
\setlength{\aboverulesep}{0pt}
\setlength{\belowrulesep}{0pt}
\renewcommand{\arraystretch}{1.25}
\resizebox{\fitwidth}{!}{%
\begin{tabular}{lccccccc}
\toprule
& & & \multicolumn{5}{c}{\textbf{Visual Disturbance}} \\
\cmidrule(lr){4-8}
\textbf{Method}
& \textbf{Mem.}
& \textbf{Basic}
& \textbf{Distractor}
& \textbf{Background}
& \textbf{Height}
& \textbf{Lighting}
& \textbf{Avg.} \\
\midrule
\multicolumn{8}{@{}l}{\emph{Memory-dependent tasks}} \\
SAM2Act+~\cite{fang2025sam2act} & \cmark & 30.0\% & 0.0\% & 0.0\% & 0.0\% & 3.3\% & 0.8\% \\
\textbf{\method} & \xmark & 20.0\% & 20.0\% & 23.3\% & 6.7\% & 13.3\% & 15.8\% \\
\textbf{\memmethod} & \cmark & 93.3\% & 73.3\% & 86.7\% & 76.7\% & 76.7\% & 78.3\% \\
\midrule
\multicolumn{8}{@{}l}{\emph{Memory-free tasks}} \\
SAM2Act+~\cite{fang2025sam2act} & \cmark & 40.0\% & 0.0\% & 0.0\% & 0.0\% & 7.5\% & 1.9\% \\
\textbf{\method} & \xmark & 95.0\% & 57.5\% & 72.5\% & 67.5\% & 67.5\% & 66.3\% \\
\textbf{\memmethod} & \cmark & 100.0\% & 70.0\% & 100.0\% & 82.5\% & 75.0\% & 81.9\% \\
\bottomrule
\end{tabular}}
\end{table}

\subsection{Ablation Studies}
\label{sec:exp:ablation}

To address Q3, we conduct two groups of ablation studies: we first ablate the core architectural designs of the base model \method\ (upper ablation rows of Table~\ref{tab:rlbench}), and then dissect the spatio-temporal memory of \memmethod\ (lower ablation rows of Table~\ref{tab:rlbench} and Table~\ref{tab:rmbench_ablation}).

\textbf{Whether we need to predict heatmaps before predicting actions.}
Replacing the convex upsampling module with a parameter-matched Transformer decoder that directly regresses target positions (Appendix~\ref{app:finetune_hparams}) causes the average success rate on RLBench to collapse from 90.5\% to 31.4\% (Table~\ref{tab:rlbench}).
The ablated model is also markedly harder to optimize, demanding a threefold larger batch size (192 vs. 64) and careful learning-rate tuning.
We attribute this gap to three properties of the heatmap as an intermediate representation: it provides denser supervision than sparse 3D position vectors, the 3D-to-2D projection injects a helpful spatial prior, and the heatmaps share the spatial structure of the input images, keeping input and output aligned.

\textbf{Whether we need to remove the 3D position input to the VLM backbone.}
Unlike typical 3D VLA models such as SpatialVLA, \method\ feeds the backbone only RGB projection images.
Fusing per-pixel 3D positions into the image features via a 3D convolutional module (Appendix~\ref{app:finetune_hparams}) injects richer spatial cues, yet degrades the success rate from 90.5\% to 56.2\% (Table~\ref{tab:rlbench}), which we attribute to the resulting shift of the image features away from the distribution seen during VLM pre-training.
For a pre-trained VLM, preserving input alignment thus outweighs adding explicit 3D inputs.

\textbf{Whether we need the 2D heatmap pre-training.}
Without the pre-training stage, \method\ w/o Pre-train fails to generalize in both language-related real-world settings and cannot even match RVT-2, whereas the full \method\ performs best in both, especially in Combination (Fig.~\ref{fig:real_results}).
We hypothesize that the 2D heatmap pre-training teaches the model to ground language semantics in image observations directly within the heatmap space, an ability that fine-tuning on a handful of robot trajectories alone cannot instill.

\textbf{Whether a continuous rotation representation outperforms a discretized one.}
Replacing the continuous 6D rotation representation (Sec.~\ref{sec:bridgevla:finetune}) with the discretized per-axis Euler-angle classification head from our preliminary conference version degrades the RLBench average from 90.5\% to 88.2\% (Table~\ref{tab:rlbench}), with the drop concentrated in tasks demanding high-precision end-effector orientations.
The 6D representation also stays robust in near-vertical gripper poses, avoiding the gimbal lock inherent to discretized Euler angles (Appendix~\ref{app:finetune_hparams}).

\textbf{Whether we need the spatial memory $\mathcal{S}$.}
Since $\mathcal{S}$ targets fine-grained geometric alignment under arm-induced occlusions (Sec.~\ref{sec:memmethod:spatial}), we ablate it mainly on RLBench, whose precision tasks are exactly where such occlusions arise.
Removing $\mathcal{S}$ lowers the RLBench average of \memmethod\ (93.7\% vs. 92.0\%), and the loss concentrates exactly in occlusion-heavy precision tasks, \eg, \emph{Sort Shape} (72.0\% vs. 60.8\%, Table~\ref{tab:rlbench}); on RMBench, whose tasks stress temporal sequencing rather than geometric alignment, the removal is nearly harmless (Table~\ref{tab:rmbench_ablation}).
This benchmark-selective effect indicates that $\mathcal{S}$ contributes complementary geometric detail for precise alignment rather than duplicating the temporal memory.

\textbf{Whether we need the temporal memory $\mathcal{T}$.}
Removing $\mathcal{T}$ collapses the RMBench success rate of \memmethod\ from 96.0\% to 21.3\%, close to the memory-free base model (18.9\%, Table~\ref{tab:rmbench_ablation}).
Notably, $\mathcal{T}$ also benefits RLBench (93.7\% vs. 91.9\%) despite its tasks not being intrinsically memory-dependent: the anchor views and neighboring keyframes provide a stable global reference and local motion cues that help general manipulation (Table~\ref{tab:rlbench}).

All the above results address Q3: heatmap-based action prediction, input alignment, and the temporal memory account for the largest gains; the heatmap pre-training underpins language-conditioned generalization; and the continuous rotation and the spatial memory contribute smaller, targeted gains on orientation-critical and occlusion-heavy tasks, respectively.


\section{Conclusion and Future Work}
\label{sec:conclusion}
This article has presented \method, an efficient 3D vision-language-action model built upon a pre-trained vision-language model (VLM)~\cite{beyer2024paligemma}, together with its memory-augmented extension \memmethod.
\method rests on a single alignment principle: 3D observations are rendered as multi-view 2D images to match the input space of the VLM, actions are expressed as 2D heatmaps in the same image space, and a scalable pre-training stage teaches the VLM to predict heatmaps before it is fine-tuned for action prediction.
\memmethod extends this principle with a unified spatio-temporal memory: temporal memory retains the interaction history to determine \emph{what to do next}, while spatial memory re-renders previously observed geometry to determine \emph{where exactly to act}.
Extensive experiments on standard and memory-dependent benchmarks, in both simulation and the real world, show that the framework learns 3D manipulation efficiently and effectively, and that the memory is strictly additive to the base policy.
Future work includes broadening pre-training to more diverse tasks such as semantic segmentation and keypoint detection, adopting more expressive action decoders, and replacing the annotation-dependent sub-goal gate with self-supervised alternatives.
Moreover, since the token-space memory injection is agnostic to timescale, 
extending this mechanism to support cross-episode or life-long memory represents a promising path toward robots that continuously accumulate, refine, and transfer manipulation skills across long operational horizons.

\bibliographystyle{IEEEtran}
\bibliography{main}

@IEEEtranBSTCTL{IEEEexample:BSTcontrol,
  CTLuse_forced_etal       = "yes",
  CTLmax_names_forced_etal = "6",
  CTLnames_show_etal       = "1"
}

@inproceedings{johns2021coarse,
  title={Coarse-to-fine imitation learning: Robot manipulation from a single demonstration},
  author={Johns, Edward},
  booktitle={2021 IEEE international conference on robotics and automation (ICRA)},
  pages={4613--4619},
  year={2021},
  organization={IEEE}
}

@inproceedings{zhao2023learning,
  title={Learning Fine-Grained Bimanual Manipulation with Low-Cost Hardware},
  author={Zhao, Tony Z and Kumar, Vikash and Levine, Sergey and Finn, Chelsea},
  booktitle={Robotics: Science and Systems (RSS)},
  year={2023}
}

@inproceedings{brohan2022rt,
  title={{RT-1}: Robotics Transformer for Real-World Control at Scale},
  author={Brohan, Anthony and Brown, Noah and Carbajal, Justice and Chebotar, Yevgen and Dabis, Joseph and Finn, Chelsea and Gopalakrishnan, Keerthana and Hausman, Karol and Herzog, Alex and Hsu, Jasmine and others},
  booktitle={Robotics: Science and Systems (RSS)},
  year={2023}
}

@misc{xiao2022masked,
  title={Masked visual pre-training for motor control},
  author={Xiao, Tete and Radosavovic, Ilija and Darrell, Trevor and Malik, Jitendra},
  note = {arXiv:2203.06173},
  year={2022}
}

@inproceedings{gervet2023act3d,
  title={{Act3D}: {3D} Feature Field Transformers for Multi-Task Robotic Manipulation},
  author={Gervet, Theophile and Xian, Zhou and Gkanatsios, Nikolaos and Fragkiadaki, Katerina},
  booktitle={Conference on Robot Learning},
  pages={3949--3965},
  year={2023},
  organization={PMLR}
}

@inproceedings{3d-da,
  title={{3D Diffuser Actor}: Policy Diffusion with {3D} Scene Representations},
  author={Ke, Tsung-Wei and Gkanatsios, Nikolaos and Fragkiadaki, Katerina},
  booktitle={Conference on Robot Learning},
  pages={1949--1974},
  year={2025}
}

@inproceedings{shridhar2023perceiver,
  title={{Perceiver-Actor}: A Multi-Task Transformer for Robotic Manipulation},
  author={Shridhar, Mohit and Manuelli, Lucas and Fox, Dieter},
  booktitle={Conference on Robot Learning},
  pages={785--799},
  year={2023},
  organization={PMLR}
}

@inproceedings{goyal2023rvt,
  title={{RVT}: Robotic View Transformer for {3D} Object Manipulation},
  author={Goyal, Ankit and Xu, Jie and Guo, Yijie and Blukis, Valts and Chao, Yu-Wei and Fox, Dieter},
  booktitle={Conference on Robot Learning},
  pages={694--710},
  year={2023},
  organization={PMLR}
}

@inproceedings{goyal2024rvt,
  title={{RVT-2}: Learning Precise Manipulation from Few Demonstrations},
  author={Goyal, Ankit and Blukis, Valts and Xu, Jie and Guo, Yijie and Chao, Yu-Wei and Fox, Dieter},
  booktitle={Robotics: Science and Systems (RSS)},
  year={2024}
}

@inproceedings{guhur2023instruction,
  title={Instruction-driven history-aware policies for robotic manipulations},
  author={Guhur, Pierre-Louis and Chen, Shizhe and Pinel, Ricardo Garcia and Tapaswi, Makarand and Laptev, Ivan and Schmid, Cordelia},
  booktitle={Conference on Robot Learning},
  pages={175--187},
  year={2023},
  organization={PMLR}
}

@inproceedings{kimopenvla,
  title={{OpenVLA}: An Open-Source Vision-Language-Action Model},
  author={Kim, Moo Jin and Pertsch, Karl and Karamcheti, Siddharth and Xiao, Ted and Balakrishna, Ashwin and Nair, Suraj and Rafailov, Rafael and Foster, Ethan P and Sanketi, Pannag R and Vuong, Quan and Kollar, Thomas and Burchfiel, Benjamin and Tedrake, Russ and Sadigh, Dorsa and Levine, Sergey and Liang, Percy and Finn, Chelsea},
  booktitle={Conference on Robot Learning},
  pages={2679--2713},
  year={2025}
}

@inproceedings{li2023vision,
  title={Vision-Language Foundation Models as Effective Robot Imitators},
  author={Li, Xinghang and Liu, Minghuan and Zhang, Hanbo and Yu, Cunjun and Xu, Jie and Wu, Hongtao and Cheang, Chilam and Jing, Ya and Zhang, Weinan and Liu, Huaping and Li, Hang and Kong, Tao},
  booktitle={International Conference on Learning Representations (ICLR)},
  year={2024}
}

@inproceedings{qu2025spatialvla,
  title={{SpatialVLA}: Exploring Spatial Representations for Visual-Language-Action Models},
  author={Qu, Delin and Song, Haoming and Chen, Qizhi and Yao, Yuanqi and Ye, Xinyi and Gu, Jiayuan and Wang, Zhigang and Ding, Yan and Zhao, Bin and Wang, Dong and Li, Xuelong},
  booktitle={Robotics: Science and Systems (RSS)},
  year={2025}
}

@inproceedings{chen2023polarnet,
  title={{PolarNet}: {3D} Point Clouds for Language-Guided Robotic Manipulation},
  author={Chen, Shizhe and Pinel, Ricardo Garcia and Schmid, Cordelia and Laptev, Ivan},
  booktitle={Conference on Robot Learning},
  pages={1761--1781},
  year={2023},
  organization={PMLR}
}

@misc{yang2025fp3,
  title={{FP3}: A {3D} Foundation Policy for Robotic Manipulation},
  author={Yang, Rujia and Chen, Geng and Wen, Chuan and Gao, Yang},
  note = {arXiv:2503.08950},
  year={2025}
}

@article{chi2024diffusionpolicy,
	author = {Cheng Chi and Zhenjia Xu and Siyuan Feng and Eric Cousineau and Yilun Du and Benjamin Burchfiel and Russ Tedrake and Shuran Song},
	title ={Diffusion Policy: Visuomotor Policy Learning via Action Diffusion},
	journal = {The International Journal of Robotics Research},
	year = {2024},
}

@inproceedings{jaegleperceiver,
  title={{Perceiver IO}: A General Architecture for Structured Inputs \& Outputs},
  author={Jaegle, Andrew and Borgeaud, Sebastian and Alayrac, Jean-Baptiste and Doersch, Carl and Ionescu, Catalin and Ding, David and Koppula, Skanda and Zoran, Daniel and Brock, Andrew and Shelhamer, Evan and others},
  booktitle={International Conference on Learning Representations},
  year={2022}
}

@inproceedings{brohan2023rt,
  title={{RT-2}: Vision-Language-Action Models Transfer Web Knowledge to Robotic Control},
  author={Brohan, Anthony and Brown, Noah and Carbajal, Justice and Chebotar, Yevgen and Chen, Xi and Choromanski, Krzysztof and Ding, Tianli and Driess, Danny and Dubey, Avinava and Finn, Chelsea and others},
  booktitle={Conference on Robot Learning},
  pages={2165--2183},
  year={2023},
  organization={PMLR}
}

@inproceedings{o2024open,
  title={Open {X-Embodiment}: Robotic Learning Datasets and {RT-X} Models},
  author={O’Neill, Abby and Rehman, Abdul and Maddukuri, Abhiram and Gupta, Abhishek and Padalkar, Abhishek and Lee, Abraham and Pooley, Acorn and Gupta, Agrim and Mandlekar, Ajay and Jain, Ajinkya and others},
  booktitle={2024 IEEE International Conference on Robotics and Automation (ICRA)},
  pages={6892--6903},
  year={2024},
  organization={IEEE}
}

@inproceedings{pertsch2025fast,
  title={{FAST}: Efficient Action Tokenization for Vision-Language-Action Models},
  author={Pertsch, Karl and Stachowicz, Kyle and Ichter, Brian and Driess, Danny and Nair, Suraj and Vuong, Quan and Mees, Oier and Finn, Chelsea and Levine, Sergey},
  booktitle={Robotics: Science and Systems (RSS)},
  year={2025}
}

@inproceedings{zhen20243d,
  title={{3D-VLA}: A {3D} Vision-Language-Action Generative World Model},
  author={Zhen, Haoyu and Qiu, Xiaowen and Chen, Peihao and Yang, Jincheng and Yan, Xin and Du, Yilun and Hong, Yining and Gan, Chuang},
  booktitle={International Conference on Machine Learning},
  pages={61229--61245},
  year={2024},
  organization={PMLR}
}

@article{james2020rlbench,
  title={{RLBench}: The Robot Learning Benchmark \& Learning Environment},
  author={James, Stephen and Ma, Zicong and Arrojo, David Rovick and Davison, Andrew J},
  journal={IEEE Robotics and Automation Letters},
  volume={5},
  number={2},
  pages={3019--3026},
  year={2020},
  publisher={IEEE}
}

@article{li2025pointvla,
  title={{PointVLA}: Injecting the {3D} World Into Vision-Language-Action Models},
  author={Li, Chengmeng and Wen, Junjie and Peng, Yaxin and Peng, Yan and Zhu, Yichen},
  journal={IEEE Robotics and Automation Letters},
  volume={11},
  number={3},
  pages={2506--2513},
  year={2026}
}

@inproceedings{james2022coarse,
  title={Coarse-to-Fine {Q}-Attention: Efficient Learning for Visual Robotic Manipulation via Discretisation},
  author={James, Stephen and Wada, Kentaro and Laidlow, Tristan and Davison, Andrew J},
  booktitle={Proceedings of the IEEE/CVF Conference on Computer Vision and Pattern Recognition},
  pages={13739--13748},
  year={2022}
}

@misc{yuan2024robopoint,
  title={{RoboPoint}: A Vision-Language Model for Spatial Affordance Prediction for Robotics},
  author={Yuan, Wentao and Duan, Jiafei and Blukis, Valts and Pumacay, Wilbert and Krishna, Ranjay and Murali, Adithyavairavan and Mousavian, Arsalan and Fox, Dieter},
  note = {arXiv:2406.10721},
  year={2024}
}

@misc{beyer2024paligemma,
  title={{PaliGemma}: A Versatile {3B} {VLM} for Transfer},
  author={Beyer, Lucas and Steiner, Andreas and Pinto, Andr{\'e} Susano and Kolesnikov, Alexander and Wang, Xiao and Salz, Daniel and Neumann, Maxim and Alabdulmohsin, Ibrahim and Tschannen, Michael and Bugliarello, Emanuele and others},
  note = {arXiv:2407.07726},
  year={2024}
}

@inproceedings{zhai2023sigmoid,
  title={Sigmoid loss for language image pre-training},
  author={Zhai, Xiaohua and Mustafa, Basil and Kolesnikov, Alexander and Beyer, Lucas},
  booktitle={Proceedings of the IEEE/CVF international conference on computer vision},
  pages={11975--11986},
  year={2023}
}

@misc{team2024gemma,
  title={Gemma: Open Models Based on {Gemini} Research and Technology},
  author={{Gemma Team} and Mesnard, Thomas and Hardin, Cassidy and Dadashi, Robert and Bhupatiraju, Surya and Pathak, Shreya and Sifre, Laurent and Rivi{\`e}re, Morgane and Kale, Mihir Sanjay and Love, Juliette and others},
  note = {arXiv:2403.08295},
  year={2024}
}

@inproceedings{teed2020raft,
  title={{RAFT}: Recurrent All-Pairs Field Transforms for Optical Flow},
  author={Teed, Zachary and Deng, Jia},
  booktitle={Computer Vision--ECCV 2020: 16th European Conference, Glasgow, UK, August 23--28, 2020, Proceedings, Part II 16},
  pages={402--419},
  year={2020},
  organization={Springer}
}

@article{li2026bridgevla,
  title={{BridgeVLA}: Input-Output Alignment for Efficient {3D} Manipulation Learning with Vision-Language Models},
  author={Li, Peiyan and Chen, Yixiang and Wu, Hongtao and Ma, Xiao and Wu, Xiangnan and Huang, Yan and Wang, Liang and Kong, Tao and Tan, Tieniu},
  journal={Advances in Neural Information Processing Systems},
  volume={38},
  pages={63635--63673},
  year={2025}
}

@inproceedings{fang2025sam2act,
  title={{SAM2Act}: Integrating Visual Foundation Model with a Memory Architecture for Robotic Manipulation},
  author={Fang, Haoquan and Grotz, Markus and Pumacay, Wilbert and Wang, Yi Ru and Fox, Dieter and Krishna, Ranjay and Duan, Jiafei},
  booktitle={International Conference on Machine Learning},
  pages={15925--15942},
  year={2025},
  organization={PMLR}
}

@misc{chen2026rmbench,
      title={{RMBench}: Memory-Dependent Robotic Manipulation Benchmark with Insights into Policy Design},
      author={Tianxing Chen and Yuran Wang and Mingleyang Li and Yan Qin and Hao Shi and Zixuan Li and Yifan Hu and Yingsheng Zhang and Kaixuan Wang and Yue Chen and Hongcheng Wang and Renjing Xu and Ruihai Wu and Yao Mu and Yaodong Yang and Hao Dong and Ping Luo},
      year={2026},
      eprint={2603.01229},
      archivePrefix={arXiv},
      primaryClass={cs.RO},
      url={https://arxiv.org/abs/2603.01229}
}

@misc{chen2025robotwin2,
  title={{RoboTwin} 2.0: A Scalable Data Generator and Benchmark with Strong Domain Randomization for Robust Bimanual Robotic Manipulation},
  author={Chen, Tianxing and Chen, Zanxin and Chen, Baijun and Cai, Zijian and Liu, Yibin and Li, Zixuan and Liang, Qiwei and Lin, Xianliang and Ge, Yiheng and Gu, Zhenyu and Deng, Weiliang and Guo, Yubin and Nian, Tian and Xie, Xuanbing and Chen, Qiangyu and Su, Kailun and Xu, Tianling and Liu, Guodong and Hu, Mengkang and Gao, Huan-ang and Wang, Kaixuan and Liang, Zhixuan and Qin, Yusen and Yang, Xiaokang and Luo, Ping and Mu, Yao},
  note = {arXiv:2506.18088},
  year={2025}
}

@misc{pumacay2024colosseum,
  title={The {Colosseum}: A Benchmark for Evaluating Generalization for Robotic Manipulation},
  author={Pumacay, Wilbert and Singh, Ishika and Duan, Jiafei and Krishna, Ranjay and Thomason, Jesse and Fox, Dieter},
  note = {arXiv:2402.08191},
  year={2024}
}

@inproceedings{rohmer2013v,
  title={{V-REP}: A Versatile and Scalable Robot Simulation Framework},
  author={Rohmer, Eric and Singh, Surya PN and Freese, Marc},
  booktitle={2013 IEEE/RSJ international conference on intelligent robots and systems},
  pages={1321--1326},
  year={2013},
  organization={IEEE}
}

@inproceedings{xiang2020sapien,
  title={{SAPIEN}: A SimulAted Part-based Interactive ENvironment},
  author={Xiang, Fanbo and Qin, Yuzhe and Mo, Kaichun and Xia, Yikuan and Zhu, Hao and Liu, Fangchen and Liu, Minghua and Jiang, Hanxiao and Yuan, Yifu and Wang, He and Yi, Li and Chang, Angel X and Guibas, Leonidas J and Su, Hao},
  booktitle={Proceedings of the IEEE/CVF Conference on Computer Vision and Pattern Recognition},
  pages={11097--11107},
  year={2020}
}

@misc{nair2022r3m,
  title={{R3M}: A Universal Visual Representation for Robot Manipulation},
  author={Nair, Suraj and Rajeswaran, Aravind and Kumar, Vikash and Finn, Chelsea and Gupta, Abhinav},
  note = {arXiv:2203.12601},
  year={2022}
}

@inproceedings{black2024pi_0,
  title={$\pi_0$: A Vision-Language-Action Flow Model for General Robot Control},
  author={Black, Kevin and Brown, Noah and Driess, Danny and Esmail, Adnan and Equi, Michael Robert and Finn, Chelsea and Fusai, Niccolo and Groom, Lachy and Hausman, Karol and Ichter, Brian and others},
  booktitle={Robotics: Science and Systems (RSS)},
  year={2025}
}

@misc{intelligence2025pi_,
  title={$\pi$0.5: a Vision-Language-Action Model with Open-World Generalization},
  author={{Physical Intelligence} and Black, Kevin and Brown, Noah and Darpinian, James and Dhabalia, Karan and Driess, Danny and Esmail, Adnan and Equi, Michael and Finn, Chelsea and Fusai, Niccolo and others},
  note = {arXiv:2504.16054},
  year={2025}
}

@misc{intelligence2025pi_06star,
  title={{$\pi^{*}_{0.6}$}: a VLA That Learns From Experience},
  author={{Physical Intelligence} and Amin, Ali and Aniceto, Raichelle and Balakrishna, Ashwin and Black, Kevin and Conley, Ken and Connors, Grace and Darpinian, James and Dhabalia, Karan and DiCarlo, Jared and others},
  note = {arXiv:2511.14759},
  year={2025}
}

@misc{intelligence2026pi_07,
  title={{$\pi_{0.7}$}: a Steerable Generalist Robotic Foundation Model with Emergent Capabilities},
  author={{Physical Intelligence} and Ai, Bo and Amin, Ali and Aniceto, Raichelle and Balakrishna, Ashwin and Balke, Greg and Black, Kevin and Bokinsky, George and Cao, Shihao and Charbonnier, Thomas and others},
  note = {arXiv:2604.15483},
  year={2026}
}

@misc{generalist2025gen0,
  author       = {{Generalist Team}},
  title        = {{GEN-0}: Embodied Foundation Models That Scale with Physical Interaction},
  howpublished = {Generalist AI Blog},
  year         = {2025},
  url          = {https://generalistai.com/blog/gen-0}
}

@misc{generalist2026gen1,
  author       = {{Generalist Team}},
  title        = {{GEN-1}: Scaling Embodied Foundation Models to Mastery},
  howpublished = {Generalist AI Blog},
  year         = {2026},
  url          = {https://generalistai.com/blog/gen-1}
}

@misc{
     genesis2026gene265,
     author = {{Genesis AI Team}},
     title = {{GENE-26.5}: Advancing Robotic Manipulation to Human Level},
     howpublished = {Genesis AI Blog},
     month = may,
     year = {2026},
     url = {https://genesis.ai/blog/gene-26-5-advancing-robotic-manipulation-to-human-level},
}

@misc{sunday2026act2preview,
  author       = {{Sunday Robotics}},
  title        = {{ACT-2} Preview: Generalizing Reliability},
  howpublished = {Sunday Robotics Blog},
  year         = {2026},
  month        = jul,
  url          = {https://www.sunday.ai/blog/act-2-preview}
}

@inproceedings{yuan2023m2t2,
  title={{M2T2}: Multi-Task Masked Transformer for Object-Centric Pick and Place},
  author={Yuan, Wentao and Murali, Adithyavairavan and Mousavian, Arsalan and Fox, Dieter},
  booktitle={Conference on Robot Learning},
  pages={3619--3630},
  year={2023},
  organization={PMLR}
}

@inproceedings{jia2024lift3d,
  title={{Lift3D} Policy: Lifting {2D} Foundation Models for Robust {3D} Robotic Manipulation},
  author={Jia, Yueru and Liu, Jiaming and Chen, Sixiang and Gu, Chenyang and Wang, Zhilve and Luo, Longzan and Li, Xiaoqi and Wang, Pengwei and Wang, Zhongyuan and Zhang, Renrui and Zhang, Shanghang},
  booktitle={Proceedings of the IEEE/CVF Conference on Computer Vision and Pattern Recognition},
  pages={17347--17358},
  year={2025}
}

@article{oquab2023dinov2,
  title={{DINOv2}: Learning Robust Visual Features without Supervision},
  author={Oquab, Maxime and Darcet, Timoth{\'e}e and Moutakanni, Th{\'e}o and Vo, Huy and Szafraniec, Marc and Khalidov, Vasil and Fernandez, Pierre and Haziza, Daniel and Massa, Francisco and El-Nouby, Alaaeldin and others},
  journal={Transactions on Machine Learning Research},
  year={2024}
}

@inproceedings{garcia2024towards,
  title={Towards Generalizable Vision-Language Robotic Manipulation: A Benchmark and {LLM}-Guided {3D} Policy},
  author={Garcia, Ricardo and Chen, Shizhe and Schmid, Cordelia},
  booktitle={2025 IEEE International Conference on Robotics and Automation (ICRA)},
  pages={8996--9002},
  year={2025},
  organization={IEEE}
}

@inproceedings{wu2024point,
  title={Point Transformer {V3}: Simpler Faster Stronger},
  author={Wu, Xiaoyang and Jiang, Li and Wang, Peng-Shuai and Liu, Zhijian and Liu, Xihui and Qiao, Yu and Ouyang, Wanli and He, Tong and Zhao, Hengshuang},
  booktitle={Proceedings of the IEEE/CVF Conference on Computer Vision and Pattern Recognition},
  pages={4840--4851},
  year={2024}
}

@inproceedings{zhou2019continuity,
  title={On the continuity of rotation representations in neural networks},
  author={Zhou, Yi and Barnes, Connelly and Lu, Jingwan and Yang, Jimei and Li, Hao},
  booktitle={Proceedings of the IEEE/CVF Conference on Computer Vision and Pattern Recognition},
  pages={5745--5753},
  year={2019}
}

@misc{yang2026memorywam,
  title={{MemoryWAM}: Efficient World Action Modeling with Persistent Memory},
  author={Yang, Sizhe and Mu, Juncheng and Wei, Tianming and Lu, Chenhao and Li, Xiaofan and Xu, Linning and Xue, Zhengrong and Yuan, Zhecheng and Lin, Dahua and Pang, Jiangmiao and Xu, Huazhe},
  note = {arXiv:2606.20562},
  year={2026}
}

@misc{yang2026wla,
  title={World-Language-Action Model for Unified World Modeling, Language Reasoning, and Action Synthesis},
  author={Yang, Yi and Liu, Zhihong and Kou, Siqi and Chen, Yiyang and Hu, Yanzhe and Zhou, Jianbo and Zhao, Boyuan and Wei, Zhijie and Xia, Xiao and Li, Xueqi and Liu, Pengfei and Deng, Zhijie},
  note = {arXiv:2606.05979},
  year={2026}
}

@misc{zheng2025xvla,
  title={{X-VLA}: Soft-Prompted Transformer as Scalable Cross-Embodiment Vision-Language-Action Model},
  author={Zheng, Jinliang and Li, Jianxiong and Wang, Zhihao and Liu, Dongxiu and Kang, Xirui and Feng, Yuchun and Zheng, Yinan and Zou, Jiayin and Chen, Yilun and Zeng, Jia and Zhang, Ya-Qin and Pang, Jiangmiao and Liu, Jingjing and Wang, Tai and Zhan, Xianyuan},
  note = {arXiv:2510.10274},
  year={2025}
}

@misc{yuan2026fastwam,
  title={{Fast-WAM}: Do World Action Models Need Test-Time Future Imagination?},
  author={Yuan, Tianyuan and Dong, Zibin and Liu, Yicheng and Zhao, Hang},
  note = {arXiv:2603.16666},
  year={2026}
}

@misc{li2026lingbotva,
  title={Causal World Modeling for Robot Control},
  author={Li, Lin and Zhang, Qihang and Luo, Yiming and Yang, Shuai and Wang, Ruilin and Han, Fei and Yu, Mingrui and Gao, Zelin and Xue, Nan and Zhu, Xing and Shen, Yujun and Xu, Yinghao},
  note = {arXiv:2601.21998},
  year={2026}
}

@inproceedings{shi2026memoryvla,
  title     = {{MemoryVLA}: Perceptual-Cognitive Memory in Vision-Language-Action Models for Robotic Manipulation},
  author    = {Shi, Hao and Xie, Bin and Liu, Yingfei and Sun, Lin and Liu, Fengrong and Wang, Tiancai and Zhou, Erjin and Fan, Haoqiang and Zhang, Xiangyu and Huang, Gao},
  booktitle = {International Conference on Learning Representations (ICLR)},
  year      = {2026}
}

@misc{lei2025robomemory,
  title   = {{RoboMemory}: A Brain-inspired Multi-memory Agentic Framework for Interactive Environmental Learning in Physical Embodied Systems},
  author  = {Lei, Mingcong and Cai, Honghao and Yang, Yuyuan and Wu, Yimou and Ren, Jinke and Cui, Zezhou and Tan, Liangchen and Hong, Junkun and Hu, Gehan and Zhu, Shuangyu and Jiang, Shaohan and Wang, Ge and Tan, Junyuan and Wan, Zhenglin and Li, Zheng and Li, Zhen and Cui, Shuguang and Zhao, Yiming and Han, Yatong},
  note = {arXiv:2508.01415},
  year    = {2025}
}

@inproceedings{zheng2024tracevla,
  title     = {{TraceVLA}: Visual Trace Prompting Enhances Spatial-Temporal Awareness for Generalist Robotic Policies},
  author    = {Zheng, Ruijie and Liang, Yongyuan and Huang, Shuaiyi and Gao, Jianfeng and Daum{\'e} III, Hal and Kolobov, Andrey and Huang, Furong and Yang, Jianwei},
  booktitle = {International Conference on Learning Representations (ICLR)},
  year      = {2025}
}

@misc{yu2026wall,
  title={{WALL-OSS}-0.5 Technical Report},
  author={Yu, Ryan and Zhang, Pushi and Liu, Starrick and Liu, Brae and Kang, Miracle and Li, Shalfun and Shi, Lights and Ma, Ellie and Yang, Ping and Pan, Chris and others},
  note = {arXiv:2605.30877},
  year={2026}
}

@misc{singh2025ogvla,
  title={{OG-VLA}: Orthographic Image Generation for {3D}-Aware Vision-Language Action Model},
  author={Singh, Ishika and Goyal, Ankit and Birchfield, Stan and Fox, Dieter and Garg, Animesh and Blukis, Valts},
  note = {arXiv:2506.01196},
  year={2025}
}

@inproceedings{zhou2024yoso,
  title={You Only Scan Once: A Dynamic Scene Reconstruction Pipeline for {6-DoF} Robotic Grasping of Novel Objects},
  author={Zhou, Lei and Wang, Haozhe and Zhang, Zhengshen and Liu, Zhiyang and Tay, Francis EH and Ang, Marcelo H.},
  booktitle={IEEE International Conference on Robotics and Automation (ICRA)},
  year={2024}
}

@misc{zheng2026memworld,
  title={{Mem-World}: Memory-Augmented Action-Conditioned World Models for Persistent Robot Manipulation},
  author={Zheng, Zirui and Yu, Jiaqian and Peng, Xiongfeng and Shi, Jun and Li, Mingyi and Zhang, Chao and Li, Weiming and Wang, Dong and Lu, Huchuan and Jia, Xu},
  note = {arXiv:2606.18960},
  year={2026}
}

@misc{gao2026gatedmemory,
  title={Gated Memory Policy},
  author={Gao, Yihuai and Liu, Jinyun and Li, Shuang and Song, Shuran},
  note = {arXiv:2604.18933},
  year={2026}
}

\clearpage
\appendix
\subsection{Network and Memory Architecture}
\label{app:arch_details}

Each memory injection block of Sec.~\ref{sec:memmethod:integration}
(Fig.~\ref{fig:bridgevla_arch}) stacks $L{=}2$ layers with 8 attention heads of dimension 128 and a
feed-forward expansion factor of 2, and operates in the
2048-dimensional patch-token space of the backbone.
The temporal memory uses two such blocks, one for the initial anchor
views and one for the dynamic keyframe bank, and the spatial memory a
third (Fig.~\ref{fig:anchor_memory}); the anchor block concatenates the three views so that attention
can track scene changes across views, whereas the other two blocks
restrict each view's tokens to the corresponding memory view.

\subsection{Pre-Training}
\label{app:pretrain_details}

All fine-tuning runs in this article warm-start from a single
2D-heatmap pre-training run that instantiates
Sec.~\ref{sec:bridgevla:pretrain} on the 120K object-detection split
of RoboPoint~\cite{yuan2024robopoint}; Fig.~\ref{fig:pretrain_dataset}
illustrates how the ground-truth heatmaps are rendered from the
detection boxes, and Fig.~\ref{fig:real_heatmaps} the predictions the
fine-tuned model still produces on such data.
The memory injection blocks and the sub-goal gate of
Sec.~\ref{sec:memmethod} are not pre-trained and are instead trained
from scratch during fine-tuning.
Whenever a stage has nothing to read, the injection block gates all
residual contributions of masked memory to zero, so the memory-free
\method\ forward pass is recovered exactly.

\subsection{Fine-Tuning Details}
\label{app:finetune_hparams}

\begin{table*}[t]
\centering
\caption{\textbf{Per-benchmark fine-tuning configuration.}
Fine-tuning is two-phase: during the initial freeze epochs the
PaliGemma backbone is frozen and only the modules outside it are
trained (the convex-upsampling modules, the action heads, and the
memory modules); the backbone is then unfrozen for the remaining
epochs, and the learning-rate warmup applies per phase.}
\label{tab:hyperparams}
\footnotesize
\setlength{\tabcolsep}{6pt}
\setlength{\aboverulesep}{0pt}
\setlength{\belowrulesep}{0pt}
\renewcommand{\arraystretch}{1.2}
\begin{tabular}{lccccc}
\toprule
\textbf{Setting} & \textbf{RLBench} & \textbf{COLOSSEUM} & \textbf{GemBench} & \textbf{RMBench} & \textbf{MemoryBench} \\
\midrule
Epochs / freeze epochs & 130 / 4 & 200 / 4 & 200 / 2 & $\sim$320 (per task) / 5 & 160 / 20 \\
Warmup steps (per phase) & 1,500 & 1,500 & 1,000 & 1,000 & 1,000 \\
Optimizer & \multicolumn{5}{c}{AdamW, $(\beta_1, \beta_2) = (0.9, 0.95)$} \\
Learning rate & $8{\times}10^{-5}$ & $8{\times}10^{-5}$ & $5{\times}10^{-5}$ & $5{\times}10^{-5}$ & $5{\times}10^{-5}$ \\
Weight decay & $10^{-2}$ & $10^{-2}$ & $10^{-3}$ & $10^{-3}$ & $10^{-3}$ \\
Batch size per GPU & 4 & 4 & 4 & 4 & 4 \\
GPUs & 32 & 32 & 32 & 8 & 8 \\
\midrule
SE(3) aug.\ (trans.\ / yaw) & 0.125 / $45^{\circ}$ & 0.125 / $45^{\circ}$ & 0.125 / $45^{\circ}$ & 0.03 / $30^{\circ}$ & 0.125 / $45^{\circ}$ \\
Stage-2 zoom jitter & 0.05 & 0.05 & 0.05 & 0.005 & 0.05 \\
\midrule
Slot budget $K$ & 2 & 2 & 2 & 12 & 2 \\
Neighboring keyframes $n$ & 2 & 2 & 2 & 2 & 2 \\
Gate $\lambda_{\mathrm{check}}$ / pos.\ wt.\ / thr. & -- & -- & -- & 1.0 / 5.5 / 0.5 & -- \\
\midrule
Rotation head & \multicolumn{5}{c}{continuous 6D~\cite{zhou2019continuity}} \\
Collision head & on & on & off & off & off \\
Arms & 1 & 1 & 1 & 2 & 1 \\
Demonstrations per task & 100 & 100 & 100 (per variation) & 50 & 100 \\
\midrule
Input cameras (resolution) & 4 ($128^2$) & 4 ($128^2$) & 4 ($256^2$) & 4 ($224^2$) & 4 ($128^2$) \\
Ortho.\ window scale & 2.0 & 2.0 & 2.0 & 0.8 & 2.0 \\
Splat radius (coarse / fine) & 0.012 / 0.012 & 0.012 / 0.012 & 0.012 / 0.012 & 0.004 / 0.012 & 0.012 / 0.012 \\
\bottomrule
\end{tabular}
\end{table*}

Table~\ref{tab:hyperparams} lists the per-benchmark fine-tuning
configuration; the paragraphs below cover the settings the table
cannot express.

\paragraph{Two-phase schedule}
During the initial freeze epochs of Table~\ref{tab:hyperparams}, the
PaliGemma backbone is frozen and gradients reach only the modules
outside it: the convex-upsampling modules, the MLP action heads, and
the memory injection blocks and sub-goal gate.
This first phase lets the modules that the pre-training of
Appendix~\ref{app:pretrain_details} does not cover adapt to the
manipulation data before the backbone is touched.
In the second phase the backbone is unfrozen and trained jointly with
these modules, except for the SigLIP vision encoder and the
language-token embedding, which remain frozen throughout fine-tuning.
The optimizer is re-initialized at the phase boundary, so the warmup
steps of Table~\ref{tab:hyperparams} apply to each phase.

\paragraph{Memory-specific settings}
Whenever memory is enabled, random in-plane 2D image augmentation is
disabled and the workspace cube is centered on fixed scene bounds
rather than on the per-frame cloud mean, since either perturbation
would break the pixel correspondence between the current observation
and the cached memory tokens.
The SE(3) augmentation of Table~\ref{tab:hyperparams}, whose
translation is a fraction of the workspace extent and whose rotation
perturbs yaw only, is instead applied jointly to the current, anchor,
and history point clouds (Sec.~\ref{sec:memmethod:training}); the
stage-2 zoom jitter perturbs the ground-truth waypoint the fine stage
zooms to during training.

\paragraph{Slot budgets}
On RMBench the budget $K{=}12$ holds the two neighboring keyframes and
up to ten sub-goal slots.
Every executed keyframe occupies a neighboring slot regardless of the
gate, and a gated keyframe enters a sub-goal slot only when it leaves
the neighboring window two steps later; when the sub-goal slots are
full, the oldest is evicted.
The other benchmarks carry no sub-goal annotations, so the gate is
disabled and the budget holds only the two neighboring keyframes,
$K{=}2$; the temporal memory there reduces to the neighboring
keyframes and the initial anchor, which suits the shorter horizons of
these tasks.

\paragraph{Rendering}
Every point cloud is rendered into three $224 \times 224$ orthographic
views regardless of the sensor resolution listed in
Table~\ref{tab:hyperparams}.
The orthographic window scale sets the extent of the rendered viewport
relative to the workspace, with values below one zooming in.

\paragraph{Ablation configurations}
Among the design-ablation rows of Table~\ref{tab:rlbench}
(Sec.~\ref{sec:exp:ablation}), \emph{w/ discretized rotation} replaces
the continuous 6D rotation head with per-axis Euler angles quantized
into $5^{\circ}$ bins and supervised with cross-entropy, as in our
conference version.
Beyond its resolution ceiling, this representation is ill-conditioned
near the gimbal-lock singularities of the Euler decomposition, where
the roll and yaw axes degenerate and orientations that are close in
$SO(3)$ may fall into distant bins, making the per-axis cross-entropy
targets discontinuous.
The continuous 6D parameterization is free of such singularities.
\emph{w/ 3D position input} adds a 3D convolutional module that encodes
per-pixel 3D positions and fuses them with the 2D image features fed to
the backbone.
\emph{w/o heatmap decoding} replaces the convex-upsampling module, of
309M parameters, with a similarly sized Transformer decoder of 303M
parameters that regresses the target positions directly under an MSE
loss, leaving all other modules unchanged.

\subsection{Training Data Preparation}
\label{app:keyframe_selection}

\paragraph{Keyframe selection}
Demonstrations are converted into consecutive-keyframe training
transitions with the keyframe-selection strategy of
PerAct~\cite{shridhar2023perceiver} in all single-arm experiments: a
time step is labeled a keyframe if the robot is stationary, if the
gripper state changes, or if it is the final step of the episode.
RMBench demonstrations instead use a bimanual variant of this
heuristic, which keeps the last frame of every segment in which both
arms are still.

\paragraph{Sub-goal labels}
\label{app:gate_labels}
The sub-goal gate of Sec.~\ref{sec:memmethod:gate} is supervised by
labels derived from the RMBench demonstrations, whose keyframes carry
per-segment language annotations.
The last keyframe of each language segment marks the completion of a
sub-goal and is labeled positive; all other keyframes are negatives.
Repeated identical language segments are kept separate rather than
merged by text identity, so every repetition of an instruction, such
as each press in \emph{Press Button}, contributes its own sub-goal
frame.
The resulting positives cover 9--15\% of keyframes depending on the
task, which the binary cross-entropy compensates with the
positive-class weight of Table~\ref{tab:hyperparams}.

\subsection{Evaluation Protocol}
\label{app:eval_protocol}

This appendix states the training data, trial counts, step budgets,
and reporting statistics used for each benchmark.

\begin{figure*}[t]
  \centering
  \includegraphics[width=0.92\textwidth]{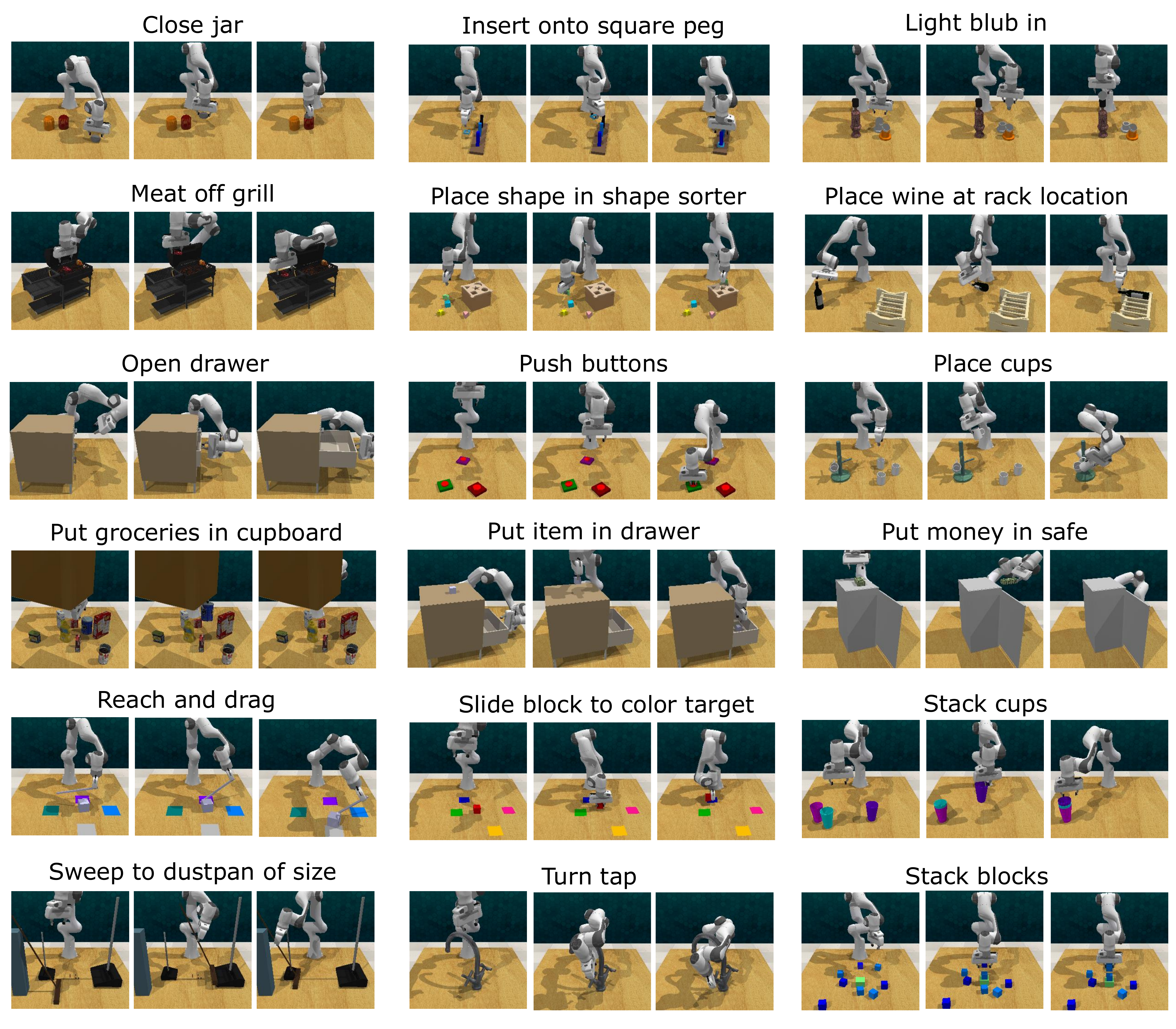}
  \caption{\textbf{The 18 RLBench Tasks.}
  Visualization of the 18 RLBench~\cite{james2020rlbench} tasks used in
  Sec.~\ref{sec:exp:rlbench}.}
  \label{fig:RLBench_VIS}
\end{figure*}

\paragraph{RLBench}
Training uses the 100 demonstrations per task provided by the
benchmark, over the 18 tasks of Fig.~\ref{fig:RLBench_VIS}.
Each configuration is evaluated over 25 episodes per task under a
25-keyframe-step budget, except \emph{Place Cups} and \emph{Stack
Blocks}, which receive 35 steps.
We report means and standard deviations over five independent
evaluation runs, except for the two design-ablation variants that
replace the heatmap head or inject 3D position input, which are
evaluated over three runs.

\paragraph{COLOSSEUM}
Policies are trained on the unperturbed RLBench data of the 20
benchmark tasks, 100 demonstrations per task, and evaluated under the
14 settings of Table~\ref{tab:colosseum}, visualized in
Fig.~\ref{fig:vis_colosseum}, with 25 trials per task and setting.
\method, \memmethod, and RVT-2 are our own training and evaluation
runs, reported as mean and variance over three test repetitions
(Tables~\ref{tab:results_bridgevla},
\ref{tab:results_bridgevla_plus},
and~\ref{tab:results_rvt2}); the remaining baseline numbers are quoted
from the benchmark release~\cite{pumacay2024colosseum}.

\paragraph{GemBench}
Policies are trained on the 16-task training split, 31 variations, and
evaluated on the 44 test tasks, 92 variations, of the four levels
L1--L4 shown in Fig.~\ref{fig:vis_gembench}.
Following the benchmark protocol~\cite{garcia2024towards}, \method\ and
\memmethod\ are each evaluated over five random seeds with 20 trials
per task variation, and Tables~\ref{tab:gembench}
and~\ref{tab:gembench_sota_cmpr_l1_detail}--\ref{tab:gembench_sota_cmpr_l4_detail}
report means over the five seeds.
Baseline numbers are quoted from~\cite{garcia2024towards}.

\begin{figure*}[t]
  \centering
  \includegraphics[width=\textwidth]{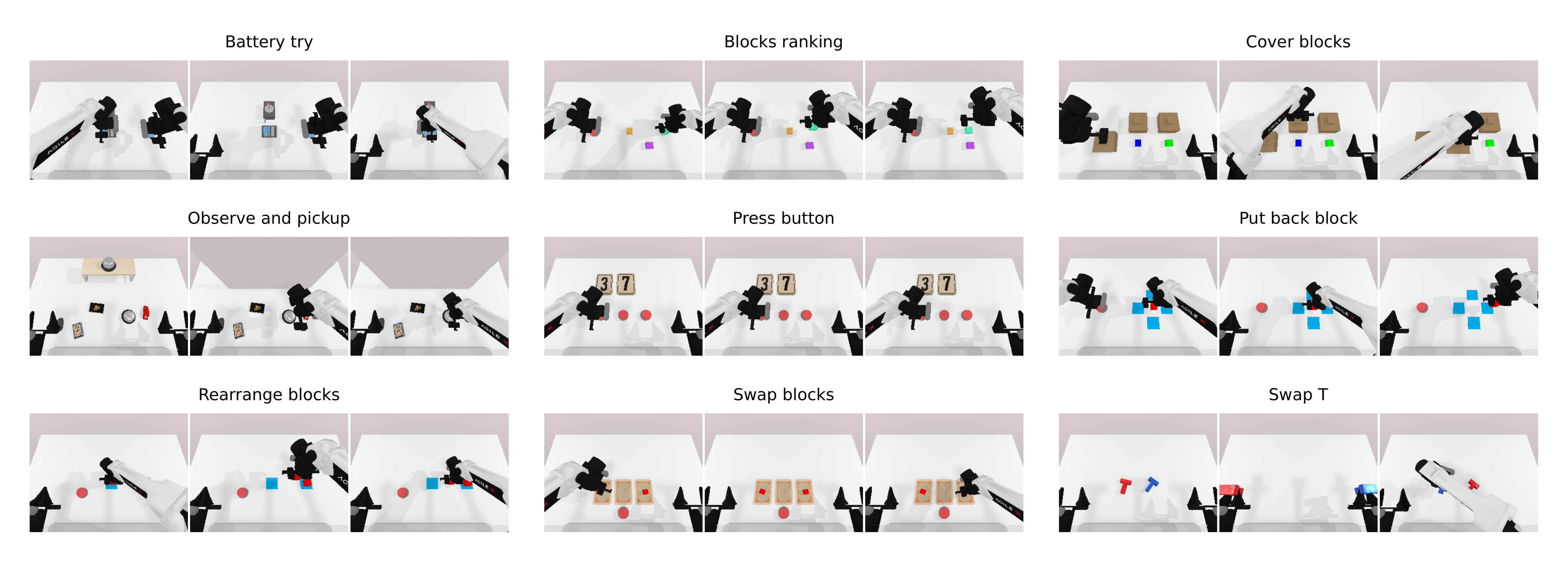}
  \caption{\textbf{The Nine RMBench Tasks.}
  One evaluation rollout per task of RMBench~\cite{chen2026rmbench},
  shown as three frames in temporal order; the dual-arm tasks span the
  short-term $M(1)$ and long-term $M(n)$ memory regimes.}
  \label{fig:vis_rmbench}
\end{figure*}

\paragraph{RMBench}
The benchmark is built within RoboTwin~2.0~\cite{chen2025robotwin2} and
simulated in SAPIEN~\cite{xiang2020sapien}; its nine dual-arm tasks are
shown in Fig.~\ref{fig:vis_rmbench}.
Following the benchmark protocol, policies are trained on 50 expert
demonstrations per task; \memmethod\ emits one action tuple per arm at
every keyframe step (Sec.~\ref{sec:bridgevla:problem}), and the
collision flag is dropped.
Every reported number is a single 100-episode evaluation under a
per-task keyframe-step limit scaled to the task horizon.
The numbers in Tables~\ref{tab:rmbench_main}
and~\ref{tab:rmbench_ablation} are obtained by training one model per
task and selecting the best-performing checkpoint of each task's training run.

\paragraph{MemoryBench}
Policies are trained on 100 demonstrations per task following the
protocol of~\cite{fang2025sam2act}.
One evaluation covers all nine task variants of the three tasks of
Fig.~\ref{fig:vis_memorybench} under a 25-step budget, and we report mean$\pm$std over five evaluation seeds.

\paragraph{Real robot}
Each of the 13 Franka tasks of Figs.~\ref{fig:basic_task1}
and~\ref{fig:basic_task2} is trained with 10 kinesthetic-teaching
demonstrations, reduced to 3 in the low-data variant of
Table~\ref{tab:app:real_episode}, and every method is evaluated over 10
trials per task in the Basic setting.
Each test scene is photographed and manually aligned across methods.
A single multi-task model is trained jointly on all tasks, and the
same checkpoint is evaluated on every task.
The six generalization settings compare \method\ and RVT-2, the two
methods that perform well in the Basic setting; their definitions are
given in Appendix~\ref{app:real_settings_vis} and the four
visual-disturbance ones are visualized in Fig.~\ref{fig:real_settings}.

The Dobot suite of Appendix~\ref{app:real_dobot} follows the same
protocol at the granularity of the language instruction rather than the
task: its five tasks, three memory-dependent and two memory-free,
comprise seven instructions, each trained with 10 kinesthetic-teaching
demonstrations, for 70 demonstrations in total.
All methods share this training set, and \method\ and \memmethod\ are
each trained jointly on all seven instructions as a single model whose
one checkpoint is evaluated on every instruction, as on the Franka
platform.
Every instruction is evaluated
over 10 trials in each of the Basic, Distractor, Background, Height,
and Lighting settings, the last four illustrated in
Fig.~\ref{fig:dobot_settings} and defined as in the Franka suite
(Appendix~\ref{app:real_settings_vis}).
The comparison here runs across all five
settings and against two baselines, the memory-free \method\ and the
memory-augmented SAM2Act+~\cite{fang2025sam2act}, so that the effect of
memory is separated from that of the backbone: \method\ shares
\memmethod's backbone but has no memory, whereas SAM2Act+ has memory
but a different backbone and retrieval scheme.
Figs.~\ref{fig:dobot_rollouts1} and~\ref{fig:dobot_rollouts2} show
\memmethod\ rollouts on the memory-dependent and memory-free
instructions, respectively.

\subsection{Computational Cost}
\label{app:compute_resources}

The running times below are those of the runs behind the reported
results.
The 2D-heatmap pre-training of Appendix~\ref{app:pretrain_details}
takes about 2 hours on 8 NVIDIA A100 GPUs.
For fine-tuning, \method\ and \memmethod\ train on 32 H20 GPUs on
RLBench, COLOSSEUM, and GemBench, as do the ablation rows of
Table~\ref{tab:rlbench}; RMBench trains one model per task on 8 H20
GPUs each, and MemoryBench trains on 8 A100 GPUs.
Real-world fine-tuning takes about 1.5 hours on 8 A100 GPUs.

Evaluation uses a single GPU per run: RMBench, GemBench, MemoryBench,
and COLOSSEUM are evaluated on one A100, and RLBench on one H20.
For real-world deployment, both models run on a machine with a single
NVIDIA RTX 4090 GPU.
Averaged over 100 trials, the end-to-end latency from point-cloud input
to action output is 0.35 seconds per prediction step for \method\ and
0.57 seconds for the full \memmethod\
(Sec.~\ref{sec:memmethod:training}).
Both figures are small relative to the remainder of the control loop:
in real-world deployment, transmitting the multi-camera observations
and physically executing the planned motion between keyframes dominate
the total time per step.

\paragraph{Memory overhead}
The totals of Sec.~\ref{sec:memmethod:training} decompose as follows.
Each of the three memory injection blocks holds 83.95M parameters and
the sub-goal gate a further 17.91M, for 269.77M in total, or 9.2\% of
the 2.92B-parameter backbone.
A cached memory entry is a single bf16 coarse-stage token grid of
3.0\,MiB, so a full memory at the RMBench budget of $K{=}12$ frames
plus the anchor occupies 39\,MiB.
Because entries are stored already encoded, the cache is the only
stored quantity that grows with $K$; the cross-attention cost of the
injection block also grows linearly in $K$ but remains negligible next
to a backbone forward pass, and the number of forward passes per step
is independent of $K$.
Per step, the memory-free policy runs two backbone forwards, coarse and
fine, and the full \memmethod\ three, the additional pass encoding the
fine-stage geometric reference.
Dual-arm deployments, whose fine stage runs once per arm on a shared
coarse trunk (Sec.~\ref{sec:memmethod:bimanual}), accordingly run five.

\renewcommand{\dbltopfraction}{0.6}

\subsection{Per-Task Results on COLOSSEUM}
\label{app:colosseum_results}

COLOSSEUM and GemBench, covered here and in
Appendix~\ref{app:gembench_results}, hold the training data fixed and
shift the test environment away from it in appearance, objects, and
instructions.
Because both suites probe per-frame perception rather than history, the
base policy \method\ carries the comparison against prior work;
\memmethod\ is reported alongside it to verify that the memory
extension preserves this robustness.

\begin{figure*}[t]
  \centering
  \includegraphics[width=0.7\textwidth]{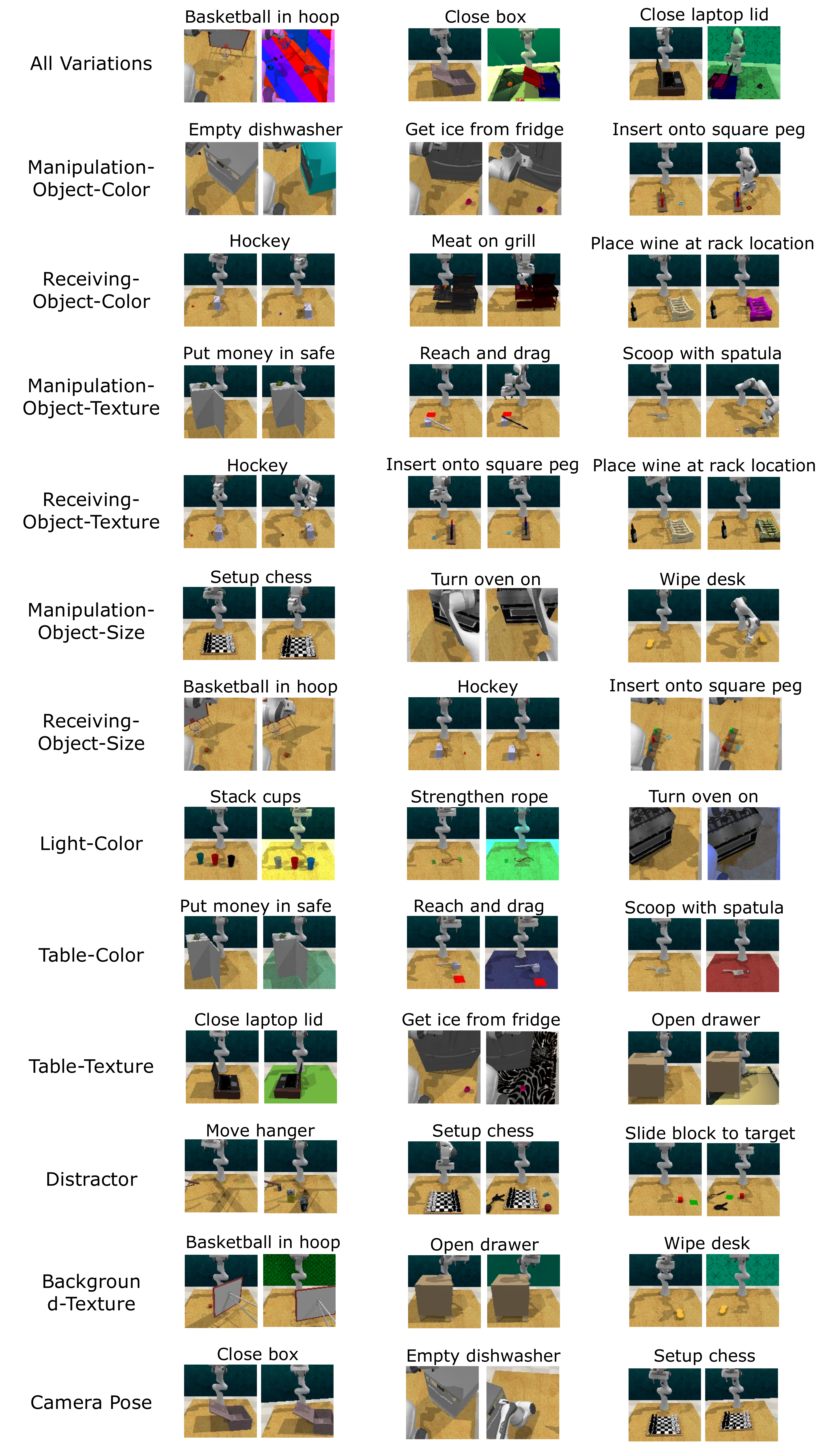}
  \caption{\textbf{Perturbations in COLOSSEUM~\cite{pumacay2024colosseum}.}
  All perturbation axes are shown except the original-RLBench variation
  setting.}
  \label{fig:vis_colosseum}
\end{figure*}

On COLOSSEUM~\cite{pumacay2024colosseum} (settings and protocol in
Appendix~\ref{app:eval_protocol}, perturbations visualized in
Fig.~\ref{fig:vis_colosseum}), we compare against
R3M-MLP~\cite{nair2022r3m} and MVP-MLP~\cite{xiao2022masked}, which
pair pre-trained 2D encoders with MLP action heads, and against the 3D
policies PerAct~\cite{shridhar2023perceiver}, RVT~\cite{goyal2023rvt},
and RVT-2~\cite{goyal2024rvt}.

Beyond the averages reported in Sec.~\ref{sec:exp:additional}, \method\
ranks best among prior methods in 13 of the 14 settings
(Table~\ref{tab:colosseum}), and how that margin is distributed
supports the alignment argument: it is widest under appearance-level
shifts, with a lead of 11 to 15 points on table texture, table color,
light color, and receptacle texture, which is exactly the nuisance
variation a VLM's 2D pre-training has seen in abundance and a policy
trained from scratch has not.
The one setting in which \method\ trails is \emph{Distractor}, at
51.8\% against 60.8\% for RVT-2.
\memmethod\ improves markedly on precisely the two settings that are
hardest for \method, reaching 38.9\% against 18.7\% under \emph{All
Perturbations} and 61.6\% against 51.8\% under \emph{Distractor}, and
between them the two variants rank first in every one of the 14
settings.
Tables~\ref{tab:results_bridgevla},
\ref{tab:results_bridgevla_plus}, and~\ref{tab:results_rvt2} break the
comparison down to the task level.

\begin{table*}[t]
\centering
\caption{\textbf{Per-Task Results of \method\ on COLOSSEUM.}
Success rates (\%) under each COLOSSEUM
perturbation~\cite{pumacay2024colosseum}, mean$\pm$variance over three
evaluation repetitions;
``--'' marks perturbation--task combinations the benchmark does not
define.}
\label{tab:results_bridgevla}
\footnotesize
\setlength{\tabcolsep}{4pt}
\setlength{\aboverulesep}{0pt}
\setlength{\belowrulesep}{0pt}
\renewcommand{\arraystretch}{1.25}
\resizebox{\fitwidth}{!}{%
\begin{tabular}{l*{15}{c}}
\toprule
\textbf{Task} & \rotatebox[origin=c]{90}{\textbf{Original}} & \rotatebox[origin=c]{90}{\textbf{All Perturbations}} & \rotatebox[origin=c]{90}{\textbf{MO-COLOR}} & \rotatebox[origin=c]{90}{\textbf{RO-COLOR}} & \rotatebox[origin=c]{90}{\textbf{MO-TEXTURE}} & \rotatebox[origin=c]{90}{\textbf{RO-TEXTURE}} & \rotatebox[origin=c]{90}{\textbf{MO-SIZE}} & \rotatebox[origin=c]{90}{\textbf{RO-SIZE}} & \rotatebox[origin=c]{90}{\textbf{Light Color}} & \rotatebox[origin=c]{90}{\textbf{Table Color}} & \rotatebox[origin=c]{90}{\textbf{Table Texture}} & \rotatebox[origin=c]{90}{\textbf{Distractor}} & \rotatebox[origin=c]{90}{\textbf{Background Texture}} & \rotatebox[origin=c]{90}{\textbf{RLBench}} & \rotatebox[origin=c]{90}{\textbf{Camera Pose}} \\
\midrule
basketball\_in\_hoop & 100.0$\pm$0.0 & 4.0$\pm$3.3 & 94.7$\pm$1.9 & 96.0$\pm$0.0 & 84.0$\pm$5.7 & -- & 100.0$\pm$0.0 & 68.0$\pm$0.0 & 100.0$\pm$0.0 & 100.0$\pm$0.0 & 100.0$\pm$0.0 & 37.3$\pm$1.9 & 100.0$\pm$0.0 & 100.0$\pm$0.0 & 100.0$\pm$0.0 \\
close\_box & 100.0$\pm$0.0 & 72.0$\pm$0.0 & 94.7$\pm$1.9 & -- & -- & -- & 93.3$\pm$1.9 & -- & 100.0$\pm$0.0 & 100.0$\pm$0.0 & 98.7$\pm$1.9 & 98.7$\pm$1.9 & 100.0$\pm$0.0 & 97.3$\pm$1.9 & 100.0$\pm$0.0 \\
close\_laptop\_lid & 100.0$\pm$0.0 & 11.1$\pm$15.7 & 82.7$\pm$3.8 & -- & -- & -- & 67.9$\pm$14.6 & -- & 89.3$\pm$8.2 & 92.0$\pm$0.0 & 97.3$\pm$3.8 & 82.7$\pm$6.8 & 96.0$\pm$3.3 & 100.0$\pm$0.0 & 96.0$\pm$0.0 \\
empty\_dishwasher & 0.0$\pm$0.0 & 0.0$\pm$0.0 & 1.3$\pm$1.9 & 1.3$\pm$1.9 & -- & 1.3$\pm$1.9 & 4.0$\pm$3.3 & 0.0$\pm$0.0 & 0.0$\pm$0.0 & 0.0$\pm$0.0 & 0.0$\pm$0.0 & 0.0$\pm$0.0 & 1.3$\pm$1.9 & 1.3$\pm$1.9 & 0.0$\pm$0.0 \\
get\_ice\_from\_fridge & 94.7$\pm$1.9 & 5.3$\pm$1.9 & 86.7$\pm$1.9 & 90.7$\pm$7.5 & 90.7$\pm$5.0 & -- & 84.0$\pm$3.3 & 73.3$\pm$1.9 & 96.0$\pm$3.3 & 98.7$\pm$1.9 & 89.3$\pm$7.5 & 56.0$\pm$8.6 & 94.7$\pm$1.9 & 96.0$\pm$3.3 & 98.7$\pm$1.9 \\
hockey & 57.3$\pm$5.0 & 9.3$\pm$3.8 & 44.0$\pm$6.5 & 50.7$\pm$8.2 & -- & 50.7$\pm$13.2 & 46.7$\pm$8.2 & 65.3$\pm$5.0 & 45.3$\pm$1.9 & 64.0$\pm$8.6 & 53.3$\pm$1.9 & 20.0$\pm$3.3 & 56.0$\pm$5.7 & 49.3$\pm$5.0 & 50.7$\pm$5.0 \\
insert\_onto\_square\_peg & 93.3$\pm$3.8 & 23.3$\pm$2.4 & 52.0$\pm$3.3 & 94.7$\pm$1.9 & -- & 76.0$\pm$8.6 & 85.3$\pm$3.8 & 70.7$\pm$3.8 & 84.0$\pm$0.0 & 88.0$\pm$3.3 & 88.0$\pm$3.3 & 44.0$\pm$11.8 & 86.7$\pm$1.9 & 77.3$\pm$5.0 & 96.0$\pm$0.0 \\
meat\_on\_grill & 96.0$\pm$0.0 & 9.3$\pm$1.9 & 32.0$\pm$0.0 & 88.0$\pm$5.7 & -- & -- & 100.0$\pm$0.0 & -- & 100.0$\pm$0.0 & 92.0$\pm$6.5 & 90.7$\pm$1.9 & 98.7$\pm$1.9 & 97.3$\pm$1.9 & 100.0$\pm$0.0 & 100.0$\pm$0.0 \\
move\_hanger & 37.3$\pm$3.8 & 2.7$\pm$3.8 & 26.7$\pm$3.8 & 46.7$\pm$3.8 & -- & -- & -- & -- & 52.0$\pm$0.0 & 84.0$\pm$0.0 & 52.0$\pm$5.7 & 52.0$\pm$5.7 & 33.3$\pm$5.0 & 42.7$\pm$1.9 & 24.0$\pm$0.0 \\
open\_drawer & 96.0$\pm$0.0 & 60.0$\pm$3.3 & 97.3$\pm$1.9 & -- & -- & -- & 90.7$\pm$1.9 & -- & 88.0$\pm$3.3 & 93.3$\pm$1.9 & 100.0$\pm$0.0 & 90.7$\pm$1.9 & 100.0$\pm$0.0 & 94.7$\pm$1.9 & 96.0$\pm$0.0 \\
place\_wine\_at\_rack\_location & 88.0$\pm$5.7 & 17.3$\pm$13.6 & 82.7$\pm$5.0 & 89.3$\pm$7.5 & -- & 92.0$\pm$6.5 & 93.3$\pm$3.8 & 90.7$\pm$3.8 & 90.7$\pm$5.0 & 97.3$\pm$1.9 & 88.0$\pm$3.3 & 74.7$\pm$3.8 & 90.7$\pm$6.8 & 92.0$\pm$3.3 & 92.0$\pm$8.6 \\
put\_money\_in\_safe & 94.7$\pm$1.9 & 6.7$\pm$5.0 & 78.7$\pm$1.9 & 74.7$\pm$1.9 & 81.3$\pm$6.8 & 89.3$\pm$5.0 & 92.0$\pm$3.3 & -- & 37.3$\pm$12.4 & 84.0$\pm$3.3 & 84.0$\pm$3.3 & 84.0$\pm$3.3 & 89.3$\pm$1.9 & 86.7$\pm$8.2 & 86.7$\pm$1.9 \\
reach\_and\_drag & 100.0$\pm$0.0 & 0.0$\pm$0.0 & 89.3$\pm$3.8 & 96.0$\pm$0.0 & 94.7$\pm$5.0 & 84.0$\pm$5.7 & 94.7$\pm$1.9 & 38.7$\pm$5.0 & 92.0$\pm$3.3 & 88.0$\pm$5.7 & 78.7$\pm$3.8 & 28.0$\pm$8.6 & 100.0$\pm$0.0 & 100.0$\pm$0.0 & 94.7$\pm$3.8 \\
scoop\_with\_spatula & 96.0$\pm$3.3 & 6.7$\pm$1.9 & 94.7$\pm$1.9 & 93.3$\pm$1.9 & 85.3$\pm$3.8 & 85.3$\pm$3.8 & 78.7$\pm$3.8 & 86.7$\pm$5.0 & 90.7$\pm$1.9 & 88.0$\pm$6.5 & 77.3$\pm$1.9 & 20.0$\pm$5.7 & 90.7$\pm$6.8 & 89.3$\pm$1.9 & 93.3$\pm$1.9 \\
setup\_chess & 10.7$\pm$1.9 & 0.0$\pm$0.0 & 1.3$\pm$1.9 & 8.0$\pm$0.0 & 8.0$\pm$3.3 & -- & 13.3$\pm$1.9 & -- & 12.0$\pm$5.7 & 21.3$\pm$8.2 & 13.3$\pm$3.8 & 5.3$\pm$1.9 & 20.0$\pm$5.7 & 16.0$\pm$5.7 & 4.0$\pm$3.3 \\
slide\_block\_to\_target & 100.0$\pm$0.0 & 24.0$\pm$3.3 & 74.7$\pm$1.9 & -- & 92.0$\pm$3.3 & -- & -- & -- & 100.0$\pm$0.0 & 100.0$\pm$0.0 & 98.7$\pm$1.9 & 84.0$\pm$9.8 & 100.0$\pm$0.0 & 100.0$\pm$0.0 & 100.0$\pm$0.0 \\
stack\_cups & 58.7$\pm$3.8 & 29.3$\pm$1.9 & 66.7$\pm$1.9 & -- & 50.7$\pm$1.9 & -- & 44.0$\pm$3.3 & -- & 62.7$\pm$1.9 & 64.0$\pm$3.3 & 65.3$\pm$8.2 & 26.7$\pm$7.5 & 73.3$\pm$8.2 & 64.0$\pm$14.2 & 72.0$\pm$8.6 \\
straighten\_rope & 61.3$\pm$6.8 & 8.0$\pm$5.7 & 16.0$\pm$5.7 & -- & 48.0$\pm$3.3 & -- & -- & -- & 61.3$\pm$9.4 & 65.3$\pm$1.9 & 54.7$\pm$8.2 & 37.3$\pm$5.0 & 70.7$\pm$8.2 & 66.7$\pm$7.5 & 72.0$\pm$6.5 \\
turn\_oven\_on & 93.3$\pm$1.9 & 85.3$\pm$3.8 & 94.7$\pm$3.8 & -- & -- & -- & 90.7$\pm$1.9 & -- & 93.3$\pm$3.8 & 94.7$\pm$7.5 & 96.0$\pm$3.3 & 96.0$\pm$3.3 & 96.0$\pm$0.0 & 88.0$\pm$3.3 & 100.0$\pm$0.0 \\
wipe\_desk & 0.0$\pm$0.0 & 0.0$\pm$0.0 & 0.0$\pm$0.0 & 0.0$\pm$0.0 & 0.0$\pm$0.0 & -- & 0.0$\pm$0.0 & -- & 0.0$\pm$0.0 & 0.0$\pm$0.0 & 0.0$\pm$0.0 & 0.0$\pm$0.0 & 0.0$\pm$0.0 & 0.0$\pm$0.0 & 0.0$\pm$0.0 \\
\midrule
Task Mean & 73.9$\pm$0.7 & 18.7$\pm$2.2 & 60.5$\pm$1.1 & 63.8$\pm$0.1 & 63.5$\pm$1.5 & 68.4$\pm$3.3 & 69.3$\pm$1.0 & 61.7$\pm$0.8 & 69.7$\pm$1.2 & 75.7$\pm$0.9 & 71.3$\pm$0.7 & 51.8$\pm$1.5 & 74.8$\pm$1.0 & 73.1$\pm$0.2 & 73.8$\pm$0.3 \\
\bottomrule
\end{tabular}%
}
\end{table*}

\begin{table*}[t]
\centering
\caption{\textbf{Per-Task Results of \memmethod\ on COLOSSEUM.}
Success rates (\%) under each COLOSSEUM
perturbation~\cite{pumacay2024colosseum}, mean$\pm$variance over three
evaluation repetitions,
under the identical protocol as Table~\ref{tab:results_bridgevla};
``--'' marks perturbation--task combinations the benchmark does not
define.}
\label{tab:results_bridgevla_plus}
\footnotesize
\setlength{\tabcolsep}{4pt}
\setlength{\aboverulesep}{0pt}
\setlength{\belowrulesep}{0pt}
\renewcommand{\arraystretch}{1.25}
\resizebox{\fitwidth}{!}{%
\begin{tabular}{l*{15}{c}}
\toprule
\textbf{Task} & \rotatebox[origin=c]{90}{\textbf{Original}} & \rotatebox[origin=c]{90}{\textbf{All Perturbations}} & \rotatebox[origin=c]{90}{\textbf{MO-COLOR}} & \rotatebox[origin=c]{90}{\textbf{RO-COLOR}} & \rotatebox[origin=c]{90}{\textbf{MO-TEXTURE}} & \rotatebox[origin=c]{90}{\textbf{RO-TEXTURE}} & \rotatebox[origin=c]{90}{\textbf{MO-SIZE}} & \rotatebox[origin=c]{90}{\textbf{RO-SIZE}} & \rotatebox[origin=c]{90}{\textbf{Light Color}} & \rotatebox[origin=c]{90}{\textbf{Table Color}} & \rotatebox[origin=c]{90}{\textbf{Table Texture}} & \rotatebox[origin=c]{90}{\textbf{Distractor}} & \rotatebox[origin=c]{90}{\textbf{Background Texture}} & \rotatebox[origin=c]{90}{\textbf{RLBench}} & \rotatebox[origin=c]{90}{\textbf{Camera Pose}} \\
\midrule
basketball\_in\_hoop & 100.0$\pm$0.0 & 41.3$\pm$3.8 & 96.0$\pm$0.0 & 100.0$\pm$0.0 & 94.7$\pm$1.9 & -- & 100.0$\pm$0.0 & 86.7$\pm$1.9 & 100.0$\pm$0.0 & 100.0$\pm$0.0 & 100.0$\pm$0.0 & 97.3$\pm$1.9 & 100.0$\pm$0.0 & 100.0$\pm$0.0 & 100.0$\pm$0.0 \\
close\_box & 93.3$\pm$1.9 & 84.0$\pm$0.0 & 94.7$\pm$1.9 & -- & -- & -- & 96.0$\pm$0.0 & -- & 96.0$\pm$3.3 & 96.0$\pm$0.0 & 97.3$\pm$1.9 & 94.7$\pm$5.0 & 98.7$\pm$1.9 & 98.7$\pm$1.9 & 96.0$\pm$0.0 \\
close\_laptop\_lid & 100.0$\pm$0.0 & 82.7$\pm$1.9 & 96.0$\pm$0.0 & -- & -- & -- & 100.0$\pm$0.0 & -- & 100.0$\pm$0.0 & 96.0$\pm$0.0 & 92.0$\pm$0.0 & 96.0$\pm$0.0 & 92.0$\pm$0.0 & 100.0$\pm$0.0 & 100.0$\pm$0.0 \\
empty\_dishwasher & 10.7$\pm$1.9 & 16.0$\pm$0.0 & 12.0$\pm$3.3 & 25.3$\pm$5.0 & -- & 20.0$\pm$5.7 & 37.3$\pm$5.0 & 20.0$\pm$8.6 & 4.0$\pm$3.3 & 10.7$\pm$1.9 & 12.0$\pm$0.0 & 14.7$\pm$1.9 & 14.7$\pm$1.9 & 6.7$\pm$5.0 & 21.3$\pm$5.0 \\
get\_ice\_from\_fridge & 96.0$\pm$3.3 & 22.7$\pm$6.8 & 88.0$\pm$0.0 & 98.7$\pm$1.9 & 93.3$\pm$1.9 & -- & 94.7$\pm$1.9 & 90.7$\pm$1.9 & 97.3$\pm$1.9 & 98.7$\pm$1.9 & 98.7$\pm$1.9 & 93.3$\pm$1.9 & 100.0$\pm$0.0 & 96.0$\pm$0.0 & 97.3$\pm$3.8 \\
hockey & 38.7$\pm$7.5 & 2.7$\pm$1.9 & 48.0$\pm$5.7 & 37.3$\pm$1.9 & -- & 26.7$\pm$5.0 & 22.7$\pm$5.0 & 28.0$\pm$3.3 & 22.7$\pm$3.8 & 37.3$\pm$6.8 & 26.7$\pm$5.0 & 16.0$\pm$0.0 & 26.7$\pm$1.9 & 29.3$\pm$1.9 & 33.3$\pm$5.0 \\
insert\_onto\_square\_peg & 24.0$\pm$6.5 & 58.7$\pm$5.0 & 18.7$\pm$8.2 & 41.3$\pm$1.9 & -- & 44.0$\pm$3.3 & 56.0$\pm$3.3 & 14.7$\pm$3.8 & 25.3$\pm$1.9 & 34.7$\pm$1.9 & 33.3$\pm$3.8 & 29.3$\pm$1.9 & 36.0$\pm$3.3 & 26.7$\pm$6.8 & 33.3$\pm$3.8 \\
meat\_on\_grill & 100.0$\pm$0.0 & 81.3$\pm$1.9 & 97.3$\pm$1.9 & 100.0$\pm$0.0 & -- & -- & 100.0$\pm$0.0 & -- & 100.0$\pm$0.0 & 100.0$\pm$0.0 & 100.0$\pm$0.0 & 100.0$\pm$0.0 & 100.0$\pm$0.0 & 100.0$\pm$0.0 & 100.0$\pm$0.0 \\
move\_hanger & 20.0$\pm$0.0 & 18.7$\pm$3.8 & 52.0$\pm$9.8 & 25.3$\pm$7.5 & -- & -- & -- & -- & 26.7$\pm$8.2 & 57.3$\pm$3.8 & 37.3$\pm$3.8 & 49.3$\pm$5.0 & 24.0$\pm$5.7 & 20.0$\pm$0.0 & 21.3$\pm$6.8 \\
open\_drawer & 100.0$\pm$0.0 & 69.3$\pm$1.9 & 100.0$\pm$0.0 & -- & -- & -- & 100.0$\pm$0.0 & -- & 100.0$\pm$0.0 & 100.0$\pm$0.0 & 100.0$\pm$0.0 & 74.7$\pm$5.0 & 100.0$\pm$0.0 & 100.0$\pm$0.0 & 100.0$\pm$0.0 \\
place\_wine\_at\_rack\_location & 96.0$\pm$3.3 & 77.3$\pm$5.0 & 98.7$\pm$1.9 & 100.0$\pm$0.0 & -- & 93.3$\pm$5.0 & 93.3$\pm$1.9 & 96.0$\pm$0.0 & 96.0$\pm$3.3 & 94.7$\pm$1.9 & 97.3$\pm$3.8 & 73.3$\pm$6.8 & 89.3$\pm$5.0 & 94.7$\pm$3.8 & 96.0$\pm$0.0 \\
put\_money\_in\_safe & 90.7$\pm$3.8 & 25.3$\pm$3.8 & 86.7$\pm$5.0 & 80.0$\pm$3.3 & 65.3$\pm$11.5 & 96.0$\pm$0.0 & 89.3$\pm$3.8 & -- & 81.3$\pm$5.0 & 86.7$\pm$1.9 & 77.3$\pm$6.8 & 78.7$\pm$5.0 & 89.3$\pm$3.8 & 78.7$\pm$3.8 & 73.3$\pm$5.0 \\
reach\_and\_drag & 86.7$\pm$6.8 & 5.3$\pm$1.9 & 85.3$\pm$5.0 & 94.7$\pm$5.0 & 82.7$\pm$3.8 & 88.0$\pm$0.0 & 92.0$\pm$3.3 & 81.3$\pm$1.9 & 86.7$\pm$5.0 & 80.0$\pm$5.7 & 86.7$\pm$1.9 & 77.3$\pm$5.0 & 93.3$\pm$1.9 & 89.3$\pm$1.9 & 76.0$\pm$5.7 \\
scoop\_with\_spatula & 93.3$\pm$1.9 & 21.3$\pm$1.9 & 93.3$\pm$1.9 & 88.0$\pm$3.3 & 93.3$\pm$1.9 & 90.7$\pm$5.0 & 78.7$\pm$1.9 & 78.7$\pm$5.0 & 88.0$\pm$6.5 & 96.0$\pm$3.3 & 90.7$\pm$1.9 & 62.7$\pm$6.8 & 92.0$\pm$3.3 & 93.3$\pm$1.9 & 89.3$\pm$1.9 \\
setup\_chess & 18.7$\pm$9.4 & 1.3$\pm$1.9 & 9.3$\pm$5.0 & 24.0$\pm$5.7 & 17.3$\pm$5.0 & -- & 25.3$\pm$5.0 & -- & 28.0$\pm$3.3 & 26.7$\pm$5.0 & 28.0$\pm$3.3 & 6.7$\pm$3.8 & 30.7$\pm$6.8 & 29.3$\pm$10.5 & 28.0$\pm$3.3 \\
slide\_block\_to\_target & 100.0$\pm$0.0 & 42.7$\pm$8.2 & 92.0$\pm$0.0 & -- & 96.0$\pm$0.0 & -- & -- & -- & 100.0$\pm$0.0 & 100.0$\pm$0.0 & 100.0$\pm$0.0 & 96.0$\pm$0.0 & 100.0$\pm$0.0 & 100.0$\pm$0.0 & 100.0$\pm$0.0 \\
stack\_cups & 41.3$\pm$5.0 & 5.3$\pm$1.9 & 36.0$\pm$5.7 & -- & 44.0$\pm$3.3 & -- & 37.3$\pm$1.9 & -- & 46.7$\pm$8.2 & 34.7$\pm$9.4 & 36.0$\pm$6.5 & 13.3$\pm$5.0 & 41.3$\pm$5.0 & 45.3$\pm$7.5 & 40.0$\pm$3.3 \\
straighten\_rope & 69.3$\pm$1.9 & 34.7$\pm$3.8 & 76.0$\pm$8.6 & -- & 70.7$\pm$6.8 & -- & -- & -- & 68.0$\pm$5.7 & 81.3$\pm$5.0 & 73.3$\pm$10.5 & 66.7$\pm$7.5 & 69.3$\pm$14.7 & 69.3$\pm$6.8 & 74.7$\pm$5.0 \\
turn\_oven\_on & 98.7$\pm$1.9 & 88.0$\pm$3.3 & 94.7$\pm$1.9 & -- & -- & -- & 93.3$\pm$1.9 & -- & 97.3$\pm$1.9 & 100.0$\pm$0.0 & 97.3$\pm$1.9 & 92.0$\pm$3.3 & 93.3$\pm$1.9 & 92.0$\pm$6.5 & 93.3$\pm$5.0 \\
wipe\_desk & 0.0$\pm$0.0 & 0.0$\pm$0.0 & 0.0$\pm$0.0 & 0.0$\pm$0.0 & 0.0$\pm$0.0 & -- & 0.0$\pm$0.0 & -- & 0.0$\pm$0.0 & 0.0$\pm$0.0 & 0.0$\pm$0.0 & 0.0$\pm$0.0 & 0.0$\pm$0.0 & 0.0$\pm$0.0 & 0.0$\pm$0.0 \\
\midrule
Task Mean & 68.9$\pm$0.5 & 38.9$\pm$0.8 & 68.7$\pm$0.7 & 62.7$\pm$0.6 & 65.7$\pm$0.4 & 65.5$\pm$1.2 & 71.5$\pm$0.3 & 62.0$\pm$0.7 & 68.2$\pm$1.0 & 71.5$\pm$0.3 & 69.2$\pm$0.7 & 61.6$\pm$0.5 & 69.5$\pm$1.2 & 68.5$\pm$0.6 & 68.7$\pm$0.7 \\
\bottomrule
\end{tabular}%
}
\end{table*}

\begin{table*}[t]
\centering
\caption{\textbf{Per-Task Results of RVT-2 on COLOSSEUM.}
Success rates (\%) of RVT-2~\cite{goyal2024rvt} under each COLOSSEUM
perturbation~\cite{pumacay2024colosseum}, trained and evaluated by us
under the protocol of Table~\ref{tab:results_bridgevla} (mean$\pm$variance
over three evaluation repetitions); ``--'' marks perturbation--task
combinations the benchmark does not define.}
\label{tab:results_rvt2}
\footnotesize
\setlength{\tabcolsep}{4pt}
\setlength{\aboverulesep}{0pt}
\setlength{\belowrulesep}{0pt}
\renewcommand{\arraystretch}{1.25}
\resizebox{\fitwidth}{!}{%
\begin{tabular}{l*{15}{c}}
\toprule
\textbf{Task} & \rotatebox[origin=c]{90}{\textbf{Original}} & \rotatebox[origin=c]{90}{\textbf{All Perturbations}} & \rotatebox[origin=c]{90}{\textbf{MO-COLOR}} & \rotatebox[origin=c]{90}{\textbf{RO-COLOR}} & \rotatebox[origin=c]{90}{\textbf{MO-TEXTURE}} & \rotatebox[origin=c]{90}{\textbf{RO-TEXTURE}} & \rotatebox[origin=c]{90}{\textbf{MO-SIZE}} & \rotatebox[origin=c]{90}{\textbf{RO-SIZE}} & \rotatebox[origin=c]{90}{\textbf{Light Color}} & \rotatebox[origin=c]{90}{\textbf{Table Color}} & \rotatebox[origin=c]{90}{\textbf{Table Texture}} & \rotatebox[origin=c]{90}{\textbf{Distractor}} & \rotatebox[origin=c]{90}{\textbf{Background Texture}} & \rotatebox[origin=c]{90}{\textbf{RLBench}} & \rotatebox[origin=c]{90}{\textbf{Camera Pose}} \\
\midrule
basketball\_in\_hoop & 100.0$\pm$0.0 & 10.0$\pm$2.0 & 99.0$\pm$1.7 & 94.0$\pm$2.0 & 97.0$\pm$1.7 & -- & 100.0$\pm$0.0 & 86.0$\pm$3.5 & 95.0$\pm$1.7 & 94.0$\pm$2.0 & 84.0$\pm$6.3 & 89.0$\pm$3.3 & 100.0$\pm$0.0 & 99.0$\pm$1.7 & 100.0$\pm$0.0 \\
close\_box & 93.0$\pm$4.4 & 36.0$\pm$8.5 & 70.0$\pm$6.6 & -- & -- & -- & 86.0$\pm$3.5 & -- & 99.0$\pm$1.7 & 97.0$\pm$1.7 & 91.0$\pm$4.4 & 93.0$\pm$3.3 & 97.0$\pm$1.7 & 94.0$\pm$2.0 & 99.0$\pm$1.7 \\
close\_laptop\_lid & 86.0$\pm$4.5 & 40.0$\pm$0.0 & 89.0$\pm$3.3 & -- & -- & -- & 62.0$\pm$2.0 & -- & 84.0$\pm$4.0 & 92.0$\pm$0.0 & 96.0$\pm$2.8 & 89.0$\pm$5.2 & 99.0$\pm$1.7 & 87.0$\pm$3.3 & 92.0$\pm$0.0 \\
empty\_dishwasher & 0.0$\pm$0.0 & 0.0$\pm$0.0 & 1.0$\pm$1.7 & 0.0$\pm$0.0 & -- & 0.0$\pm$0.0 & 0.0$\pm$0.0 & 0.0$\pm$0.0 & 0.0$\pm$0.0 & 0.0$\pm$0.0 & 0.0$\pm$0.0 & 0.0$\pm$0.0 & 0.0$\pm$0.0 & 0.0$\pm$0.0 & 0.0$\pm$0.0 \\
get\_ice\_from\_fridge & 95.0$\pm$1.7 & 11.0$\pm$4.4 & 88.0$\pm$5.7 & 77.0$\pm$5.2 & 89.0$\pm$1.7 & -- & 78.0$\pm$3.5 & 79.0$\pm$3.3 & 83.0$\pm$5.9 & 89.0$\pm$1.7 & 70.0$\pm$4.5 & 86.0$\pm$4.5 & 81.0$\pm$5.2 & 96.0$\pm$2.8 & 96.0$\pm$2.8 \\
hockey & 19.0$\pm$4.4 & 0.0$\pm$0.0 & 26.0$\pm$4.5 & 30.0$\pm$4.5 & -- & 40.0$\pm$4.9 & 24.0$\pm$2.8 & 13.0$\pm$3.3 & 12.0$\pm$8.5 & 15.0$\pm$3.3 & 9.0$\pm$3.3 & 10.0$\pm$6.0 & 14.0$\pm$2.0 & 17.0$\pm$3.3 & 19.0$\pm$3.3 \\
insert\_onto\_square\_peg & 31.0$\pm$3.3 & 0.0$\pm$0.0 & 13.0$\pm$1.7 & 35.0$\pm$9.5 & -- & 32.0$\pm$2.8 & 33.3$\pm$8.6 & 21.0$\pm$1.7 & 30.0$\pm$2.0 & 9.0$\pm$1.7 & 4.0$\pm$4.9 & 9.0$\pm$3.3 & 35.0$\pm$3.3 & 35.0$\pm$1.7 & 23.0$\pm$4.4 \\
meat\_on\_grill & 100.0$\pm$0.0 & 89.0$\pm$1.7 & 100.0$\pm$0.0 & 100.0$\pm$0.0 & -- & -- & 100.0$\pm$0.0 & -- & 99.0$\pm$1.7 & 98.0$\pm$2.0 & 100.0$\pm$0.0 & 99.0$\pm$1.7 & 100.0$\pm$0.0 & 100.0$\pm$0.0 & 100.0$\pm$0.0 \\
move\_hanger & 91.0$\pm$5.2 & 0.0$\pm$0.0 & 61.0$\pm$4.4 & 83.0$\pm$18.4 & -- & -- & -- & -- & 55.0$\pm$5.9 & 69.0$\pm$5.9 & 29.0$\pm$5.2 & 92.0$\pm$2.8 & 94.0$\pm$2.0 & 87.0$\pm$4.4 & 22.0$\pm$2.0 \\
open\_drawer & 99.0$\pm$1.7 & 25.0$\pm$4.4 & 63.0$\pm$4.4 & -- & -- & -- & 92.0$\pm$0.0 & -- & 88.0$\pm$0.0 & 92.0$\pm$0.0 & 99.0$\pm$1.7 & 86.0$\pm$8.2 & 100.0$\pm$0.0 & 95.0$\pm$1.7 & 95.0$\pm$1.7 \\
place\_wine\_at\_rack\_location & 96.0$\pm$4.9 & 28.0$\pm$6.3 & 74.0$\pm$4.5 & 98.0$\pm$2.0 & -- & 93.0$\pm$5.2 & 87.0$\pm$3.3 & 90.0$\pm$6.6 & 81.0$\pm$7.1 & 87.0$\pm$4.4 & 95.0$\pm$6.6 & 83.0$\pm$3.3 & 89.0$\pm$5.9 & 96.0$\pm$2.8 & 91.0$\pm$5.2 \\
put\_money\_in\_safe & 77.0$\pm$4.4 & 9.0$\pm$1.7 & 45.0$\pm$3.3 & 22.0$\pm$3.5 & 55.0$\pm$6.6 & 73.0$\pm$3.3 & 69.0$\pm$1.7 & -- & 56.0$\pm$2.8 & 70.0$\pm$4.5 & 72.0$\pm$6.3 & 82.0$\pm$6.6 & 79.0$\pm$3.3 & 77.0$\pm$8.7 & 62.0$\pm$6.0 \\
reach\_and\_drag & 86.0$\pm$6.6 & 0.0$\pm$0.0 & 72.0$\pm$5.7 & 80.0$\pm$5.7 & 60.0$\pm$6.9 & 67.0$\pm$5.9 & 87.0$\pm$6.6 & 55.0$\pm$4.4 & 68.0$\pm$2.8 & 76.0$\pm$2.8 & 71.0$\pm$5.2 & 61.0$\pm$6.6 & 88.0$\pm$2.8 & 86.0$\pm$3.5 & 81.0$\pm$5.9 \\
scoop\_with\_spatula & 89.0$\pm$5.2 & 2.0$\pm$3.5 & 75.0$\pm$4.4 & 87.0$\pm$3.3 & 84.0$\pm$4.9 & 92.0$\pm$7.5 & 94.0$\pm$4.5 & 83.0$\pm$5.9 & 54.0$\pm$2.0 & 79.0$\pm$5.2 & 74.0$\pm$6.0 & 83.0$\pm$5.9 & 92.0$\pm$2.8 & 91.0$\pm$1.7 & 89.0$\pm$4.4 \\
setup\_chess & 3.0$\pm$1.7 & 0.0$\pm$0.0 & 0.0$\pm$0.0 & 4.0$\pm$2.8 & 4.0$\pm$4.0 & -- & 17.0$\pm$7.1 & -- & 7.0$\pm$5.2 & 7.0$\pm$3.3 & 9.0$\pm$7.1 & 14.0$\pm$4.5 & 14.0$\pm$3.5 & 16.0$\pm$8.9 & 9.0$\pm$3.3 \\
slide\_block\_to\_target & 100.0$\pm$0.0 & 11.0$\pm$4.4 & 45.0$\pm$1.7 & -- & 97.0$\pm$1.7 & -- & -- & -- & 84.0$\pm$4.9 & 96.0$\pm$0.0 & 83.0$\pm$5.2 & 82.0$\pm$8.7 & 100.0$\pm$0.0 & 100.0$\pm$0.0 & 100.0$\pm$0.0 \\
stack\_cups & 35.0$\pm$5.2 & 0.0$\pm$0.0 & 47.0$\pm$5.9 & -- & 45.0$\pm$5.9 & -- & 23.0$\pm$4.4 & -- & 18.0$\pm$2.0 & 16.0$\pm$4.0 & 13.0$\pm$9.5 & 19.0$\pm$7.7 & 24.0$\pm$2.8 & 43.0$\pm$9.1 & 40.0$\pm$2.8 \\
straighten\_rope & 66.0$\pm$11.5 & 0.0$\pm$0.0 & 25.0$\pm$3.3 & -- & 66.0$\pm$10.0 & -- & -- & -- & 53.0$\pm$1.7 & 68.0$\pm$2.8 & 39.0$\pm$11.4 & 42.0$\pm$7.2 & 72.0$\pm$8.5 & 69.0$\pm$6.6 & 75.0$\pm$4.4 \\
turn\_oven\_on & 91.0$\pm$4.4 & 50.0$\pm$10.8 & 68.0$\pm$4.9 & -- & -- & -- & 83.0$\pm$1.7 & -- & 95.0$\pm$3.3 & 97.0$\pm$1.7 & 95.0$\pm$3.3 & 96.0$\pm$0.0 & 96.0$\pm$4.9 & 89.0$\pm$7.1 & 96.0$\pm$2.8 \\
wipe\_desk & 0.0$\pm$0.0 & 0.0$\pm$0.0 & 0.0$\pm$0.0 & 0.0$\pm$0.0 & 0.0$\pm$0.0 & -- & 0.0$\pm$0.0 & -- & 0.0$\pm$0.0 & 0.0$\pm$0.0 & 0.0$\pm$0.0 & 0.0$\pm$0.0 & 0.0$\pm$0.0 & 0.0$\pm$0.0 & 0.0$\pm$0.0 \\
\midrule
Task Mean & 67.8$\pm$1.5 & 15.6$\pm$0.8 & 53.0$\pm$0.9 & 54.6$\pm$0.6 & 59.7$\pm$0.7 & 56.7$\pm$1.4 & 60.9$\pm$0.9 & 53.4$\pm$1.5 & 58.0$\pm$1.1 & 62.6$\pm$0.9 & 56.6$\pm$0.9 & 60.8$\pm$0.5 & 68.7$\pm$1.1 & 68.8$\pm$1.3 & 64.4$\pm$0.5 \\
\bottomrule
\end{tabular}%
}
\end{table*}

\subsection{Per-Task Results on GemBench}
\label{app:gembench_results}

\begin{table}[!t]
\centering
\caption{\textbf{Results on GemBench.}
Success rates (\%) on the four generalization levels of
GemBench~\cite{garcia2024towards} (protocol in
Appendix~\ref{app:eval_protocol}).
Baselines are quoted from~\cite{garcia2024towards}, except the 3D
Diffuser Actor average, recomputed as the mean of its four levels (its
source prints 44.0).
Our rows were trained and evaluated by us on keyframes only,
without demo augmentation.
Best result per column in bold.}
\label{tab:gembench}
\footnotesize
\setlength{\tabcolsep}{3pt}
\setlength{\aboverulesep}{0pt}
\setlength{\belowrulesep}{0pt}
\renewcommand{\arraystretch}{1.25}
\resizebox{\fitwidth}{!}{%
\begin{tabular}{lccccc}
\toprule
& \textbf{Avg.} & \textbf{L1} & \textbf{L2} & \textbf{L3} & \textbf{L4} \\
\multirow{-2}{*}{\textbf{Method}}
& \textbf{SR (\%) $\uparrow$} & \textbf{Placement} & \textbf{Rigid} & \textbf{Articulated} & \textbf{Long-Horizon} \\
\midrule
Hiveformer~\cite{guhur2023instruction} & 30.4 & 60.3$_{\pm 1.5}$ & 26.1$_{\pm 1.4}$ & 35.1$_{\pm 1.7}$ & 0.0$_{\pm 0.0}$ \\
PolarNet~\cite{chen2023polarnet}       & 38.4 & 77.7$_{\pm 0.9}$ & 37.1$_{\pm 1.4}$ & 38.5$_{\pm 1.7}$ & 0.1$_{\pm 0.2}$ \\
3D Diffuser Actor~\cite{3d-da}         & 43.1 & 91.9$_{\pm 0.8}$ & 43.4$_{\pm 2.8}$ & 37.0$_{\pm 2.2}$ & 0.0$_{\pm 0.0}$ \\
RVT-2~\cite{goyal2024rvt}              & 44.0 & 89.1$_{\pm 0.8}$ & 51.0$_{\pm 2.3}$ & 36.0$_{\pm 2.2}$ & 0.0$_{\pm 0.0}$ \\
3D-LOTUS~\cite{garcia2024towards}      & 45.7 & \textbf{94.3}$_{\pm 1.4}$ & 49.9$_{\pm 2.2}$ & 38.1$_{\pm 1.1}$ & 0.3$_{\pm 0.3}$ \\
3D-LOTUS++~\cite{garcia2024towards}    & 48.0 & 68.7$_{\pm 0.6}$ & 64.5$_{\pm 0.9}$ & 41.5$_{\pm 1.8}$ & \textbf{17.4}$_{\pm 0.4}$ \\
\midrule
\textbf{\method (ours)} & 50.0 & 91.1$_{\pm 1.1}$ & 65.0$_{\pm 1.3}$ & \textbf{43.8}$_{\pm 1.2}$ & 0.0$_{\pm 0.0}$ \\
\textbf{\memmethod (ours)} & \textbf{51.1} & 88.6$_{\pm 1.1}$ & \textbf{68.9}$_{\pm 1.8}$ & 38.5$_{\pm 0.9}$ & 8.2$_{\pm 1.0}$ \\
\bottomrule
\end{tabular}%
}
\end{table}

\begin{table}[!t]
\centering
\caption{\textbf{Results on MemoryBench.}
Success rates (\%) on the three MemoryBench
tasks~\cite{fang2025sam2act}.
Baselines are quoted from~\cite{fang2025sam2act} (mean$\pm$std over
four runs; the Avg.\ deviation is the spread across the three tasks);
\method\ and \memmethod\ are mean$\pm$std over five evaluation seeds
(Appendix~\ref{app:eval_protocol}).
Best result per task in bold.}
\label{tab:memorybench}
\footnotesize
\setlength{\tabcolsep}{3.5pt}
\setlength{\aboverulesep}{0pt}
\setlength{\belowrulesep}{0pt}
\renewcommand{\arraystretch}{1.05}
\resizebox{\fitwidth}{!}{%
\begin{tabular}{lcccc}
\toprule
& \textbf{Avg.} & \textbf{Reopen} & \textbf{Put Block} & \textbf{Rearrange} \\
\multirow{-2}{*}{\textbf{Method}} & \textbf{SR (\%) $\uparrow$} & \textbf{Drawer} & \textbf{Back} & \textbf{Block} \\
\midrule
RVT-2~\cite{goyal2024rvt}        & 54.0$\pm$5.3  & 60.0$\pm$0.0 & 50.0$\pm$2.3  & 52.0$\pm$3.3 \\
SAM2Act~\cite{fang2025sam2act}   & 55.0$\pm$24.3 & 48.0$\pm$0.0 & 35.0$\pm$3.8  & 82.0$\pm$2.3 \\
SAM2Act+~\cite{fang2025sam2act}  & 94.3$\pm$9.0  & 84.0$\pm$0.0 & \textbf{100.0$\pm$0.0} & \textbf{99.0$\pm$2.0} \\
\midrule
\textbf{\method\ (ours)}         & 11.3$\pm$0.8  & 29.6$\pm$4.3 & 2.8$\pm$1.8   & 1.6$\pm$2.6 \\
\textbf{\memmethod\ (ours)}      & \textbf{99.7$\pm$0.3} & \textbf{100.0$\pm$0.0} & 99.8$\pm$0.4 & \textbf{99.2$\pm$1.1} \\
\bottomrule
\end{tabular}%
}
\end{table}

\begin{figure*}[t]
  \centering
  \includegraphics[width=0.88\textwidth]{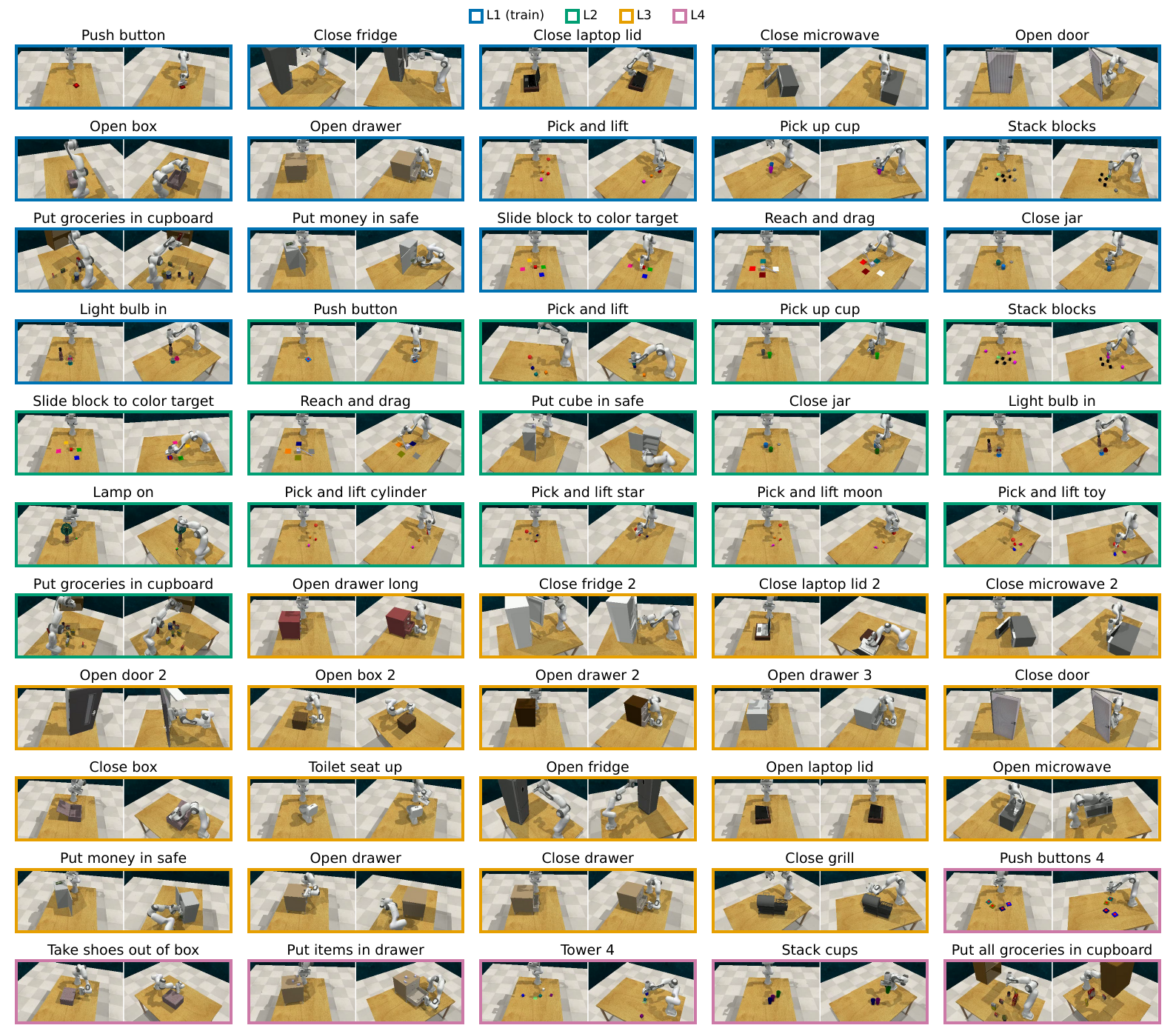}
  \caption{\textbf{The GemBench Task Suite.}
  One representative variation of every task of
  GemBench~\cite{garcia2024towards}, shown as the first and final
  frame of an evaluation rollout.
  Border colors denote the generalization level: \emph{L1} (blue,
  novel placements), \emph{L2} (green, novel rigid objects), \emph{L3}
  (orange, novel articulated objects), and \emph{L4} (pink, novel
  long-horizon tasks).}
  \label{fig:vis_gembench}
\end{figure*}

GemBench~\cite{garcia2024towards} grades generalization hierarchically
over four levels, from the training tasks under changed placements
(L1) through novel rigid (L2) and articulated (L3) objects to novel
long-horizon compositions (L4); Fig.~\ref{fig:vis_gembench} visualizes
the suite and Appendix~\ref{app:eval_protocol} states the protocol.
Alongside Hiveformer~\cite{guhur2023instruction},
PolarNet~\cite{chen2023polarnet}, 3D Diffuser Actor~\cite{3d-da}, and
RVT-2~\cite{goyal2024rvt}, we compare against the benchmark's own
3D-LOTUS, a language-conditioned point-cloud
transformer~\cite{garcia2024towards,wu2024point}, and 3D-LOTUS++, which
wraps the same controller in LLM task planning and VLM object
grounding~\cite{garcia2024towards}.

Behind the best average success rates reported in
Sec.~\ref{sec:exp:additional}, the per-level breakdown of
Table~\ref{tab:gembench} shows that \method\ is competitive with the
strongest specialized 3D policies on the seen tasks of L1, leads the
articulated-object level L3, and scores 0.0\% on L4, while \memmethod\
attains the best L2 result at 68.9\% and lifts L4 off zero to 8.2\%,
at the cost of a few points on L1 and L3.
On L4, where every end-to-end baseline sits at or near zero, the gain
of \memmethod\ comes almost entirely from \emph{PushButtons4}
(Table~\ref{tab:gembench_sota_cmpr_l4_detail}): the neighboring
keyframes in the temporal memory let the policy recall the button it
has just pressed and proceed to the next, state that the current frame
alone does not reveal.
Tables~\ref{tab:gembench_sota_cmpr_l1_detail}--\ref{tab:gembench_sota_cmpr_l4_detail}
report the full per-task success rates behind
Table~\ref{tab:gembench}.

\begin{table*}[p]
\centering
\caption{\textbf{Per-Task Results on GemBench L1 (novel placements).}
Success rates (\%), mean$\pm$std over five random seeds with 20
trials per task variation;
columns are denoted \emph{task+variation}. Baseline numbers are
quoted from~\cite{garcia2024towards}. Best result per column in
bold.}
\label{tab:gembench_sota_cmpr_l1_detail}
\footnotesize
\setlength{\tabcolsep}{4pt}
\setlength{\aboverulesep}{0pt}
\setlength{\belowrulesep}{0pt}
\renewcommand{\arraystretch}{1.05}
\resizebox{\fitwidth}{!}{%
\begin{tabular}{lccccccccccc}
\toprule
\textbf{Method} & \textbf{Avg.} & \begin{tabular}[c]{@{}c@{}}Close\\ Fridge+0\end{tabular} & \begin{tabular}[c]{@{}c@{}}Close\\ Jar+15\end{tabular} & \begin{tabular}[c]{@{}c@{}}Close\\ Jar+16\end{tabular} & \begin{tabular}[c]{@{}c@{}}CloseLaptop\\ Lid+0\end{tabular} & \begin{tabular}[c]{@{}c@{}}Close\\ Microwave+0\end{tabular} & \begin{tabular}[c]{@{}c@{}}LightBulb\\ In+17\end{tabular} & \begin{tabular}[c]{@{}c@{}}LightBulb\\ In+19\end{tabular} & \begin{tabular}[c]{@{}c@{}}Open\\ Box+0\end{tabular} & \begin{tabular}[c]{@{}c@{}}Open\\ Door+0\end{tabular} & \begin{tabular}[c]{@{}c@{}}Open\\ Drawer+0\end{tabular} \\
\midrule
Hiveformer~\cite{guhur2023instruction} & 60.3$_{\pm 1.5}$ & 96$_{\pm 4.2}$ & 64$_{\pm 13.9}$ & 92$_{\pm 2.7}$ & 90$_{\pm 3.5}$ & 88$_{\pm 7.6}$ & 12$_{\pm 4.5}$ & 13$_{\pm 6.7}$ & 4$_{\pm 4.2}$ & 53$_{\pm 15.2}$ & 15$_{\pm 12.2}$ \\
PolarNet~\cite{chen2023polarnet} & 77.6$_{\pm 0.9}$ & 99$_{\pm 2.2}$ & 99$_{\pm 2.2}$ & 99$_{\pm 2.2}$ & 95$_{\pm 3.5}$ & 98$_{\pm 2.7}$ & 72$_{\pm 12.5}$ & 71$_{\pm 6.5}$ & 32$_{\pm 11.5}$ & 69$_{\pm 8.9}$ & 61$_{\pm 12.4}$ \\
3D Diffuser Actor~\cite{3d-da} & 91.9$_{\pm 0.8}$ & \textbf{100}$_{\pm 0.0}$ & \textbf{100}$_{\pm 0.0}$ & \textbf{100}$_{\pm 0.0}$ & 99$_{\pm 2.2}$ & \textbf{100}$_{\pm 0.0}$ & 85$_{\pm 5.0}$ & 88$_{\pm 2.7}$ & 11$_{\pm 2.2}$ & 96$_{\pm 4.2}$ & 82$_{\pm 9.1}$ \\
RVT-2~\cite{goyal2024rvt} & 89.0$_{\pm 0.8}$ & 77$_{\pm 11.0}$ & 97$_{\pm 4.5}$ & 98$_{\pm 2.7}$ & 77$_{\pm 13.0}$ & \textbf{100}$_{\pm 0.0}$ & \textbf{93}$_{\pm 5.7}$ & \textbf{91}$_{\pm 8.2}$ & 7$_{\pm 4.5}$ & \textbf{98}$_{\pm 4.5}$ & \textbf{93}$_{\pm 5.7}$ \\
3D-LOTUS~\cite{garcia2024towards} & \textbf{94.3}$_{\pm 3.5}$ & 96$_{\pm 3.7}$ & \textbf{100}$_{\pm 0.0}$ & \textbf{100}$_{\pm 0.0}$ & 98$_{\pm 2.5}$ & 98$_{\pm 4.0}$ & 84$_{\pm 7.4}$ & 85$_{\pm 9.5}$ & \textbf{99}$_{\pm 2.0}$ & 77$_{\pm 2.5}$ & 83$_{\pm 8.7}$ \\
3D-LOTUS++~\cite{garcia2024towards} & 68.7$_{\pm 0.6}$ & 95$_{\pm 0.0}$ & \textbf{100$_{\pm 0.0}$} & 99$_{\pm 2.0}$ & 28$_{\pm 2.5}$ & 87$_{\pm 5.1}$ & 55$_{\pm 10.5}$ & 45$_{\pm 8.9}$ & 55$_{\pm 8.9}$ & 79$_{\pm 9.7}$ & 68$_{\pm 12.5}$ \\
\midrule
\textbf{\method\ (ours)} & 91.1$_{\pm 1.1}$ & 99$_{\pm 2.0}$ & 98$_{\pm 4.0}$ & \textbf{100}$_{\pm 0.0}$ & 97$_{\pm 2.5}$ & 85$_{\pm 5.5}$ & 90$_{\pm 5.5}$ & 87$_{\pm 7.5}$ & 76$_{\pm 10.2}$ & 70$_{\pm 12.3}$ & 86$_{\pm 5.8}$ \\
\textbf{\memmethod\ (ours)} & 88.6$_{\pm 1.1}$ & 98$_{\pm 2.4}$ & \textbf{100}$_{\pm 0.0}$ & \textbf{100}$_{\pm 0.0}$ & \textbf{100}$_{\pm 0.0}$ & 93$_{\pm 2.4}$ & 91$_{\pm 4.9}$ & 90$_{\pm 4.5}$ & 25$_{\pm 18.2}$ & 90$_{\pm 6.3}$ & 74$_{\pm 9.7}$ \\
\midrule
\textbf{Method} & \begin{tabular}[c]{@{}c@{}}Open\\ Drawer+2\end{tabular} & \begin{tabular}[c]{@{}c@{}}Pick\&\\ Lift+0\end{tabular} & \begin{tabular}[c]{@{}c@{}}Pick\&\\ Lift+2\end{tabular} & \begin{tabular}[c]{@{}c@{}}Pick\&\\ Lift+7\end{tabular} & \begin{tabular}[c]{@{}c@{}}PickUp\\ Cup+8\end{tabular} & \begin{tabular}[c]{@{}c@{}}PickUp\\ Cup+9\end{tabular} & \begin{tabular}[c]{@{}c@{}}PickUp\\ Cup+11\end{tabular} & \begin{tabular}[c]{@{}c@{}}Push\\ Button+0\end{tabular} & \begin{tabular}[c]{@{}c@{}}Push\\ Button+3\end{tabular} & \begin{tabular}[c]{@{}c@{}}Push\\ Button+4\end{tabular} & \begin{tabular}[c]{@{}c@{}}PutIn\\ Cupboard+0\end{tabular} \\
\midrule
Hiveformer~\cite{guhur2023instruction} & 59$_{\pm 7.4}$ & 86$_{\pm 4.2}$ & 92$_{\pm 6.7}$ & 93$_{\pm 2.7}$ & 83$_{\pm 7.6}$ & 69$_{\pm 12.9}$ & 61$_{\pm 19.8}$ & 84$_{\pm 11.9}$ & 68$_{\pm 6.7}$ & 87$_{\pm 7.6}$ & 34$_{\pm 8.2}$ \\
PolarNet~\cite{chen2023polarnet} & 90$_{\pm 7.1}$ & 92$_{\pm 9.1}$ & 84$_{\pm 7.4}$ & 88$_{\pm 5.7}$ & 82$_{\pm 7.6}$ & 79$_{\pm 4.2}$ & 72$_{\pm 10.4}$ & \textbf{100}$_{\pm 0.0}$ & \textbf{100}$_{\pm 0.0}$ & 99$_{\pm 2.2}$ & 52$_{\pm 7.6}$ \\
3D Diffuser Actor~\cite{3d-da} & 97$_{\pm 4.5}$ & \textbf{99}$_{\pm 2.2}$ & 99$_{\pm 2.2}$ & 99$_{\pm 2.2}$ & 96$_{\pm 2.2}$ & 97$_{\pm 4.5}$ & 98$_{\pm 2.7}$ & 98$_{\pm 2.7}$ & 96$_{\pm 4.2}$ & 98$_{\pm 2.7}$ & 85$_{\pm 5.0}$ \\
RVT-2~\cite{goyal2024rvt} & 94$_{\pm 4.2}$ & \textbf{99}$_{\pm 2.2}$ & 98$_{\pm 2.7}$ & \textbf{100}$_{\pm 0.0}$ & \textbf{99}$_{\pm 2.2}$ & \textbf{99}$_{\pm 2.2}$ & \textbf{99}$_{\pm 2.2}$ & \textbf{100}$_{\pm 0.0}$ & \textbf{100}$_{\pm 0.0}$ & \textbf{100}$_{\pm 0.0}$ & 88$_{\pm 8.4}$ \\
3D-LOTUS~\cite{garcia2024towards} & 93$_{\pm 6.0}$ & \textbf{99}$_{\pm 2.0}$ & \textbf{100}$_{\pm 0.0}$ & 99$_{\pm 2.0}$ & 97$_{\pm 4.0}$ & 96$_{\pm 3.7}$ & 94$_{\pm 4.9}$ & 99$_{\pm 2.0}$ & 99$_{\pm 2.0}$ & \textbf{100}$_{\pm 0.0}$ & \textbf{89}$_{\pm 5.8}$ \\
3D-LOTUS++~\cite{garcia2024towards} & 75$_{\pm 4.5}$ & 97$_{\pm 6.0}$ & 94$_{\pm 3.7}$ & 93$_{\pm 5.1}$ & 86$_{\pm 8.0}$ & 88$_{\pm 6.8}$ & 91$_{\pm 4.9}$ & \textbf{100$_{\pm 0.0}$} & \textbf{100$_{\pm 0.0}$} & \textbf{100$_{\pm 0.0}$} & 1$_{\pm 2.0}$ \\
\midrule
\textbf{\method\ (ours)} & \textbf{99}$_{\pm 2.0}$ & \textbf{99}$_{\pm 2.0}$ & \textbf{100}$_{\pm 0.0}$ & 98$_{\pm 2.5}$ & 96$_{\pm 2.0}$ & 94$_{\pm 3.7}$ & \textbf{99}$_{\pm 2.0}$ & \textbf{100}$_{\pm 0.0}$ & 98$_{\pm 4.0}$ & 98$_{\pm 4.0}$ & 74$_{\pm 6.6}$ \\
\textbf{\memmethod\ (ours)} & 95$_{\pm 3.2}$ & \textbf{99}$_{\pm 2.0}$ & 98$_{\pm 2.4}$ & 98$_{\pm 2.4}$ & 89$_{\pm 3.7}$ & 90$_{\pm 6.3}$ & 91$_{\pm 4.9}$ & \textbf{100}$_{\pm 0.0}$ & 97$_{\pm 2.4}$ & 97$_{\pm 2.4}$ & 82$_{\pm 8.7}$ \\
\midrule
\textbf{Method} & \begin{tabular}[c]{@{}c@{}}PutIn\\ Cupboard+3\end{tabular} & \begin{tabular}[c]{@{}c@{}}PutMoney\\ InSafe+0\end{tabular} & \begin{tabular}[c]{@{}c@{}}PutMoney\\ InSafe+1\end{tabular} & \begin{tabular}[c]{@{}c@{}}Reach\&\\ Drag+14\end{tabular} & \begin{tabular}[c]{@{}c@{}}Reach\&\\ Drag+18\end{tabular} & \begin{tabular}[c]{@{}c@{}}Slide\\ Block+0\end{tabular} & \begin{tabular}[c]{@{}c@{}}Slide\\ Block+1\end{tabular} & \begin{tabular}[c]{@{}c@{}}Stack\\ Blocks+30\end{tabular} & \begin{tabular}[c]{@{}c@{}}Stack\\ Blocks+36\end{tabular} & \begin{tabular}[c]{@{}c@{}}Stack\\ Blocks+39\end{tabular} &  \\
\midrule
Hiveformer~\cite{guhur2023instruction} & 74$_{\pm 6.5}$ & 85$_{\pm 3.5}$ & 88$_{\pm 2.7}$ & 37$_{\pm 5.7}$ & 32$_{\pm 7.6}$ & 99$_{\pm 2.2}$ & 91$_{\pm 12.4}$ & 6$_{\pm 5.5}$ & 7$_{\pm 4.5}$ & 6$_{\pm 4.2}$ &  \\
PolarNet~\cite{chen2023polarnet} & \textbf{88}$_{\pm 4.5}$ & 93$_{\pm 4.5}$ & 95$_{\pm 5.0}$ & 99$_{\pm 2.2}$ & 99$_{\pm 2.2}$ & \textbf{100}$_{\pm 0.0}$ & 0$_{\pm 0.0}$ & 34$_{\pm 10.8}$ & 30$_{\pm 9.4}$ & 36$_{\pm 12.9}$ &  \\
3D Diffuser Actor~\cite{3d-da} & 82$_{\pm 11.5}$ & \textbf{95}$_{\pm 5.0}$ & 98$_{\pm 2.7}$ & \textbf{100}$_{\pm 0.0}$ & 99$_{\pm 2.2}$ & \textbf{100}$_{\pm 0.0}$ & 89$_{\pm 4.2}$ & 88$_{\pm 7.6}$ & 85$_{\pm 6.1}$ & 89$_{\pm 5.5}$ &  \\
RVT-2~\cite{goyal2024rvt} & 80$_{\pm 6.1}$ & 93$_{\pm 8.4}$ & 96$_{\pm 8.5}$ & 85$_{\pm 10.0}$ & 94$_{\pm 2.2}$ & \textbf{100}$_{\pm 0.0}$ & 37$_{\pm 6.7}$ & 88$_{\pm 5.7}$ & \textbf{93}$_{\pm 2.7}$ & 88$_{\pm 11.5}$ &  \\
3D-LOTUS~\cite{garcia2024towards} & 72$_{\pm 11.2}$ & 94$_{\pm 3.7}$ & \textbf{99}$_{\pm 2.0}$ & 99$_{\pm 2.0}$ & \textbf{100}$_{\pm 0.0}$ & \textbf{100}$_{\pm 0.0}$ & \textbf{100}$_{\pm 0.0}$ & \textbf{94}$_{\pm 5.8}$ & 91$_{\pm 6.6}$ & \textbf{90}$_{\pm 4.5}$ & \\
3D-LOTUS++~\cite{garcia2024towards} & 2$_{\pm 2.5}$ & 22$_{\pm 6.8}$ & 16$_{\pm 4.9}$ & 94$_{\pm 3.7}$ & 62$_{\pm 8.7}$ & \textbf{100$_{\pm 0.0}$} & 65$_{\pm 5.5}$ & 86$_{\pm 5.8}$ & 20$_{\pm 4.5}$ & 28$_{\pm 13.6}$ & \\
\midrule
\textbf{\method\ (ours)} & 84$_{\pm 6.6}$ & 79$_{\pm 9.7}$ & 86$_{\pm 3.7}$ & 96$_{\pm 5.8}$ & 97$_{\pm 4.0}$ & \textbf{100}$_{\pm 0.0}$ & 90$_{\pm 5.5}$ & 77$_{\pm 8.1}$ & 87$_{\pm 4.0}$ & 85$_{\pm 7.8}$ & \\
\textbf{\memmethod\ (ours)} & 76$_{\pm 3.7}$ & 92$_{\pm 6.8}$ & 98$_{\pm 4.0}$ & 85$_{\pm 5.5}$ & 83$_{\pm 6.0}$ & \textbf{100}$_{\pm 0.0}$ & 91$_{\pm 6.6}$ & 77$_{\pm 12.1}$ & 76$_{\pm 5.8}$ & 73$_{\pm 10.8}$ & \\
\bottomrule
\end{tabular}%
}
\end{table*}

\begin{table*}[p]
\centering
\caption{\textbf{Per-Task Results on GemBench L2 (novel rigid objects).}
Success rates (\%), mean$\pm$std over five random seeds with 20
trials per task variation;
columns are denoted \emph{task+variation}. Baseline numbers are
quoted from~\cite{garcia2024towards}. Best result per column in
bold.}
\label{tab:gembench_sota_cmpr_l2_detail}
\footnotesize
\setlength{\tabcolsep}{4pt}
\setlength{\aboverulesep}{0pt}
\setlength{\belowrulesep}{0pt}
\renewcommand{\arraystretch}{1.05}
\resizebox{\fitwidth}{!}{%
\begin{tabular}{lcccccccccc}
\toprule
\textbf{Method} & \textbf{Avg.} & \begin{tabular}[c]{@{}c@{}}Push\\ Button+13\end{tabular} & \begin{tabular}[c]{@{}c@{}}Push\\ Button+15\end{tabular} & \begin{tabular}[c]{@{}c@{}}Push\\ Button+17\end{tabular} & \begin{tabular}[c]{@{}c@{}}Pick\&\\ Lift+14\end{tabular} & \begin{tabular}[c]{@{}c@{}}Pick\&\\ Lift+16\end{tabular} & \begin{tabular}[c]{@{}c@{}}Pick\&\\ Lift+18\end{tabular} & \begin{tabular}[c]{@{}c@{}}PickUp\\ Cup+10\end{tabular} & \begin{tabular}[c]{@{}c@{}}PickUp\\ Cup+12\end{tabular} & \begin{tabular}[c]{@{}c@{}}PickUp\\ Cup+13\end{tabular} \\
\midrule
Hiveformer~\cite{guhur2023instruction} & 26.1$_{\pm 1.4}$ & 97$_{\pm 2.7}$ & 85$_{\pm 10.0}$ & 88$_{\pm 2.7}$ & 21$_{\pm 6.5}$ & 9$_{\pm 4.2}$ & 8$_{\pm 6.7}$ & 30$_{\pm 7.1}$ & 22$_{\pm 13.5}$ & 26$_{\pm 10.6}$ \\
PolarNet~\cite{chen2023polarnet} & 37.1$_{\pm 1.4}$ & 100$_{\pm 0.0}$ & \textbf{100$_{\pm 0.0}$} & 85$_{\pm 7.9}$ & 3$_{\pm 4.5}$ & 1$_{\pm 2.2}$ & 0$_{\pm 0.0}$ & 48$_{\pm 11.0}$ & 46$_{\pm 8.9}$ & 16$_{\pm 6.5}$ \\
3D Diffuser Actor~\cite{3d-da} & 43.4$_{\pm 2.8}$ & 87$_{\pm 13.0}$ & 81$_{\pm 6.5}$ & 60$_{\pm 9.4}$ & 9$_{\pm 4.2}$ & 18$_{\pm 9.1}$ & 0$_{\pm 0.0}$ & 84$_{\pm 5.5}$ & 60$_{\pm 11.7}$ & 62$_{\pm 13.0}$ \\
RVT-2~\cite{goyal2024rvt} & 51.0$_{\pm 2.3}$ & \textbf{100$_{\pm 0.0}$} & \textbf{100$_{\pm 0.0}$} & \textbf{100$_{\pm 0.0}$} & 47$_{\pm 7.6}$ & 29$_{\pm 9.6}$ & 8$_{\pm 4.5}$ & 81$_{\pm 8.2}$ & 59$_{\pm 9.6}$ & 72$_{\pm 9.7}$ \\
3D-LOTUS~\cite{garcia2024towards} & 49.9$_{\pm 2.2}$ & 99$_{\pm 2.0}$ & \textbf{100$_{\pm 0.0}$} & \textbf{100$_{\pm 0.0}$} & 3$_{\pm 2.5}$ & 18$_{\pm 8.7}$ & 33$_{\pm 9.3}$ & 89$_{\pm 3.7}$ & 78$_{\pm 8.7}$ & 57$_{\pm 7.5}$ \\
3D-LOTUS++~\cite{garcia2024towards} & 64.5$_{\pm 0.9}$ & 99$_{\pm 2.0}$ & \textbf{100$_{\pm 0.0}$} & 99$_{\pm 2.0}$ & \textbf{94$_{\pm 3.7}$} & \textbf{96$_{\pm 3.7}$} & \textbf{95$_{\pm 3.2}$} & 79$_{\pm 4.9}$ & 89$_{\pm 9.7}$ & 84$_{\pm 10.2}$ \\
\midrule
\textbf{\method\ (ours)} & 65.0$_{\pm 1.3}$ & \textbf{100}$_{\pm 0.0}$ & \textbf{100}$_{\pm 0.0}$ & \textbf{100}$_{\pm 0.0}$ & 74$_{\pm 9.7}$ & 89$_{\pm 4.9}$ & 0$_{\pm 0.0}$ & \textbf{91}$_{\pm 3.7}$ & \textbf{90}$_{\pm 3.2}$ & \textbf{90}$_{\pm 6.3}$ \\
\textbf{\memmethod\ (ours)} & \textbf{68.9}$_{\pm 1.8}$ & \textbf{100}$_{\pm 0.0}$ & 99$_{\pm 2.0}$ & 96$_{\pm 3.7}$ & 78$_{\pm 8.1}$ & 89$_{\pm 5.8}$ & 31$_{\pm 8.6}$ & 89$_{\pm 4.9}$ & 86$_{\pm 9.7}$ & 86$_{\pm 9.7}$ \\
\midrule
\textbf{Method} & \begin{tabular}[c]{@{}c@{}}Stack\\ Blocks+24\end{tabular} & \begin{tabular}[c]{@{}c@{}}Stack\\ Blocks+27\end{tabular} & \begin{tabular}[c]{@{}c@{}}Stack\\ Blocks+33\end{tabular} & \begin{tabular}[c]{@{}c@{}}Slide\\ Block+2\end{tabular} & \begin{tabular}[c]{@{}c@{}}Slide\\ Block+3\end{tabular} & \begin{tabular}[c]{@{}c@{}}Close\\ Jar+3\end{tabular} & \begin{tabular}[c]{@{}c@{}}Close\\ Jar+4\end{tabular} & \begin{tabular}[c]{@{}c@{}}LightBulb\\ In+1\end{tabular} & \begin{tabular}[c]{@{}c@{}}LightBulb\\ In+2\end{tabular} & \begin{tabular}[c]{@{}c@{}}Lamp\\ On+0\end{tabular} \\
\midrule
Hiveformer~\cite{guhur2023instruction} & 0$_{\pm 0.0}$ & 4$_{\pm 4.2}$ & 0$_{\pm 0.0}$ & 0$_{\pm 0.0}$ & 0$_{\pm 0.0}$ & 0$_{\pm 0.0}$ & 0$_{\pm 0.0}$ & 4$_{\pm 4.2}$ & 0$_{\pm 0.0}$ & 7$_{\pm 4.5}$ \\
PolarNet~\cite{chen2023polarnet} & 1$_{\pm 2.2}$ & 2$_{\pm 2.7}$ & 6$_{\pm 8.2}$ & 0$_{\pm 0.0}$ & 0$_{\pm 0.0}$ & 20$_{\pm 10.6}$ & 82$_{\pm 5.7}$ & 22$_{\pm 11.5}$ & 17$_{\pm 8.4}$ & \textbf{14}$_{\pm 10.8}$ \\
3D Diffuser Actor~\cite{3d-da} & 66$_{\pm 13.9}$ & 82$_{\pm 2.7}$ & 50$_{\pm 14.6}$ & 0$_{\pm 0.0}$ & 0$_{\pm 0.0}$ & 23$_{\pm 16.8}$ & 82$_{\pm 5.7}$ & 51$_{\pm 17.8}$ & 60$_{\pm 10.0}$ & 7$_{\pm 7.6}$ \\
RVT-2~\cite{goyal2024rvt} & 18$_{\pm 4.5}$ & 56$_{\pm 16.7}$ & 45$_{\pm 13.7}$ & 0$_{\pm 0.0}$ & 1$_{\pm 2.2}$ & 7$_{\pm 7.6}$ & 77$_{\pm 5.7}$ & 68$_{\pm 14.4}$ & 6$_{\pm 6.5}$ & 0$_{\pm 0.0}$ \\
3D-LOTUS~\cite{garcia2024towards} & 13$_{\pm 8.1}$ & 40$_{\pm 9.5}$ & 69$_{\pm 5.8}$ & 0$_{\pm 0.0}$ & 0$_{\pm 0.0}$ & 71$_{\pm 5.8}$ & 90$_{\pm 4.5}$ & 24$_{\pm 4.9}$ & 41$_{\pm 8.6}$ & 0$_{\pm 0.0}$ \\
3D-LOTUS++~\cite{garcia2024towards} & 22$_{\pm 9.3}$ & \textbf{83}$_{\pm 7.5}$ & 59$_{\pm 3.7}$ & \textbf{27}$_{\pm 9.8}$ & 5$_{\pm 3.2}$ & \textbf{98}$_{\pm 2.5}$ & \textbf{96}$_{\pm 3.7}$ & 56$_{\pm 9.7}$ & 43$_{\pm 7.5}$ & 2$_{\pm 2.0}$ \\
\midrule
\textbf{\method\ (ours)} & 61$_{\pm 10.7}$ & 51$_{\pm 13.2}$ & \textbf{79}$_{\pm 8.6}$ & 12$_{\pm 9.3}$ & 3$_{\pm 4.0}$ & 66$_{\pm 6.6}$ & 88$_{\pm 4.0}$ & 66$_{\pm 8.6}$ & 74$_{\pm 5.8}$ & 7$_{\pm 4.0}$ \\
\textbf{\memmethod\ (ours)} & \textbf{69}$_{\pm 10.7}$ & 59$_{\pm 6.6}$ & 73$_{\pm 6.8}$ & 11$_{\pm 6.6}$ & \textbf{46}$_{\pm 5.8}$ & 95$_{\pm 4.5}$ & 90$_{\pm 3.2}$ & \textbf{76}$_{\pm 7.3}$ & \textbf{93}$_{\pm 5.1}$ & 11$_{\pm 8.6}$ \\
\midrule
\textbf{Method} & \begin{tabular}[c]{@{}c@{}}Reach\&\\ Drag+5\end{tabular} & \begin{tabular}[c]{@{}c@{}}Reach\&\\ Drag+7\end{tabular} & \begin{tabular}[c]{@{}c@{}}PutCube\\ InSafe+0\end{tabular} & \begin{tabular}[c]{@{}c@{}}Pick\&Lift\\ Cylinder+0\end{tabular} & \begin{tabular}[c]{@{}c@{}}Pick\&Lift\\ Star+0\end{tabular} & \begin{tabular}[c]{@{}c@{}}Pick\&Lift\\ Moon+0\end{tabular} & \begin{tabular}[c]{@{}c@{}}Pick\&Lift\\ Toy+0\end{tabular} & \begin{tabular}[c]{@{}c@{}}PutIn\\ Cupboard+7\end{tabular} & \begin{tabular}[c]{@{}c@{}}PutIn\\ Cupboard+8\end{tabular} &  \\
\midrule
Hiveformer~\cite{guhur2023instruction} & 1$_{\pm 2.2}$ & 0$_{\pm 0.0}$ & 4$_{\pm 2.2}$ & 78$_{\pm 5.7}$ & 73$_{\pm 7.6}$ & 88$_{\pm 2.7}$ & 87$_{\pm 4.5}$ & 0$_{\pm 0.0}$ & 0$_{\pm 0.0}$ &  \\
PolarNet~\cite{chen2023polarnet} & 61$_{\pm 8.2}$ & 10$_{\pm 6.1}$ & \textbf{40}$_{\pm 14.1}$ & 93$_{\pm 6.7}$ & 88$_{\pm 8.4}$ & 93$_{\pm 6.7}$ & 90$_{\pm 3.5}$ & 0$_{\pm 0.0}$ & 0$_{\pm 0.0}$ &  \\
3D Diffuser Actor~\cite{3d-da} & 0$_{\pm 0.0}$ & 64$_{\pm 6.5}$ & 3$_{\pm 2.7}$ & \textbf{99}$_{\pm 2.2}$ & 43$_{\pm 17.9}$ & 91$_{\pm 9.6}$ & 30$_{\pm 9.4}$ & 0$_{\pm 0.0}$ & \textbf{3}$_{\pm 4.5}$ &  \\
RVT-2~\cite{goyal2024rvt} & 91$_{\pm 2.2}$ & 89$_{\pm 6.5}$ & 6$_{\pm 5.5}$ & 98$_{\pm 2.7}$ & 98$_{\pm 4.5}$ & 94$_{\pm 4.2}$ & 78$_{\pm 8.4}$ & 0$_{\pm 0.0}$ & 0$_{\pm 0.0}$ &  \\
3D-LOTUS~\cite{garcia2024towards} & \textbf{95}$_{\pm 4.5}$ & 18$_{\pm 10.8}$ & 25$_{\pm 5.5}$ & 88$_{\pm 8.7}$ & 69$_{\pm 6.6}$ & 80$_{\pm 8.4}$ & \textbf{96}$_{\pm 3.7}$ & 0$_{\pm 0.0}$ & 0$_{\pm 0.0}$  & \\
3D-LOTUS++~\cite{garcia2024towards} & 94$_{\pm 2.0}$ & 64$_{\pm 12.4}$ & 37$_{\pm 5.1}$ & 91$_{\pm 2.0}$ & 94$_{\pm 3.7}$ & 29$_{\pm 6.6}$ & 71$_{\pm 2.0}$ & \textbf{1}$_{\pm 2.0}$ & 0$_{\pm 0.0}$  & \\
\midrule
\textbf{\method\ (ours)} & 94$_{\pm 3.7}$ & \textbf{96}$_{\pm 3.7}$ & 3$_{\pm 2.5}$ & 98$_{\pm 2.5}$ & \textbf{99}$_{\pm 2.0}$ & \textbf{95}$_{\pm 3.2}$ & 93$_{\pm 5.1}$ & 0$_{\pm 0.0}$ & 0$_{\pm 0.0}$ & \\
\textbf{\memmethod\ (ours)} & 90$_{\pm 7.1}$ & 80$_{\pm 6.3}$ & 4$_{\pm 3.7}$ & 91$_{\pm 3.7}$ & 97$_{\pm 2.4}$ & \textbf{95}$_{\pm 3.2}$ & \textbf{96}$_{\pm 5.8}$ & 0$_{\pm 0.0}$ & 0$_{\pm 0.0}$ & \\
\bottomrule
\end{tabular}%
}
\end{table*}

\begin{table*}[p]
\centering
\caption{\textbf{Per-Task Results on GemBench L3 (novel articulated objects).}
Success rates (\%), mean$\pm$std over five random seeds with 20
trials per task variation;
columns are denoted \emph{task+variation}. Baseline numbers are
quoted from~\cite{garcia2024towards}. Best result per column in
bold.}
\label{tab:gembench_sota_cmpr_l3_detail}
\footnotesize
\setlength{\tabcolsep}{4pt}
\setlength{\aboverulesep}{0pt}
\setlength{\belowrulesep}{0pt}
\renewcommand{\arraystretch}{1.05}
\resizebox{\fitwidth}{!}{%
\begin{tabular}{lcccccccc}
\toprule
\textbf{Method} & \textbf{Avg.} & \begin{tabular}[c]{@{}c@{}}Close\\ Door+0\end{tabular} & \begin{tabular}[c]{@{}c@{}}Close\\ Box+0\end{tabular} & \begin{tabular}[c]{@{}c@{}}Close\\ Fridge2+0\end{tabular} & \begin{tabular}[c]{@{}c@{}}CloseLaptop\\ Lid2+0\end{tabular} & \begin{tabular}[c]{@{}c@{}}Close\\ Microwave2+0\end{tabular} & \begin{tabular}[c]{@{}c@{}}Open\\ Door2+0\end{tabular} & \begin{tabular}[c]{@{}c@{}}Open\\ Box2+0\end{tabular} \\
\midrule
Hiveformer~\cite{guhur2023instruction} & 35.1$_{\pm 1.7}$ & 0$_{\pm 0.0}$ & 1$_{\pm 2.2}$ & 34$_{\pm 9.6}$ & 52$_{\pm 9.1}$ & 15$_{\pm 7.1}$ & 32$_{\pm 11.5}$ & 5$_{\pm 3.5}$ \\
PolarNet~\cite{chen2023polarnet} & 38.5$_{\pm 1.7}$ & 0$_{\pm 0.0}$ & 0$_{\pm 0.0}$ & 78$_{\pm 5.7}$ & 26$_{\pm 8.2}$ & 74$_{\pm 6.5}$ & 33$_{\pm 6.7}$ & 23$_{\pm 8.4}$ \\
3D Diffuser Actor~\cite{3d-da} & 37.0$_{\pm 2.2}$ & 0$_{\pm 0.0}$ & 0$_{\pm 0.0}$ & \textbf{97}$_{\pm 2.7}$ & 23$_{\pm 6.7}$ & 88$_{\pm 7.6}$ & \textbf{86}$_{\pm 7.4}$ & 67$_{\pm 9.8}$ \\
RVT-2~\cite{goyal2024rvt} & 36.0$_{\pm 2.2}$ & 1$_{\pm 2.2}$ & 2$_{\pm 2.7}$ & 72$_{\pm 6.7}$ & 42$_{\pm 14.0}$ & 71$_{\pm 8.9}$ & 79$_{\pm 6.5}$ & 5$_{\pm 6.1}$ \\
3D-LOTUS~\cite{garcia2024towards} & 38.1$_{\pm 1.1}$ & 0$_{\pm 0.0}$ & \textbf{58}$_{\pm 8.1}$ & 36$_{\pm 9.7}$ & 54$_{\pm 10.7}$ & 85$_{\pm 7.1}$ & 42$_{\pm 6.8}$ & 11$_{\pm 6.6}$ \\
3D-LOTUS++~\cite{garcia2024towards} & 41.5$_{\pm 1.8}$ & \textbf{1}$_{\pm 2.0}$ & 29$_{\pm 8.6}$ & 93$_{\pm 2.5}$ & 50$_{\pm 9.5}$ & \textbf{99}$_{\pm 2.0}$ & 52$_{\pm 10.3}$ & 16$_{\pm 8.0}$ \\
\midrule
\textbf{\method\ (ours)} & \textbf{43.8}$_{\pm 1.2}$ & 0$_{\pm 0.0}$ & 1$_{\pm 2.0}$ & 95$_{\pm 5.5}$ & \textbf{77}$_{\pm 4.0}$ & 54$_{\pm 10.2}$ & 68$_{\pm 10.8}$ & \textbf{74}$_{\pm 4.9}$ \\
\textbf{\memmethod\ (ours)} & 38.5$_{\pm 0.9}$ & 0$_{\pm 0.0}$ & 0$_{\pm 0.0}$ & \textbf{97}$_{\pm 4.0}$ & 15$_{\pm 5.5}$ & 34$_{\pm 9.2}$ & 73$_{\pm 5.1}$ & 30$_{\pm 10.5}$ \\
\midrule
\textbf{Method} & \begin{tabular}[c]{@{}c@{}}Open\\ Drawer2+0\end{tabular} & \begin{tabular}[c]{@{}c@{}}Open\\ Drawer3+0\end{tabular} & \begin{tabular}[c]{@{}c@{}}OpenDrawer\\ Long+0\end{tabular} & \begin{tabular}[c]{@{}c@{}}OpenDrawer\\ Long+1\end{tabular} & \begin{tabular}[c]{@{}c@{}}OpenDrawer\\ Long+2\end{tabular} & \begin{tabular}[c]{@{}c@{}}OpenDrawer\\ Long+3\end{tabular} & \begin{tabular}[c]{@{}c@{}}Toilet\\ SeatUp+0\end{tabular} & \begin{tabular}[c]{@{}c@{}}Open\\ Fridge+0\end{tabular} \\
\midrule
Hiveformer~\cite{guhur2023instruction} & 59$_{\pm 11.9}$ & 39$_{\pm 11.9}$ & 78$_{\pm 8.4}$ & 82$_{\pm 4.5}$ & 49$_{\pm 4.2}$ & 57$_{\pm 11.5}$ & 6$_{\pm 4.2}$ & 0$_{\pm 0.0}$ \\
PolarNet~\cite{chen2023polarnet} & \textbf{91}$_{\pm 4.2}$ & 29$_{\pm 8.2}$ & 84$_{\pm 11.9}$ & \textbf{88}$_{\pm 5.7}$ & \textbf{63}$_{\pm 8.4}$ & 37$_{\pm 7.6}$ & 2$_{\pm 2.7}$ & 4$_{\pm 2.2}$ \\
3D Diffuser Actor~\cite{3d-da} & 19$_{\pm 8.2}$ & 1$_{\pm 2.2}$ & 15$_{\pm 5.0}$ & 35$_{\pm 13.7}$ & 26$_{\pm 9.6}$ & 79$_{\pm 12.9}$ & 0$_{\pm 0.0}$ & 7$_{\pm 5.7}$ \\
RVT-2~\cite{goyal2024rvt} & 81$_{\pm 11.9}$ & 0$_{\pm 0.0}$ & \textbf{84}$_{\pm 8.2}$ & 39$_{\pm 10.8}$ & 11$_{\pm 8.9}$ & 75$_{\pm 6.1}$ & 7$_{\pm 5.7}$ & 0$_{\pm 0.0}$ \\
3D-LOTUS~\cite{garcia2024towards} & 90$_{\pm 3.2}$ & 22$_{\pm 8.1}$ & 56$_{\pm 13.9}$ & 33$_{\pm 11.2}$ & 17$_{\pm 8.1}$ & 75$_{\pm 6.3}$ & 0$_{\pm 0.0}$ & 4$_{\pm 5.8}$ \\
3D-LOTUS++~\cite{garcia2024towards} & 70$_{\pm 5.5}$ & 41$_{\pm 4.9}$ & 72$_{\pm 4.0}$ & 52$_{\pm 10.8}$ & 23$_{\pm 8.1}$ & 78$_{\pm 5.1}$ & \textbf{8}$_{\pm 5.1}$ & 0$_{\pm 0.0}$ \\
\midrule
\textbf{\method\ (ours)} & 65$_{\pm 6.3}$ & \textbf{87}$_{\pm 6.0}$ & 59$_{\pm 8.6}$ & 34$_{\pm 8.0}$ & 18$_{\pm 10.3}$ & \textbf{85}$_{\pm 8.4}$ & 6$_{\pm 5.8}$ & 7$_{\pm 2.5}$ \\
\textbf{\memmethod\ (ours)} & 85$_{\pm 9.5}$ & 82$_{\pm 11.2}$ & 52$_{\pm 8.7}$ & 43$_{\pm 8.1}$ & 11$_{\pm 9.7}$ & 83$_{\pm 6.8}$ & 1$_{\pm 2.0}$ & \textbf{11}$_{\pm 10.7}$ \\
\midrule
\textbf{Method} & \begin{tabular}[c]{@{}c@{}}OpenLaptop\\ Lid+0\end{tabular} & \begin{tabular}[c]{@{}c@{}}Open\\ Microwave+0\end{tabular} & \begin{tabular}[c]{@{}c@{}}PutMoney\\ InSafe+2\end{tabular} & \begin{tabular}[c]{@{}c@{}}Open\\ Drawer+1\end{tabular} & \begin{tabular}[c]{@{}c@{}}Close\\ Drawer+0\end{tabular} & \begin{tabular}[c]{@{}c@{}}Close\\ Grill+0\end{tabular} &  &  \\
\midrule
Hiveformer~\cite{guhur2023instruction} & \textbf{100}$_{\pm 0.0}$ & 0$_{\pm 0.0}$ & 0$_{\pm 0.0}$ & 0$_{\pm 0.0}$ & 83$_{\pm 5.7}$ & 44$_{\pm 10.8}$ &  &  \\
PolarNet~\cite{chen2023polarnet} & \textbf{100}$_{\pm 0.0}$ & 0$_{\pm 0.0}$ & 1$_{\pm 2.2}$ & 4$_{\pm 4.2}$ & 29$_{\pm 11.9}$ & 42$_{\pm 11.5}$ &  &  \\
3D Diffuser Actor~\cite{3d-da} & \textbf{100}$_{\pm 0.0}$ & 0$_{\pm 0.0}$ & 2$_{\pm 4.5}$ & 0$_{\pm 0.0}$ & 66$_{\pm 7.4}$ & \textbf{65}$_{\pm 13.7}$ &  &  \\
RVT-2~\cite{goyal2024rvt} & 93$_{\pm 5.7}$ & 0$_{\pm 0.0}$ & 0$_{\pm 0.0}$ & \textbf{6}$_{\pm 2.2}$ & 78$_{\pm 8.4}$ & 9$_{\pm 4.2}$ &  &  \\
3D-LOTUS~\cite{garcia2024towards} & \textbf{100}$_{\pm 0.0}$ & 0$_{\pm 0.0}$ & 0$_{\pm 0.0}$ & 0$_{\pm 0.0}$ & \textbf{87}$_{\pm 8.1}$ & 29$_{\pm 6.6}$ &  &  \\
3D-LOTUS++~\cite{garcia2024towards} & 86$_{\pm 6.6}$ & 0$_{\pm 0.0}$ & \textbf{13}$_{\pm 8.1}$ & 0$_{\pm 0.0}$ & 69$_{\pm 5.8}$ & 19$_{\pm 13.9}$ &  &  \\
\midrule
\textbf{\method\ (ours)} & 95$_{\pm 0.0}$ & 0$_{\pm 0.0}$ & 2$_{\pm 2.5}$ & 0$_{\pm 0.0}$ & 58$_{\pm 12.9}$ & 35$_{\pm 12.3}$ &  &  \\
\textbf{\memmethod\ (ours)} & \textbf{100}$_{\pm 0.0}$ & 0$_{\pm 0.0}$ & 0$_{\pm 0.0}$ & 0$_{\pm 0.0}$ & 51$_{\pm 9.7}$ & 40$_{\pm 7.7}$ & & \\
\bottomrule
\end{tabular}%
}
\end{table*}

\begin{table*}[p]
\centering
\caption{\textbf{Per-Task Results on GemBench L4 (novel long-horizon tasks).}
Success rates (\%), mean$\pm$std over five random seeds with 20
trials per task variation;
columns are denoted \emph{task+variation}. Baseline numbers are
quoted from~\cite{garcia2024towards}. Best result per column in
bold.}
\label{tab:gembench_sota_cmpr_l4_detail}
\footnotesize
\setlength{\tabcolsep}{4pt}
\setlength{\aboverulesep}{0pt}
\setlength{\belowrulesep}{0pt}
\renewcommand{\arraystretch}{1.05}
\resizebox{\fitwidth}{!}{%
\begin{tabular}{lccccccc}
\toprule
\textbf{Method} & \textbf{Avg.} & \begin{tabular}[c]{@{}c@{}}Push\\ Buttons4+1\end{tabular} & \begin{tabular}[c]{@{}c@{}}Push\\ Buttons4+2\end{tabular} & \begin{tabular}[c]{@{}c@{}}Push\\ Buttons4+3\end{tabular} & \begin{tabular}[c]{@{}c@{}}TakeShoes\\ OutOfBox+0\end{tabular} & \begin{tabular}[c]{@{}c@{}}PutItems\\ InDrawer+0\end{tabular} & \begin{tabular}[c]{@{}c@{}}PutItems\\ InDrawer+2\end{tabular} \\
\midrule
Hiveformer~\cite{guhur2023instruction} & 0$_{\pm 0.0}$ & 0$_{\pm 0.0}$ & 0$_{\pm 0.0}$ & 0$_{\pm 0.0}$ & 0$_{\pm 0.0}$ & 0$_{\pm 0.0}$ & 0$_{\pm 0.0}$ \\
PolarNet~\cite{chen2023polarnet} & 0.1$_{\pm 0.2}$ & 1$_{\pm 2.2}$ & 0$_{\pm 0.0}$ & 0$_{\pm 0.0}$ & 0$_{\pm 0.0}$ & 0$_{\pm 0.0}$ & 0$_{\pm 0.0}$ \\
3D Diffuser Actor~\cite{3d-da} & 0$_{\pm 0.0}$ & 0$_{\pm 0.0}$ & 0$_{\pm 0.0}$ & 0$_{\pm 0.0}$ & 0$_{\pm 0.0}$ & 0$_{\pm 0.0}$ & 0$_{\pm 0.0}$ \\
RVT-2~\cite{goyal2024rvt} & 0$_{\pm 0.0}$ & 0$_{\pm 0.0}$ & 0$_{\pm 0.0}$ & 0$_{\pm 0.0}$ & 0$_{\pm 0.0}$ & 0$_{\pm 0.0}$ & 0$_{\pm 0.0}$ \\
3D-LOTUS~\cite{garcia2024towards} & 0.3$_{\pm 0.3}$ & 3$_{\pm 4.0}$ & 0$_{\pm 0.0}$ & 0$_{\pm 0.0}$ & 0$_{\pm 0.0}$ & 0$_{\pm 0.0}$ & 0$_{\pm 0.0}$ \\
3D-LOTUS++~\cite{garcia2024towards} & \textbf{17.4}$_{\pm 0.4}$ & \textbf{76}$_{\pm 7.4}$ & \textbf{49}$_{\pm 8.6}$ & \textbf{37}$_{\pm 8.1}$ & 0$_{\pm 0.0}$ & 0$_{\pm 0.0}$ & 0$_{\pm 0.0}$ \\
\midrule
\textbf{\method\ (ours)} & 0$_{\pm 0.0}$ & 0$_{\pm 0.0}$ & 0$_{\pm 0.0}$ & 0$_{\pm 0.0}$ & 0$_{\pm 0.0}$ & 0$_{\pm 0.0}$ & 0$_{\pm 0.0}$ \\
\textbf{\memmethod\ (ours)} & 8.2$_{\pm 1.0}$ & 68$_{\pm 8.1}$ & 27$_{\pm 9.3}$ & 3$_{\pm 2.4}$ & 0$_{\pm 0.0}$ & 0$_{\pm 0.0}$ & 0$_{\pm 0.0}$ \\
\midrule
\textbf{Method} & \begin{tabular}[c]{@{}c@{}}PutItems\\ InDrawer+4\end{tabular} & Tower4+1 & Tower4+3 & \begin{tabular}[c]{@{}c@{}}Stack\\ Cups+0\end{tabular} & \begin{tabular}[c]{@{}c@{}}Stack\\ Cups+3\end{tabular} & \begin{tabular}[c]{@{}c@{}}PutAllGroceries\\ InCupboard+0\end{tabular} &  \\
\midrule
Hiveformer~\cite{guhur2023instruction} & 0$_{\pm 0.0}$ & 0$_{\pm 0.0}$ & 0$_{\pm 0.0}$ & 0$_{\pm 0.0}$ & 0$_{\pm 0.0}$ & 0$_{\pm 0.0}$ &  \\
PolarNet~\cite{chen2023polarnet} & 0$_{\pm 0.0}$ & 0$_{\pm 0.0}$ & 0$_{\pm 0.0}$ & 0$_{\pm 0.0}$ & 0$_{\pm 0.0}$ & 0$_{\pm 0.0}$ &  \\
3D Diffuser Actor~\cite{3d-da} & 0$_{\pm 0.0}$ & 0$_{\pm 0.0}$ & 0$_{\pm 0.0}$ & 0$_{\pm 0.0}$ & 0$_{\pm 0.0}$ & 0$_{\pm 0.0}$ &  \\
RVT-2~\cite{goyal2024rvt} & 0$_{\pm 0.0}$ & 0$_{\pm 0.0}$ & 0$_{\pm 0.0}$ & 0$_{\pm 0.0}$ & 0$_{\pm 0.0}$ & 0$_{\pm 0.0}$ &  \\
3D-LOTUS~\cite{garcia2024towards} & 0$_{\pm 0.0}$ & 0$_{\pm 0.0}$ & 0$_{\pm 0.0}$ & 0$_{\pm 0.0}$ & 0$_{\pm 0.0}$ & 0$_{\pm 0.0}$ &  \\
3D-LOTUS++~\cite{garcia2024towards} & 0$_{\pm 0.0}$ & \textbf{17}$_{\pm 10.8}$ & \textbf{30}$_{\pm 13.4}$ & 0$_{\pm 0.0}$ & 0$_{\pm 0.0}$ & 0$_{\pm 0.0}$ & \\
\midrule
\textbf{\method\ (ours)} & 0$_{\pm 0.0}$ & 0$_{\pm 0.0}$ & 0$_{\pm 0.0}$ & 0$_{\pm 0.0}$ & 0$_{\pm 0.0}$ & 0$_{\pm 0.0}$ & \\
\textbf{\memmethod\ (ours)} & 0$_{\pm 0.0}$ & 0$_{\pm 0.0}$ & 1$_{\pm 2.0}$ & 0$_{\pm 0.0}$ & 0$_{\pm 0.0}$ & 0$_{\pm 0.0}$ & \\
\bottomrule
\end{tabular}%
}
\end{table*}

\subsection{Per-Task Results on MemoryBench}
\label{app:memorybench_results}

\begin{figure*}[t]
  \centering
  \includegraphics[width=0.8\textwidth]{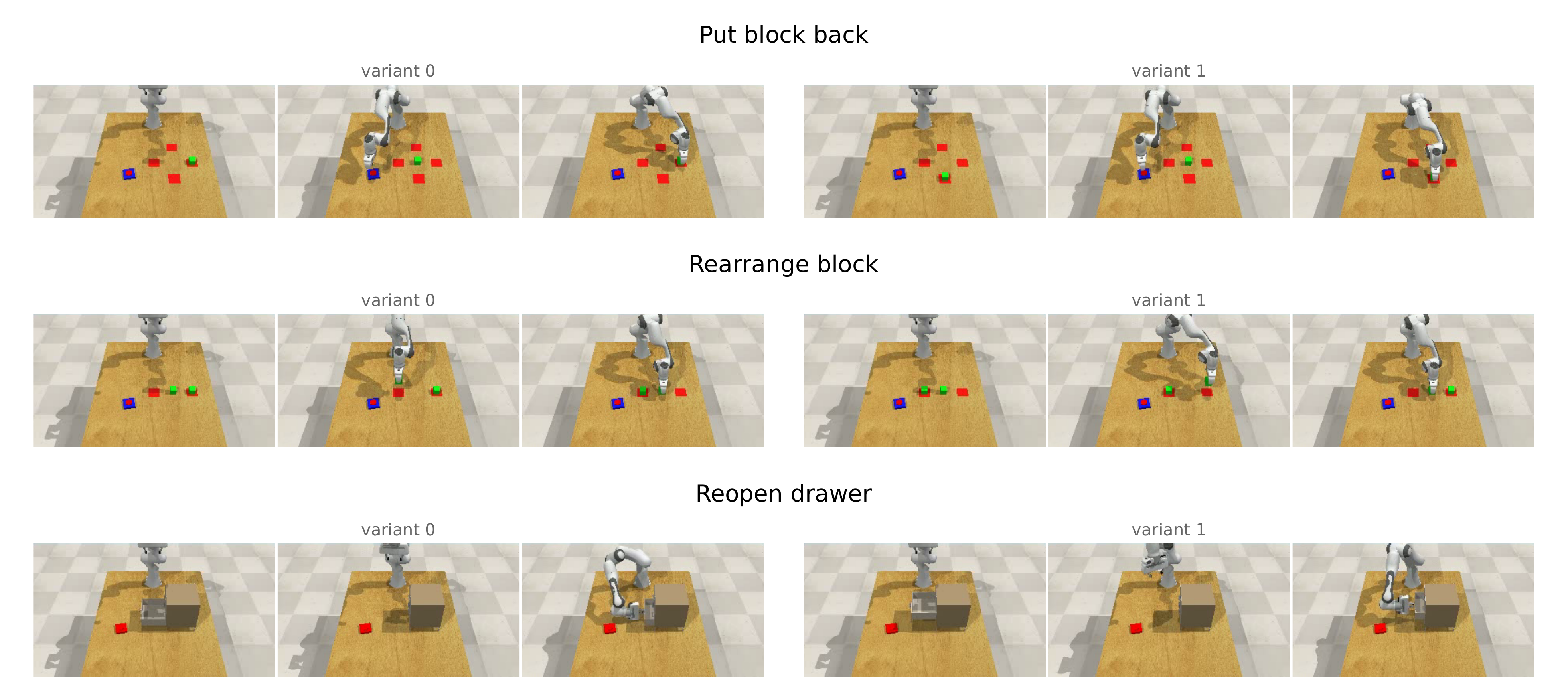}
  \caption{\textbf{The Three MemoryBench Tasks.}
  Two variants of each MemoryBench~\cite{fang2025sam2act} task, each
  shown as three rollout frames in which the robot's own intervention
  erases the evidence a later step depends on.}
  \label{fig:vis_memorybench}
\end{figure*}

MemoryBench~\cite{fang2025sam2act} repeats the memory-dependent test of
Sec.~\ref{sec:exp:rmbench} in a single-arm setting and a different
simulator: its three tasks extend RLBench so that pressing a button
erases the visual evidence a later step depends on
(Fig.~\ref{fig:vis_memorybench}).
Table~\ref{tab:memorybench} compares against RVT-2~\cite{goyal2024rvt}
and against SAM2Act and SAM2Act+~\cite{fang2025sam2act}, whose
SAM2-style memory bank is the closest prior instantiation of visual
episodic memory in a keyframe policy.
\memmethod\ solves the benchmark almost completely, with its margin
over SAM2Act+ concentrated on \emph{Reopen Drawer} (100\% against
84\%), whereas the memory-free \method\ collapses to
$11.3{\pm}0.8\%$; every seed of \memmethod\ scores at or above 99.3\%
overall, so the spread in Table~\ref{tab:memorybench} reflects a
handful of failed episodes rather than run-to-run instability.

\subsection{Per-Task Memory Ablations on RMBench}
\label{app:rmbench_ablation}

Table~\ref{tab:rmbench_ablation} reports the per-task success rates
behind the RMBench memory ablation of Sec.~\ref{sec:exp:ablation},
whose four rows form the memory $2{\times}2$ factorial.
Removing the temporal memory $\mathcal{T}$ collapses precisely the
tasks that require tracking progress or past attempts, with \emph{Press
Button} falling from 93\% to 0\%, \emph{Blocks Ranking Try} from 100\%
to 1\%, and \emph{Rearrange Blocks} from 100\% to 11\%, whereas
removing the spatial memory $\mathcal{S}$ leaves every task within a
few points of the full model, with the largest single-task change on
\emph{Cover Blocks} (91\% vs.\ 99\%).

\begin{table*}[p]
\centering
\caption{\textbf{Memory ablations on RMBench.}
Success rates (\%) over 100 episodes per task for the full \memmethod\ and
its memory ablations: the two single-memory variants (indented) and the
memory-free \method.}
\label{tab:rmbench_ablation}
\footnotesize
\setlength{\tabcolsep}{4pt}
\setlength{\aboverulesep}{0pt}
\setlength{\belowrulesep}{0pt}
\renewcommand{\arraystretch}{1.05}
\begin{tabular}{l cccccc ccccc c}
\toprule
& \multicolumn{6}{c}{\textbf{$M(1)$ tasks}}
& \multicolumn{5}{c}{\textbf{$M(n)$ tasks}} & \\
\cmidrule(lr){2-7} \cmidrule(lr){8-12}
& \textbf{Observe \&} & \textbf{Rearrange} & \textbf{Put Back} & \textbf{Swap} & \textbf{Swap} &
& \textbf{Battery} & \textbf{Blocks} & \textbf{Cover} & \textbf{Press} &
& \textbf{Overall} \\
\multirow{-2}{*}{\textbf{Variant}}
& \textbf{Pick Up} & \textbf{Blocks} & \textbf{Block} & \textbf{Blocks} & \textbf{T} & \multirow{-2}{*}{\textbf{\textit{Avg.}}}
& \textbf{Try} & \textbf{Ranking Try} & \textbf{Blocks} & \textbf{Button} & \multirow{-2}{*}{\textbf{\textit{Avg.}}}
& \textbf{Avg.} \\
\midrule
\textbf{Full \memmethod}                    & 81 & 100 & 100 & 99 & 96 & 95.2 & 96 & 100 & 99 & 93 & 97.0 & 96.0 \\
\tbranch w/o~$\mathcal{S}$ (spatial memory) & 85 & 100 & 100 & 99 & 97 & 96.2 & 95 & 100 & 91 & 92 & 94.5 & 95.4 \\
\tlast w/o~$\mathcal{T}$ (temporal memory)  & 74 & 11  & 38  & 1  & 11 & 27.0 & 51 & 1   & 5  & 0  & 14.3 & 21.3 \\
\method\ (no memory)                        & 75 & 0   & 1   & 11 & 8  & 19.0 & 72 & 0   & 3  & 0  & 18.8 & 18.9 \\
\bottomrule
\end{tabular}
\end{table*}
\begin{table*}[!t]
\centering
\caption{\textbf{Per-Task Real-Robot Results on Franka: 3 vs.\ 10 Demonstrations.}
Success counts of \method\ over 10 trials per task in the Basic
setting (Appendix~\ref{app:real_franka}) when trained with 3 or 10
demonstrations per task.}
\label{tab:app:real_episode}
\footnotesize
\setlength{\tabcolsep}{3pt}
\setlength{\aboverulesep}{0pt}
\setlength{\belowrulesep}{0pt}
\renewcommand{\arraystretch}{1.05}
\begin{tabular}{lcc@{\hspace{3em}}lcc}
\toprule
\textbf{Task} & \textbf{3 demos} & \textbf{10 demos}
& \textbf{Task} & \textbf{3 demos} & \textbf{10 demos} \\
\midrule
Put the RedBull can in the top shelf        & 9/10  & 10/10
& Place the red block in the purple plate    & 10/10 & 10/10 \\
Put the soda can in the bottom shelf        & 9/10  & 9/10
& Place the yellow block in the green plate  & 10/10 & 10/10 \\
Put the RedBull can in the bottom shelf     & 10/10 & 10/10
& Press sanitizer                            & 10/10 & 10/10 \\
Put the coke can in the top shelf           & 10/10 & 10/10
& Put the zebra in the upper drawer          & 9/10  & 9/10 \\
Place the red block in the blue plate       & 10/10 & 10/10
& Put the giraffe in the lower drawer        & 10/10 & 9/10 \\
Place the orange block in the green plate   & 10/10 & 10/10
& Put the zebra in the lower drawer          & 10/10 & 10/10 \\
Put the wolf in the upper drawer            & 7/10  & 9/10
&                                            &       &       \\
\bottomrule
\end{tabular}
\end{table*}
\begin{table*}[!t]
\centering
\caption{\textbf{Per-instruction results on the real Dobot platform.}
Success counts over 10 trials per language instruction and setting;
Tables~\ref{tab:real_dobot_basic} and~\ref{tab:real_dobot_generalization}
aggregate these counts.
\emph{Left:} the three memory-dependent tasks; \emph{right:} the four
memory-free instructions.}
\label{tab:app:real_dobot_all}
\footnotesize
\setlength{\tabcolsep}{3pt}
\setlength{\aboverulesep}{0pt}
\setlength{\belowrulesep}{0pt}
\renewcommand{\arraystretch}{1.05}
\begin{tabular}[t]{lccccc}
\toprule
\textbf{Method} & \textbf{Basic} & \textbf{Distractor}
& \textbf{Background} & \textbf{Height} & \textbf{Lighting} \\
\midrule
\multicolumn{6}{@{}l@{}}{\makecell[l]{\textbf{Cover Blocks}\\[1pt]
\emph{``Put lids on the blocks, then uncover the blue block''}}} \\
SAM2Act+~\cite{fang2025sam2act} & 2/10  & 0/10 & 0/10  & 0/10 & 0/10 \\
\textbf{\method}               & 0/10  & 0/10 & 0/10  & 0/10 & 0/10 \\
\textbf{\memmethod}            & 10/10 & 6/10 & 10/10 & 8/10 & 8/10 \\
\midrule
\multicolumn{6}{@{}l@{}}{\makecell[l]{\textbf{Press Button}\\[1pt]
\emph{``Press the blue button three times,}\\
\emph{then press the yellow button''}}} \\
SAM2Act+~\cite{fang2025sam2act} & 0/10 & 0/10 & 0/10 & 0/10 & 0/10 \\
\textbf{\method}               & 0/10 & 0/10 & 0/10 & 0/10 & 0/10 \\
\textbf{\memmethod}            & 10/10 & 8/10 & 8/10 & 8/10 & 9/10 \\
\midrule
\multicolumn{6}{@{}l@{}}{\makecell[l]{\textbf{Swap Eggplant}\\[1pt]
\emph{``Swap the two eggplants on the plate''}}} \\
SAM2Act+~\cite{fang2025sam2act} & 7/10 & 0/10 & 0/10 & 0/10 & 1/10 \\
\textbf{\method}               & 6/10 & 6/10 & 7/10 & 2/10 & 4/10 \\
\textbf{\memmethod}            & 8/10 & 8/10 & 8/10 & 7/10 & 6/10 \\
\bottomrule
\end{tabular}
\hspace{2.5em}
\begin{tabular}[t]{lccccc}
\toprule
\textbf{Method} & \textbf{Basic} & \textbf{Distractor}
& \textbf{Background} & \textbf{Height} & \textbf{Lighting} \\
\midrule
\multicolumn{6}{@{}l@{}}{\makecell[l]{\textbf{Put in Drawer} (upper)\\[1pt]
\emph{``Put the watermelon in the upper drawer''}}} \\
SAM2Act+~\cite{fang2025sam2act} & 7/10  & 0/10 & 0/10  & 0/10 & 2/10 \\
\textbf{\method}               & 10/10 & 3/10 & 6/10 & 8/10 & 5/10 \\
\textbf{\memmethod}            & 10/10 & 3/10 & 10/10 & 8/10 & 4/10 \\
\midrule
\multicolumn{6}{@{}l@{}}{\makecell[l]{\textbf{Put in Drawer} (lower)\\[1pt]
\emph{``Put the watermelon in the lower drawer''}}} \\
SAM2Act+~\cite{fang2025sam2act} & 5/10  & 0/10 & 0/10  & 0/10 & 1/10 \\
\textbf{\method}               & 10/10 & 2/10 & 6/10 & 2/10 & 4/10 \\
\textbf{\memmethod}            & 10/10 & 5/10 & 10/10 & 7/10 & 6/10 \\
\midrule
\multicolumn{6}{@{}l@{}}{\makecell[l]{\textbf{Put on Shelf} (upper)\\[1pt]
\emph{``Put the soda water in the top shelf''}}} \\
SAM2Act+~\cite{fang2025sam2act} & 2/10  & 0/10  & 0/10  & 0/10  & 0/10 \\
\textbf{\method}               & 10/10 & 9/10  & 10/10 & 10/10 & 10/10 \\
\textbf{\memmethod}            & 10/10 & 10/10 & 10/10 & 10/10 & 10/10 \\
\midrule
\multicolumn{6}{@{}l@{}}{\makecell[l]{\textbf{Put on Shelf} (lower)\\[1pt]
\emph{``Put the red bull in the bottom shelf''}}} \\
SAM2Act+~\cite{fang2025sam2act} & 2/10  & 0/10  & 0/10  & 0/10 & 0/10 \\
\textbf{\method}               & 8/10  & 9/10  & 7/10  & 7/10 & 8/10 \\
\textbf{\memmethod}            & 10/10 & 10/10 & 10/10 & 8/10 & 10/10 \\
\bottomrule
\end{tabular}
\end{table*}

\subsection{General Manipulation on the Franka Platform}
\label{app:real_franka}

This appendix and
Appendices~\ref{app:real_dobot}--\ref{app:real_settings_vis} supplement
Sec.~\ref{sec:exp:real_world} with the per-task results, the baseline
failure modes, and the setting definitions of both real-robot suites.

\begin{figure*}[t]
  \centering
  \includegraphics[width=\textwidth]{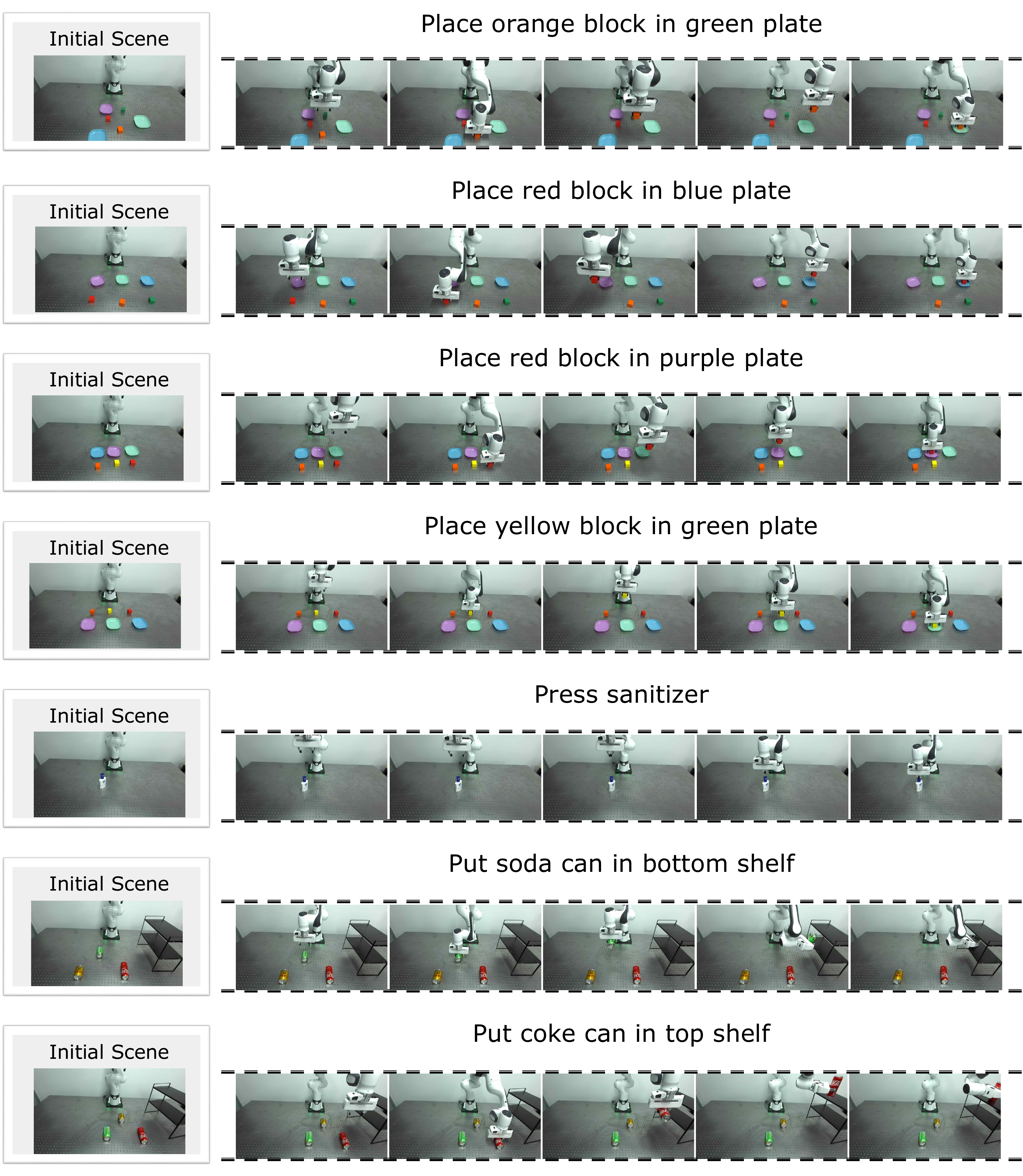}
  \caption{\textbf{Real-Robot Rollouts (I).}
  \method\ rollouts on the real-robot task suite of
  Appendix~\ref{app:real_franka}.}
  \label{fig:basic_task1}
\end{figure*}

\begin{figure*}[t]
  \centering
  \includegraphics[width=\textwidth]{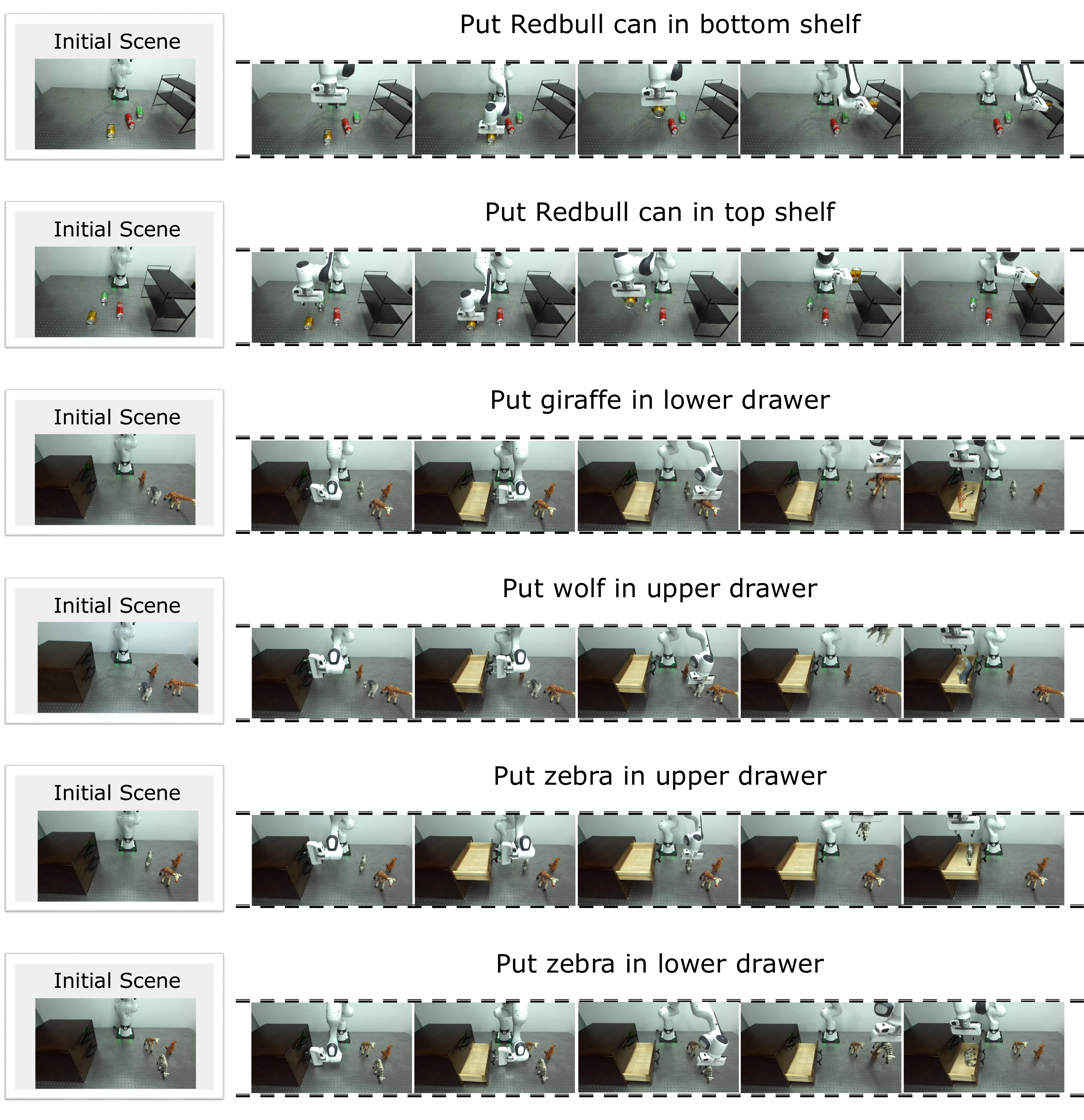}
  \caption{\textbf{Real-Robot Rollouts (II).}
  \method\ rollouts on the real-robot task suite of
  Appendix~\ref{app:real_franka}.}
  \label{fig:basic_task2}
\end{figure*}

\paragraph{Setup}
The 13 tasks of the Franka suite range from simple pick-and-place to
long-horizon drawer tasks, each spanning 3--9 keyframes
(Table~\ref{tab:real_robot}); Figs.~\ref{fig:basic_task1}
and~\ref{fig:basic_task2} show \method\ rollouts.
Demonstrations are collected by kinesthetic teaching: the manipulator
is moved to the keypoints of an expert trajectory, which are then
played back to record the observation and action at each keypoint.
Training and evaluation follow the protocol of
Appendix~\ref{app:eval_protocol}, and the six generalization settings
are defined in Appendix~\ref{app:real_settings_vis}.

\paragraph{Baselines}
The baselines span the design space laid out in
Sec.~\ref{sec:introduction}: SpatialVLA~\cite{qu2025spatialvla}, a 3D
VLA that injects 3D information through Ego3D position encoding,
trained with 10 and additionally with 50 trajectories per task;
$\pi_{0.5}$~\cite{intelligence2025pi_}, a 2D VLA whose flow-matching action
expert sits on the same PaliGemma backbone as \method;
ACT~\cite{zhao2023learning}, trained single-task per task; and
RVT-2~\cite{goyal2024rvt}, the projection-based 3D policy closest in
design to \method.

\paragraph{Data efficiency}
Among the baselines of Table~\ref{tab:real_robot}, the contrast with
$\pi_{0.5}$ is the cleanest real-world evidence for the alignment argument
of Sec.~\ref{sec:introduction}.
The two models share the same pre-trained backbone and differ in the
interface, heatmap prediction in the projected views versus action
generation through a separate flow-matching expert, so a gap of over 75
points at 10 demonstrations isolates the input--output alignment, not
the backbone, as the source of sample efficiency.
The failure modes of all baselines are recorded in
Appendix~\ref{app:real_baselines}, and the 3-demonstration variant in
Appendix~\ref{app:real_episode_analysis}.

\paragraph{Generalization}
Because only RVT-2 and \method\ perform well in the Basic setting, the
six generalization settings compare these two models
(Fig.~\ref{fig:real_results}).
The remaining failure mode of \method\ is \emph{Category}, where the
policy sometimes ignores the unseen target object and moves directly
to the destination.
This is not forgetting of the pre-trained grounding: fed samples from
the pre-training dataset after action fine-tuning, the model still
predicts accurate heatmaps
(Appendix~\ref{app:real_visualize_pretrain_heatmap}).
We attribute the gap instead to a residual domain mismatch: the 2D
pre-training images are mostly third-person views unlike the
orthographic robot renders, and their grounding supervision is object
localization, whereas manipulation targets keypoints that need not lie
on an object.
The pre-training ablation of Sec.~\ref{sec:exp:ablation} traces this
instruction-level generalization to the 2D-heatmap pre-training
itself.

\subsection{Memory-Dependent Manipulation on the Dobot Platform}
\label{app:real_dobot}

The Franka suite tests only the base policy: its near-saturated success
shows that these 13 tasks are solvable from the current frame alone.
We therefore evaluate \memmethod\ on a second, held-out embodiment, a
6-DoF Dobot CR5A collaborative arm fitted with a ChangingTek
CTAG2F90-C parallel-jaw electric gripper and observed, as in the Franka
setup, by a single static ZED~2i stereo camera.

\begin{figure*}[t]
  \centering
  \includegraphics[width=0.72\textwidth]{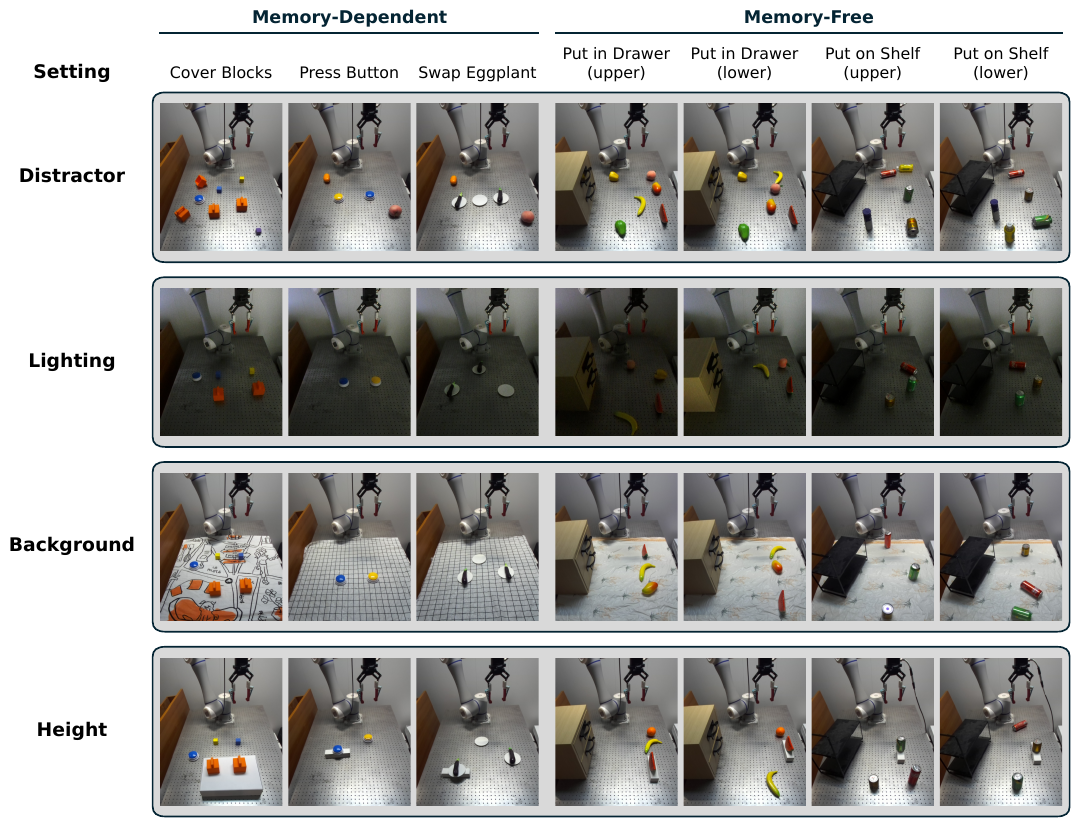}
  \caption{\textbf{The Distractor, Lighting, Background, and Height
  Settings on the Dobot Platform.}
  Initial scene of every instruction of the Dobot suite (columns, named
  as in Table~\ref{tab:app:real_dobot_all}) under each of the four
  visual-disturbance settings (rows;
  Appendix~\ref{app:real_dobot}).
  Frames in the \emph{Lighting} row are gamma-darkened for display
  where the camera's auto-exposure compensated for the reduced
  illumination; the policy receives the raw frames.}
  \label{fig:dobot_settings}
\end{figure*}

The three memory-dependent tasks transplant the memory families of
RMBench (Sec.~\ref{sec:exp:rmbench}) into the real world, each
constructed so that the current observation underdetermines the next
action.
\emph{Press Button} is the counting family: the robot must press the
blue button exactly three times and then press the yellow button once,
a final press that requires the policy to know when the counting is
complete.
In \emph{Cover Blocks}, the robot first covers the two
different-colored blocks in the workspace and must then uncover only
the block of the instructed color, which is solvable only from the
color-to-location bindings formed before the covers went on.
\emph{Swap Eggplant} is the rearrangement family: two look-alike
eggplants must each end up on the other's initial plate, using an
initially empty third plate as a buffer, so the correct next placement
depends on which eggplant has already been moved, which the current
frame does not reveal.
The two memory-free tasks, \emph{Put in Drawer} and \emph{Put on
Shelf}, are solvable from the current frame alone and test whether the
memory extension costs general manipulation capability.
Training data, baselines, and the five evaluation settings are stated
in Appendix~\ref{app:eval_protocol};
Figs.~\ref{fig:dobot_rollouts1} and~\ref{fig:dobot_rollouts2} show
\memmethod\ rollouts, Fig.~\ref{fig:dobot_settings} the four
visual-disturbance settings, and Table~\ref{tab:app:real_dobot_all}
the per-instruction counts behind Tables~\ref{tab:real_dobot_basic}
and~\ref{tab:real_dobot_generalization}.

\subsection{Real-Robot Baseline Failure Modes}
\label{app:real_baselines}

Tables~\ref{tab:real_robot} and~\ref{tab:app:real_dobot_all} aggregate
the comparisons on the two platforms; the notes below record how each
baseline fails.

\paragraph{SpatialVLA~\cite{qu2025spatialvla}}
Trained with 10 trajectories per task, SpatialVLA fails on nearly all
tasks, typically without even moving toward the correct target object.
Raising the training set to 50 trajectories per task recovers some
performance, 28.5\% against 3.1\%, but it remains far behind \method,
particularly on harder tasks such as \emph{Put the Giraffe in the Lower
Drawer}.

\paragraph{$\pi_{0.5}$~\cite{intelligence2025pi_}}
$\pi_{0.5}$ achieves occasional success on simple pick-and-place tasks but
consistently fails on more complex, long-horizon tasks, such as
\emph{Put Zebra in Drawer}.
We further observe that its motions are often unstable and that it tends to
close the gripper prematurely.
In contrast, \method performs reliably across all evaluated
tasks.

\paragraph{ACT~\cite{zhao2023learning}}
ACT generalizes poorly in space: it succeeds in regions densely covered
by the demonstrations but often fails when the target lies near the
workspace boundary.
This is consistent with its design, since ACT models actions under a
Gaussian prior, which assigns low probability to peripheral regions.

\paragraph{RVT-2~\cite{goyal2024rvt}}
RVT-2 is the strongest baseline of the Franka suite and solves most
tasks, but it is less robust than \method: it sometimes grasps a block
imprecisely or places an object inaccurately, and its gap to \method\
widens further in the generalization settings.

\paragraph{SAM2Act+~\cite{fang2025sam2act}}
The failures of SAM2Act+ on the memory-dependent Dobot tasks match the
memory-management analysis of Sec.~\ref{sec:exp:real_world}.
In \emph{Cover Blocks}, it cannot tell which cover hides the instructed
block: by the time this decision is made, about 11 history steps have
already accumulated, and its fixed-size memory window is filled with
near-duplicate frames, so the block colors seen at the start are
largely lost.
In \emph{Press Button}, it has no explicit sense of how many presses
have been completed, and keeps pressing the button endlessly.

\subsection{Sample Efficiency with 3 vs.\ 10 Demonstrations}
\label{app:real_episode_analysis}

Table~\ref{tab:app:real_episode} lists the per-task success counts of
\method\ when trained with 3 rather than 10 demonstrations per task:
the policy stays at or above 7/10 on every task and matches the
10-demonstration model on most of them.

\subsection{Preservation of Object Grounding after Fine-Tuning}
\label{app:real_visualize_pretrain_heatmap}

\begin{figure*}[t]
  \centering
  \includegraphics[width=0.85\textwidth]{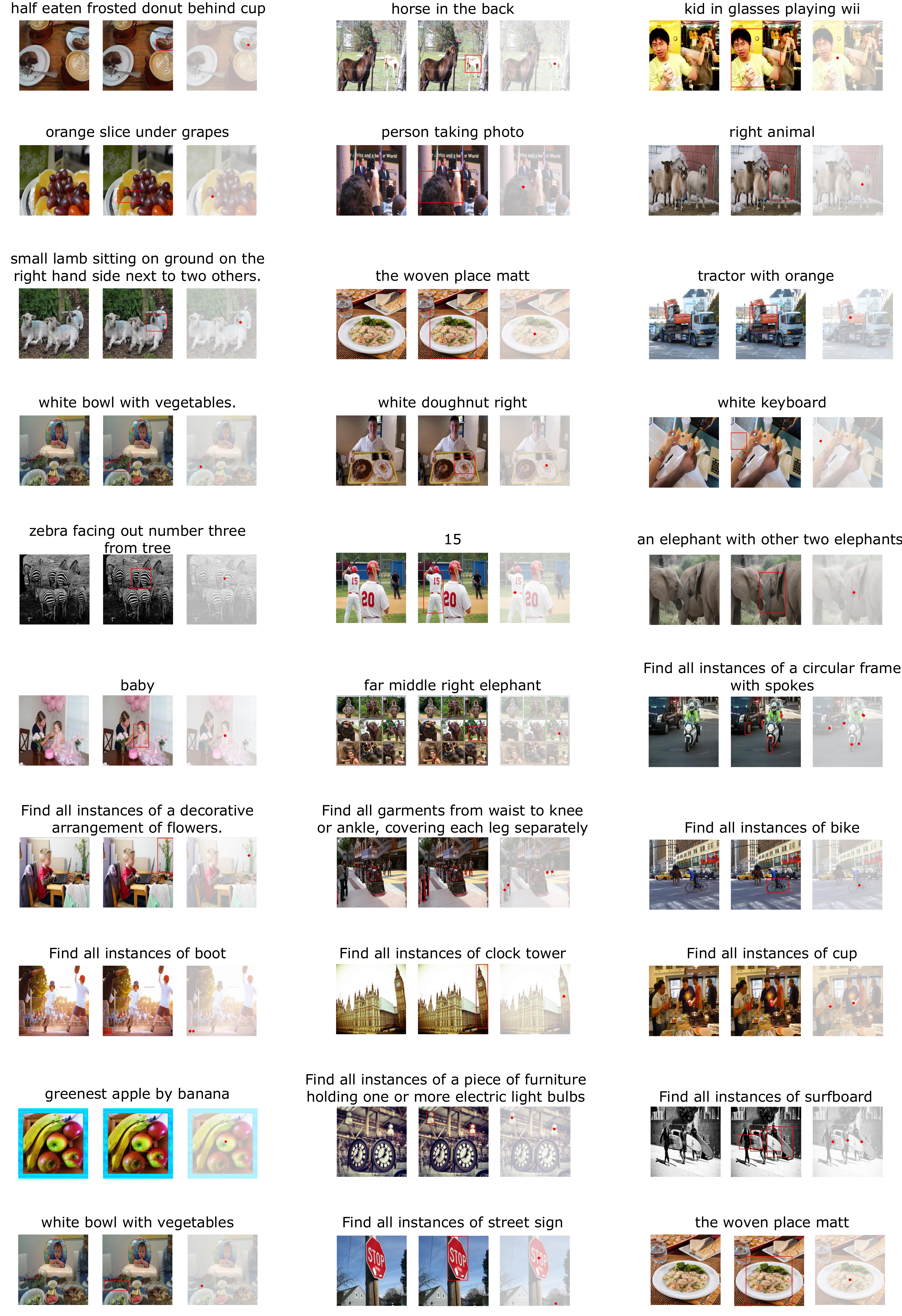}
  \caption{\textbf{Ground-Truth Heatmap Construction on Detection
  Data.}
  For each sample: the original image (\emph{left}), the bounding
  boxes of the objects of interest (\emph{middle}), and the
  ground-truth heatmap rendered from the box centers (\emph{right}).}
  \label{fig:pretrain_dataset}
\end{figure*}

\begin{figure*}[t]
  \centering
  \includegraphics[width=0.72\textwidth]{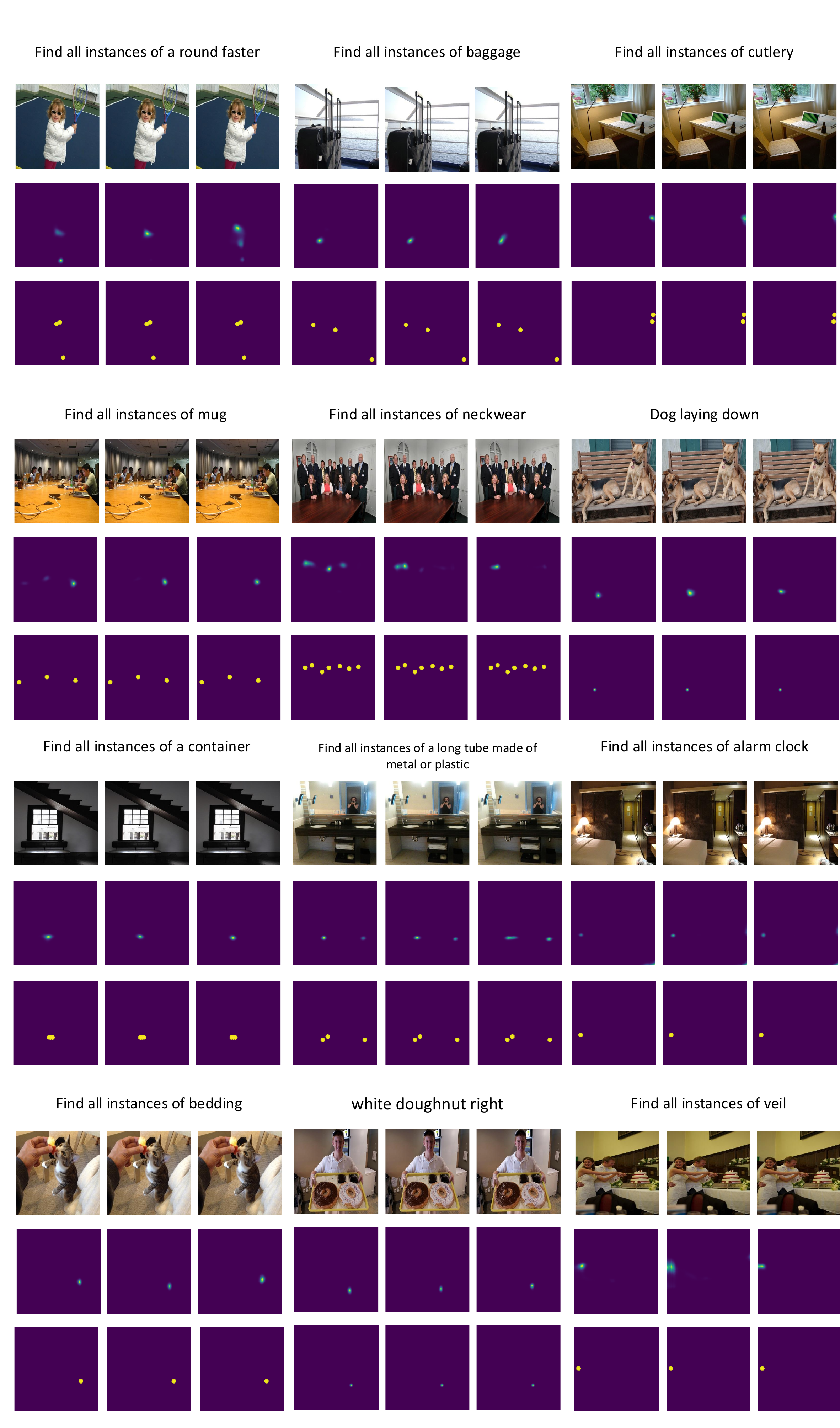}
  \caption{\textbf{Predictions on Pre-Training Data after
  Fine-Tuning.}
  Each input image is repeated three times to mimic the multi-view
  input format of fine-tuning.
  Rows per sample: input image, predicted heatmaps, ground truth.
  Samples are not cherry-picked.}
  \label{fig:real_heatmaps}
\end{figure*}

Even after fine-tuning on robot action data, \method\ retains the
object grounding installed by the 2D-heatmap pre-training.
Fig.~\ref{fig:pretrain_dataset} illustrates how the pre-training
targets are constructed on detection data, as truncated Gaussians
rendered at the annotated box centers and normalized into one
distribution (Sec.~\ref{sec:bridgevla:pretrain}).
Fig.~\ref{fig:real_heatmaps} then visualizes the fine-tuned model's
predictions on pre-training samples, with each input image repeated
three times to simulate the multi-view input of fine-tuning.
These samples are not cherry-picked, which confirms that \method\ does
not forget its pre-training knowledge after 3D action fine-tuning.

\subsection{Real-Robot Generalization Settings}
\label{app:real_settings_vis}

\begin{figure*}[t]
  \centering
  \includegraphics[width=0.72\textwidth]{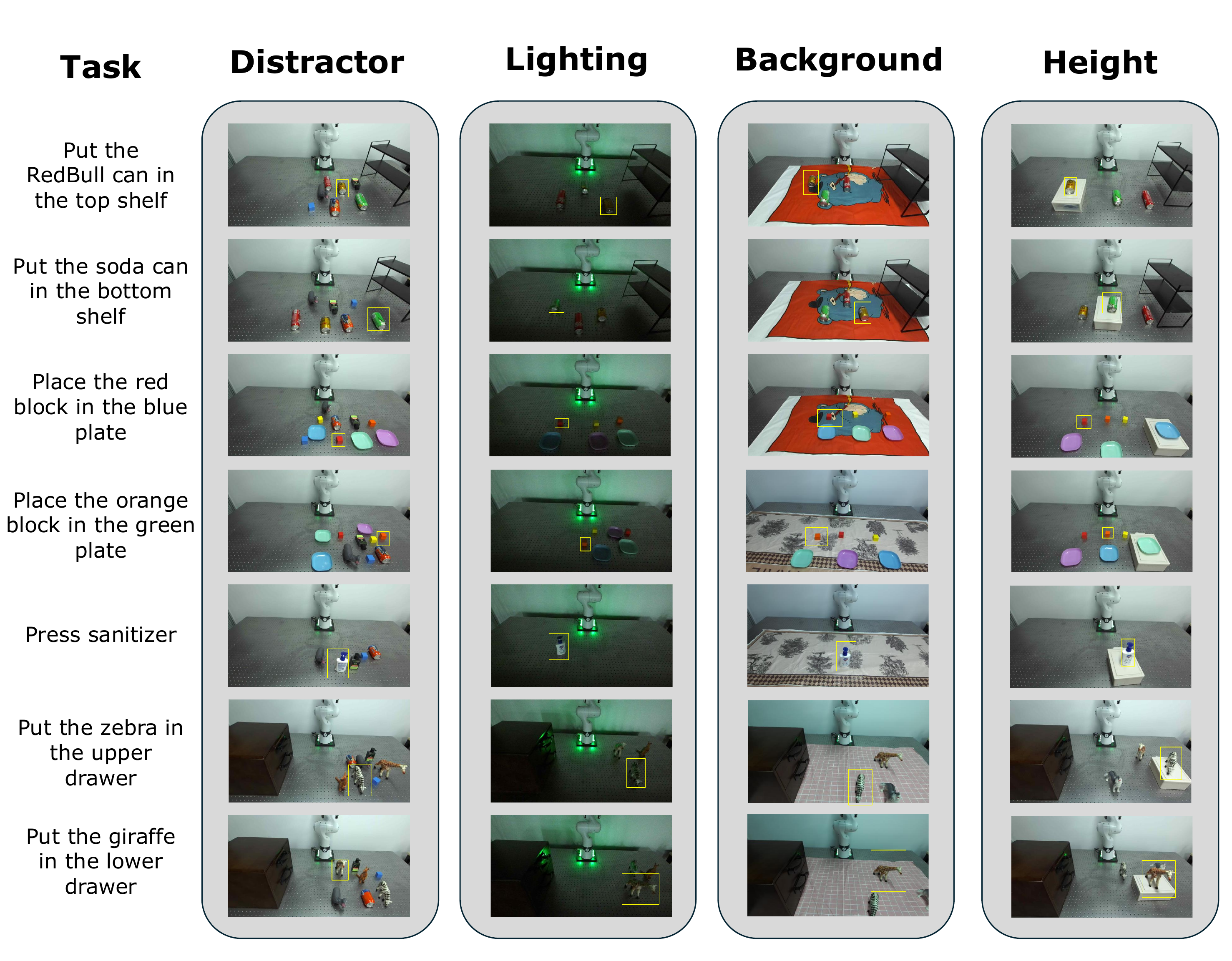}
  \caption{\textbf{The Distractor, Lighting, Background, and Height
  Settings.}
  Visualization of the four visual-disturbance settings of the
  real-robot evaluation (Appendix~\ref{app:real_franka}).}
  \label{fig:real_settings}
\end{figure*}

\begin{figure*}[t]
  \centering
  \includegraphics[width=\textwidth]{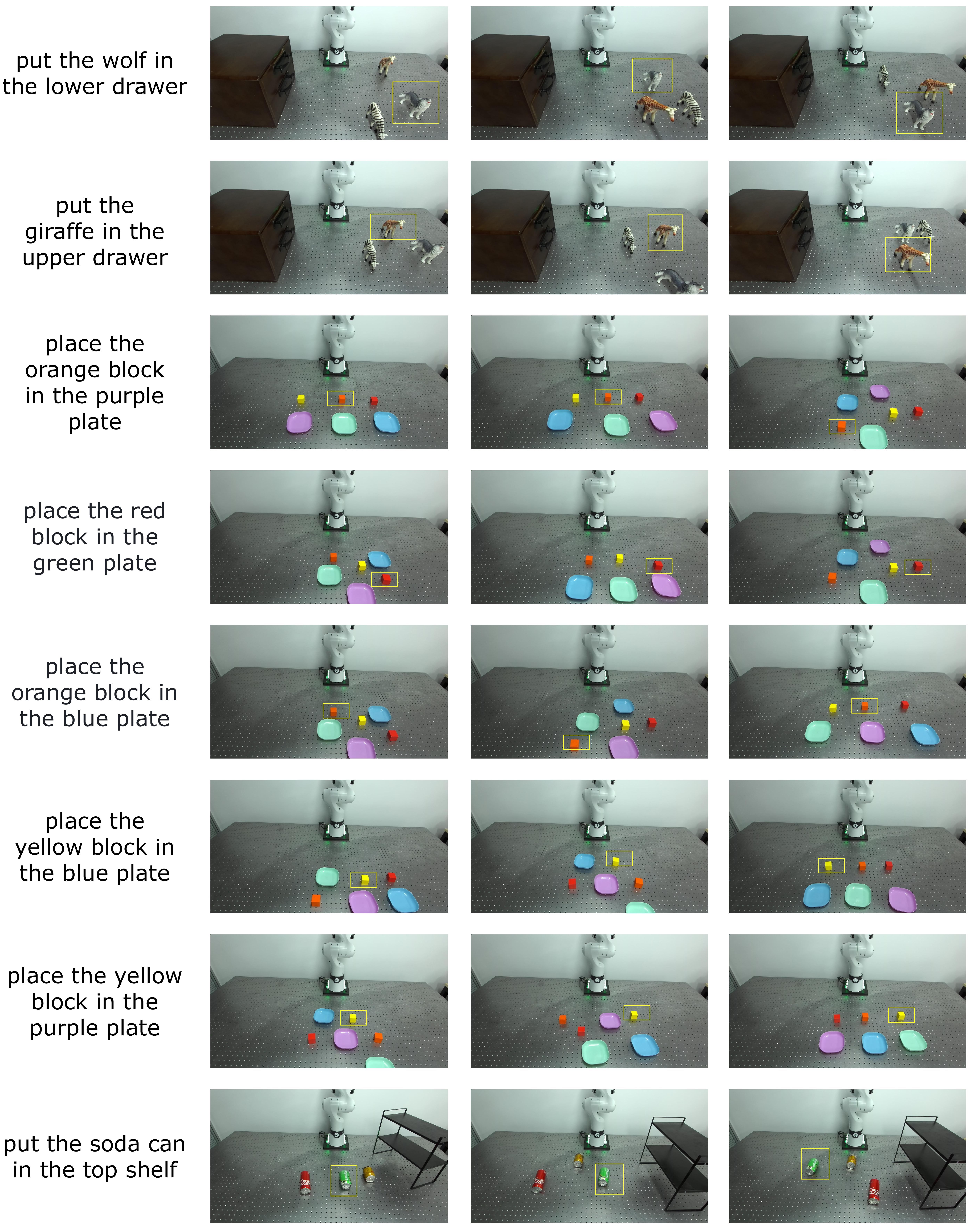}
  \caption{\textbf{The Combination Setting (I).}
  During training, the manipulated objects and skills are seen, but
  their combinations are unseen.}
  \label{fig:combination_within}
\end{figure*}

\begin{figure*}[t]
  \centering
  \includegraphics[width=\textwidth]{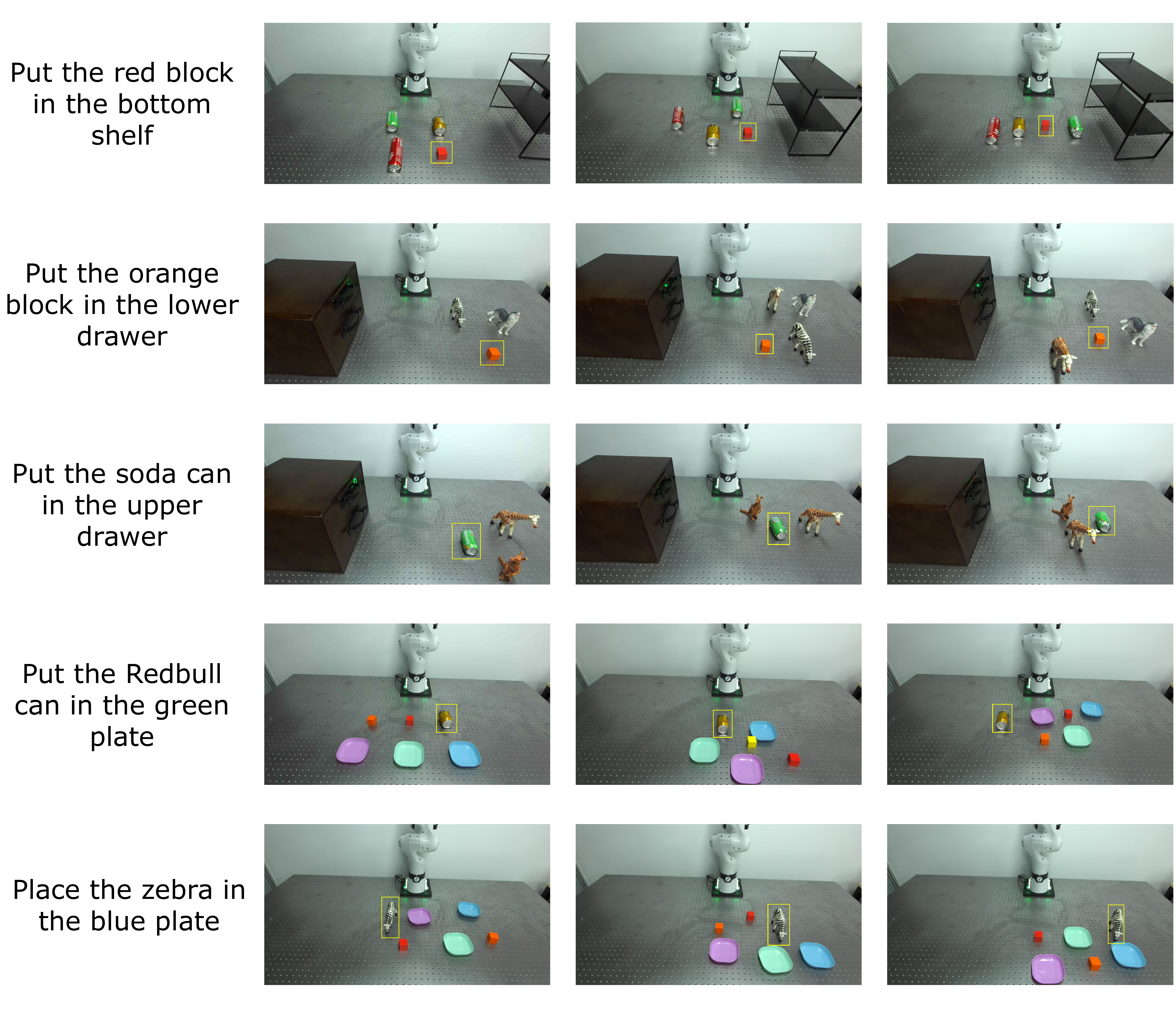}
  \caption{\textbf{The Combination Setting (II).}
  During training, the manipulated objects and skills are seen, but
  their combinations are unseen.}
  \label{fig:combination_across}
\end{figure*}

\begin{figure*}[t]
  \centering
  \includegraphics[width=0.85\textwidth]{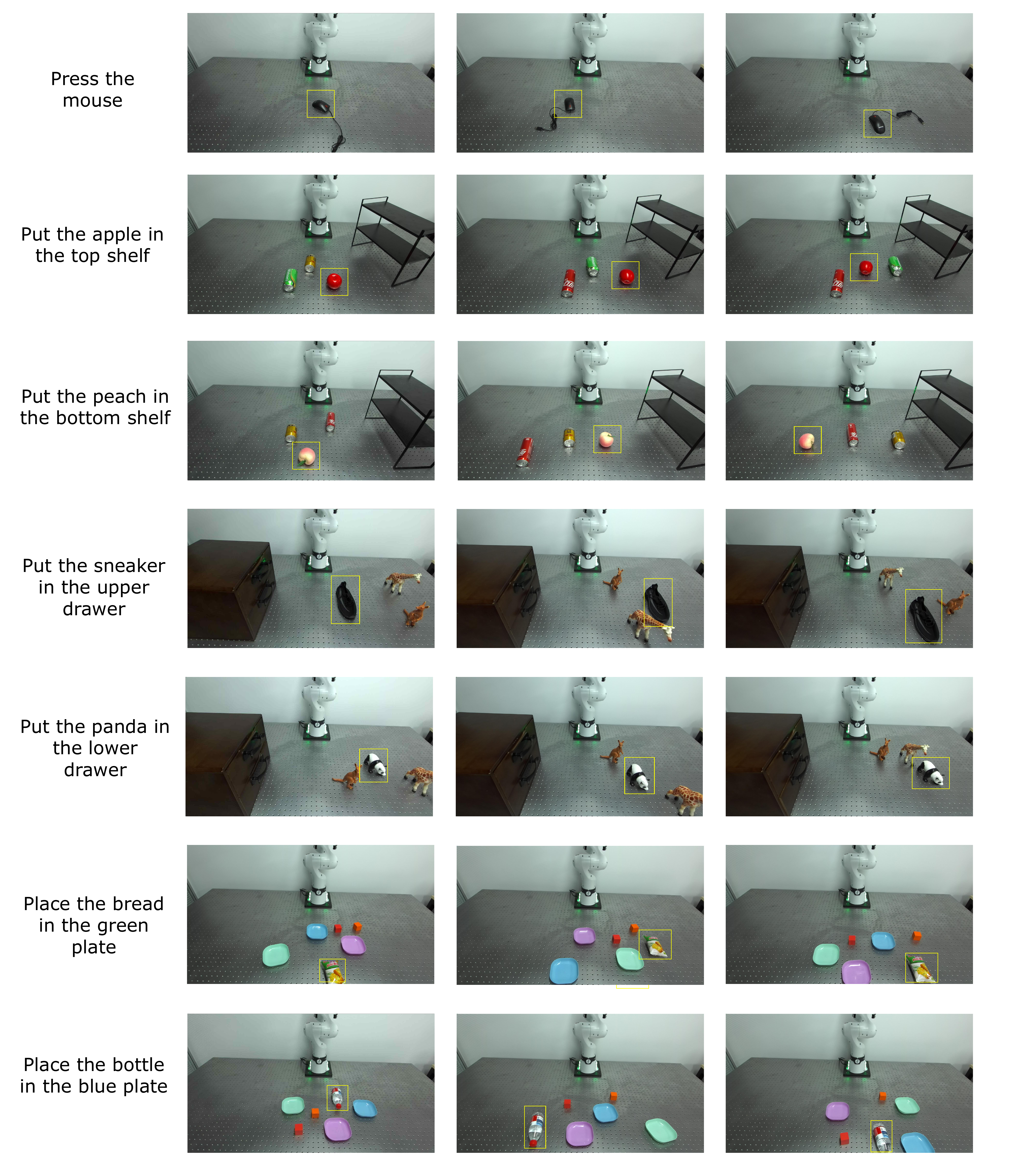}
  \caption{\textbf{The Category Setting.}
  In total, we evaluate on 7 objects from categories that are unseen
  during training.}
  \label{fig:category}
\end{figure*}

The six generalization settings are defined as follows.
\emph{Distractor} adds distractor objects visually similar to at least
one target object; \emph{Lighting} turns the lights off;
\emph{Background} changes the tablecloth, in three variants; and
\emph{Height} raises some objects onto a drawer or box.
\emph{Combination} pairs objects and skills that were each seen in
training into 13 pairings never demonstrated together, and
\emph{Category} introduces 7 objects from categories unseen in the
robot training data.
Fig.~\ref{fig:real_settings} shows the four visual-disturbance
settings, Figs.~\ref{fig:combination_within}
and~\ref{fig:combination_across} the Combination setting, and
Fig.~\ref{fig:category} the Category setting.

\balance

\clearpage

\begin{figure*}[t]
  \centering
  \includegraphics[width=0.62\textwidth]{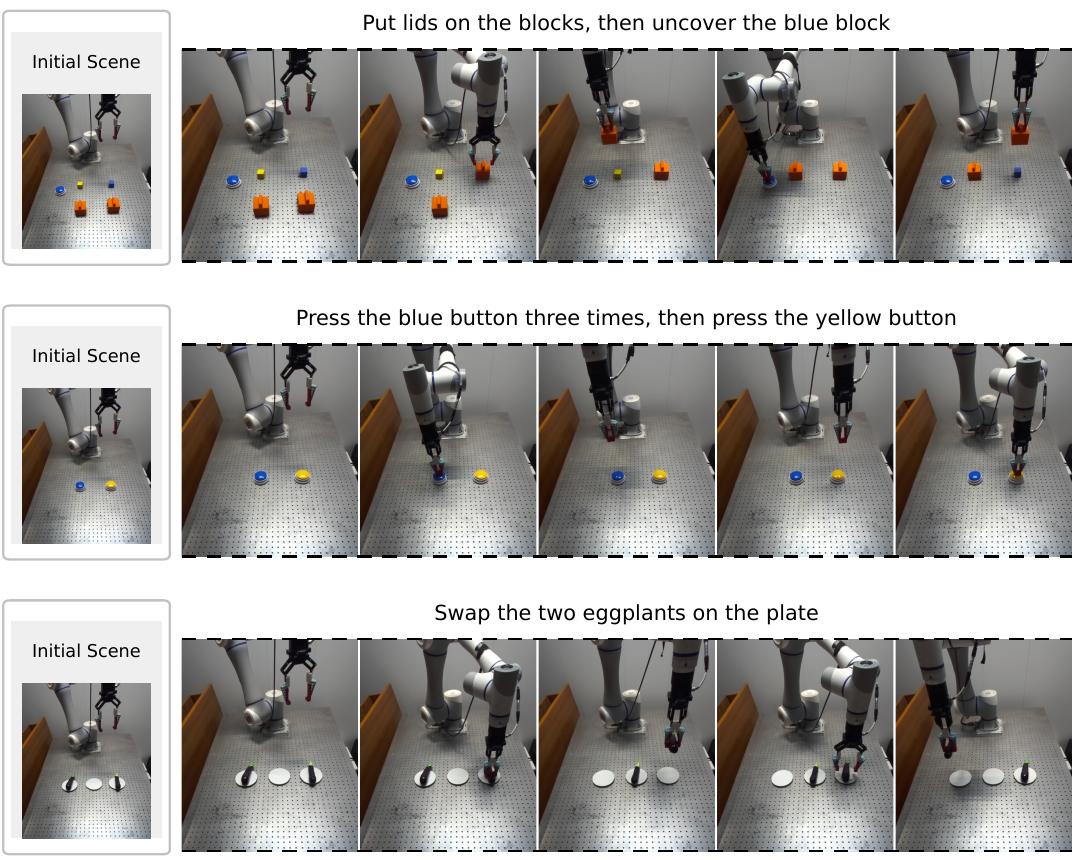}
  \caption{\textbf{Dobot Rollouts (I): Memory-Dependent Tasks.}
  \memmethod\ rollouts on the three memory-dependent instructions of
  the Dobot suite (Appendix~\ref{app:real_dobot}) in the Basic
  setting; each strip shows five keyframes of one successful episode.}
  \label{fig:dobot_rollouts1}
\end{figure*}

\begin{figure*}[t]
  \centering
  \includegraphics[width=0.62\textwidth]{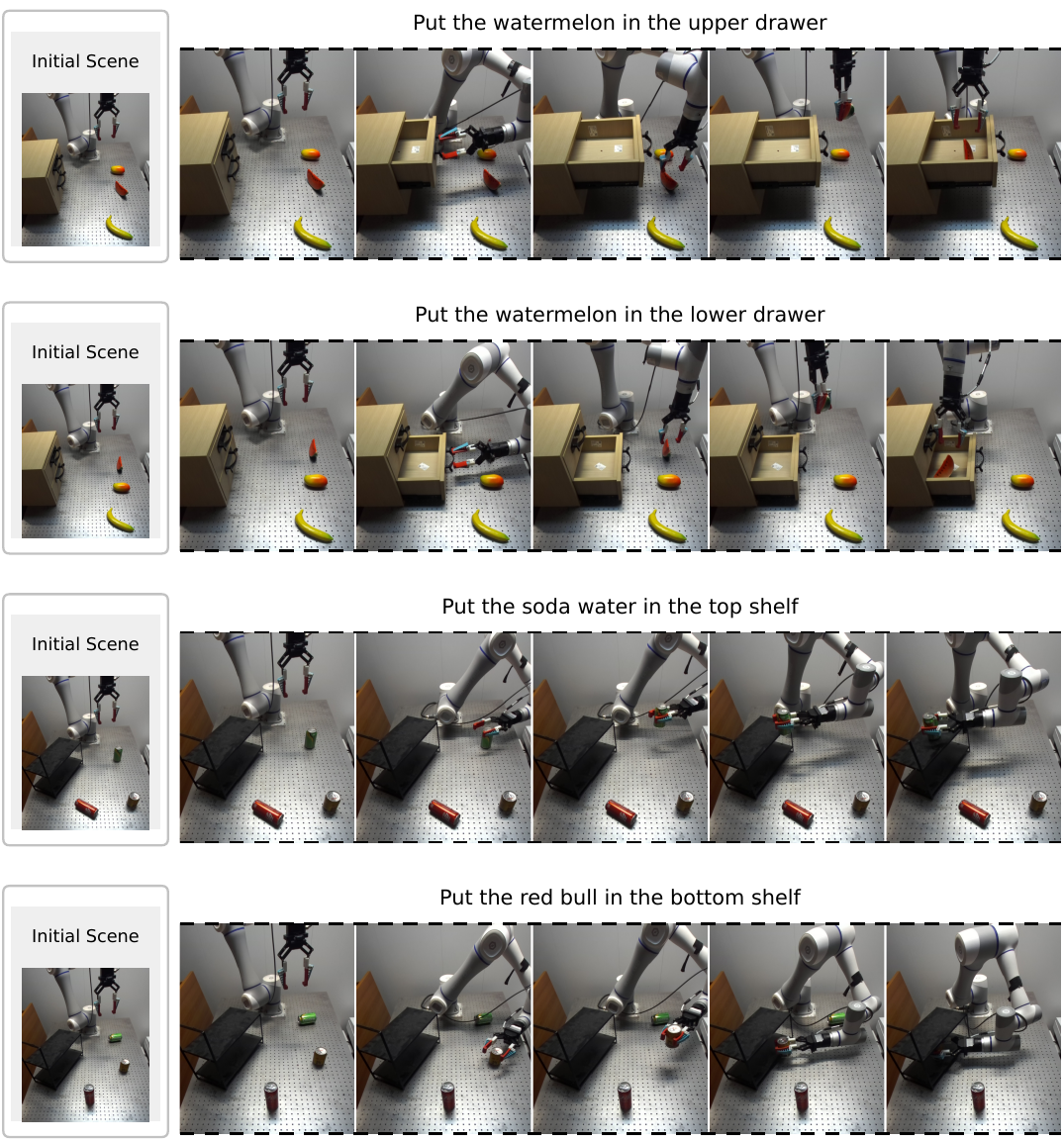}
  \caption{\textbf{Dobot Rollouts (II): Memory-Free Tasks.}
  \memmethod\ rollouts on the four memory-free instructions of the
  Dobot suite in the Basic setting, laid out as in
  Fig.~\ref{fig:dobot_rollouts1}.}
  \label{fig:dobot_rollouts2}
\end{figure*}

\end{document}